\documentclass[journal,compsoc,onecolumn]{IEEEtran}
\usepackage[utf8]{inputenc}
\usepackage{amsfonts}
\usepackage{amsmath,bm}
\usepackage{amssymb}
\usepackage{mathtools}
 \usepackage{subfigure}
\usepackage{graphicx}
\graphicspath{ {Images/} }
\usepackage{lipsum}
\usepackage{fancyhdr}
\usepackage{algorithm} 
\usepackage{algpseudocode} 
\usepackage{framed} 
\usepackage{multirow} 
\usepackage{adjustbox} 
\usepackage[flushleft]{threeparttable} 
\usepackage{booktabs} 
\usepackage{listofitems} 
\usepackage{adjustbox}
\usepackage{cite}
\usepackage{tabularx}

\usepackage[colorlinks,linkcolor=blue]{hyperref} 
\hypersetup{draft} 

\usepackage{pifont}
\usepackage{array}
\newcolumntype{L}[1]{>{\raggedright\let\newline\\\arraybackslash\hspace{0pt}}m{#1}}
\newcolumntype{C}[1]{>{\centering\let\newline\\\arraybackslash\hspace{0pt}}m{#1}}
\newcolumntype{R}[1]{>{\raggedleft\let\newline\\\arraybackslash\hspace{0pt}}m{#1}}

\usepackage{xcolor}

\usepackage{xcolor}

\usepackage{ragged2e} 

\begin{document}
%
\title{Generative Adversarial Networks in Computer Vision: A Survey and Taxonomy}


\author{Zhengwei Wang,
        Qi She,
        Tom\'as E. Ward
\thanks{Zhengwei Wang is with V-SENSE, School of Computer Science and Statistics, Trinity College Dublin, Dublin, Ireland. e-mail: villa.wang.zhengwei@gmail.com}
\thanks{Qi She is with ByteDance AI Lab, Beijing, China. e-mail: sheqi1991@gmail.com}
\thanks{Tom\'as E. Ward is with Insight Centre for Data Analytics, Dublin City University, Dublin, Ireland. e-mail: tomas.ward@dcu.ie}
\thanks{\textbf{Accepted by ACM Computing Surveys, 23 November 2020}}
}

\IEEEcompsoctitleabstractindextext{
\begin{abstract}
\justify
Generative adversarial networks (GANs) have been extensively studied in the past few years. Arguably their most significant impact has been in the area of computer vision where great advances have been made in challenges such as plausible image generation, image-to-image translation, facial attribute manipulation and similar domains. Despite the significant successes achieved to date, applying GANs to real-world problems still poses significant challenges, three of which we focus on here. These are: (1) the generation of high quality images, (2) diversity of image generation, and (3) stable training. Focusing on the degree to which popular GAN technologies have made progress against these challenges, we provide a detailed review of the state of the art in GAN-related research in the published scientific literature. We further structure this review through a convenient taxonomy we have adopted based on variations in GAN architectures and loss functions.  While several reviews for GANs have been presented to date, none have considered the status of this field based on their progress towards addressing practical challenges relevant to computer vision.  Accordingly, we review and critically discuss the most popular architecture-variant, and loss-variant GANs, for tackling these challenges. Our objective is to provide an overview as well as a critical analysis of the status of GAN research in terms of relevant progress towards important computer vision application requirements. As we do this we also discuss the most compelling applications in computer vision in which GANs have demonstrated considerable success along with some suggestions for future research directions. Code related to the GAN-variants studied in this work is summarized on \url{https://github.com/sheqi/GAN_Review}.
\end{abstract}

}

\maketitle

%

\IEEEpeerreviewmaketitle

\section{Introduction}
\IEEEPARstart{G}{enerative} adversarial networks (GANs) are attracting growing interest in the deep learning community~\cite{goodfellow2014generative,liu2016coupled,salimans2016improved,turkoglu2019layer,wu2017gp,pan2017salgan}. GANs have been applied to various domains such as computer vision~\cite{dziugaite2015training,ma2017pose,vondrick2016generating,yang2017high,odena2017conditional,li2015generative,zhu2016generative,lassner2017generative}, natural language processing~\cite{fedus2018maskgan,yang2017semi,dai2017good,jetchev2016texture}, time series synthesis~\cite{donahue2018synthesizing,hartmann2018eeg,esteban2017real,li2019mad,brophy2019quick}, semantic segmentation~\cite{zhu2016adversarial,luc2016semantic,dong2017semantic,qiu2017deep,souly2017semi} etc. GANs belong to the family of generative models in machine learning. Compared to other generative models e.g., variational autoencoders, GANs offer advantages such as an ability to handle sharp estimated density functions, efficiently generating desired samples, eliminating deterministic bias and with good compatibility with the internal neural architecture~\cite{goodfellow2016nips}. These properties have allowed GANs to enjoy great success especially in the field of computer vision e.g., plausible image generation~\cite{karras2018style,wang2018high,poole2016improved,choe2017face,zhang2017stackgan}, image-to-image translation~\cite{zhu2017unpaired,liu2016coupled,zhu2017toward,tomei2018art2real,liu2017unsupervised,isola2017image,choi2018stargan,ma2018gan}, image super-resolution~\cite{ledig2017photo,wang2018esrgan,mahapatra2017image,dong2017semantic,mahapatra2019image} and image completion~\cite{yu2018generative,yeh2017semantic,dolhansky2018eye,chen2018high,li2017generative}. 

However, GANs are not without problems. The two most significant are that they are hard to train and that they are difficult to evaluate. In terms of being hard to train, it is non-trivial for the discriminator and generator to achieve Nash equilibrium during the training and it is common for the generator to fail to learn well, the full distribution of the datasets. This is the well-known mode collapse issue. Lots of work has been carried out in this area~\cite{kossaifi2018gagan,dai2018adversarial,kodali2017convergence,li2017dualing}. In terms of evaluation, the primary issue is how best to measure the dissimilarity between the real distribution of the target $p_{r}$ and the generated distribution $p_{g}$. Unfortunately accurate estimation of $p_{r}$ is not possible. Thus, it is challenging to produce good estimations of the correspondence between $p_{r}$ and $p_{g}$. Previous work has proposed various evaluation metrics for GANs~\cite{borji2019pros,xu2018empirical,gulrajani2017improved,heusel2017gans,gretton2012kernel,wang2018use,wang2019neuroscore,barratt2018note,theis2015note}. The first aspect concerns the performance for GANs directly e.g., image quality, image diversity and stable training. In this work, we are going to study existing GAN-variants that handle this aspect in the area of computer vision while those readers interested in the second aspect can consult ~\cite{borji2019pros,theis2015note}. 

Much of current GAN research can be considered in terms of the following two objecetives: (1) Improving training, and (2) the deployment of GANs for real-world applications. The former seeks to improve GANs performance and is therefore a foundation for the latter, i.e. applications. Considering the many published results which deal with GAN training improvement, we present a succinct review on the most important GAN-variants that focus on this aspect in this paper. The improvement of the training process provides benefits in terms of GANs performance as follows: (1) improvements in the generated image diversity (also known as mode diversity), (2) increases in generated image quality, and (3) more stable training such as remedying the vanishing gradient issue for the generator. In order to improve the performance as mentioned above, modification for GANs can be done from either the architectural side or the loss perspective. We will study the GAN-variants coming from both sides that improve the performance for GANs. 
\begin{figure*}[htbp]
    \centering
    \includegraphics[width=.7\textwidth]{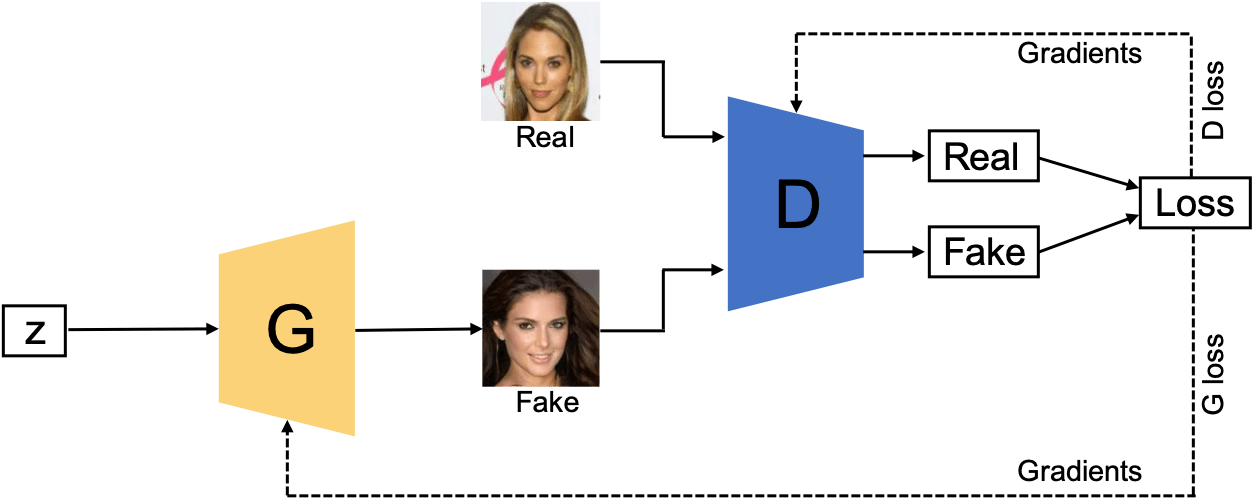}
    \caption[Architecture of a GAN.]{Architecture of a GAN. Two deep neural networks (discriminator ($D$) and generator $G$)) are synchronously trained during the learning stage. The discriminator is optimized in order to distinguish between real images and generated images while the generator is trained for the sake of fooling the discriminator from discerning between real images and generated images.}
    \label{fig:gan-architecture}
\end{figure*}
The rest of the paper is organized as follows: (1) We introduce related review work for GANs and illustrate the difference between those reviews and this work; (2) We give a brief introduction to GANs; (3) We review the architecture-variant GANs in the literature; (4) We review the loss-variant GANs in the literature; (5) We introduce some GAN-based applications mianly in the area of computer vision; (6) We introduce evaluation metrics for GANs and also compared GAN-variants discussed in this paper by using part of metrics i.e., Inception Score and Fre\'chet Inception Distance(FID); (7) We summarize the GAN-variants in this study and illustrate their difference and relationships and also discuss several avenues for future research regarding GANs and (8) We conclude this review and preview likely future research work in the area of GANs.  

Many GAN-variants have been proposed in the literature to improve performance. These can be divided into two types: (1) Architecture-variants. The first proposed GAN used fully-connected neural networks~\cite{goodfellow2014generative} so specific types of architecture may be beneficial for specific applications e.g., convolutional neural networks (CNNs) for images and recurrent neural networks (RNNs) for time series data; and (2) Loss-variants. Here different variations of the loss function~\eqref{eq:GAN-formula} are explored to enable more stable learning of $G$.

\begin{figure}[!htbp]
    \centering
    \includegraphics[width=1\textwidth]{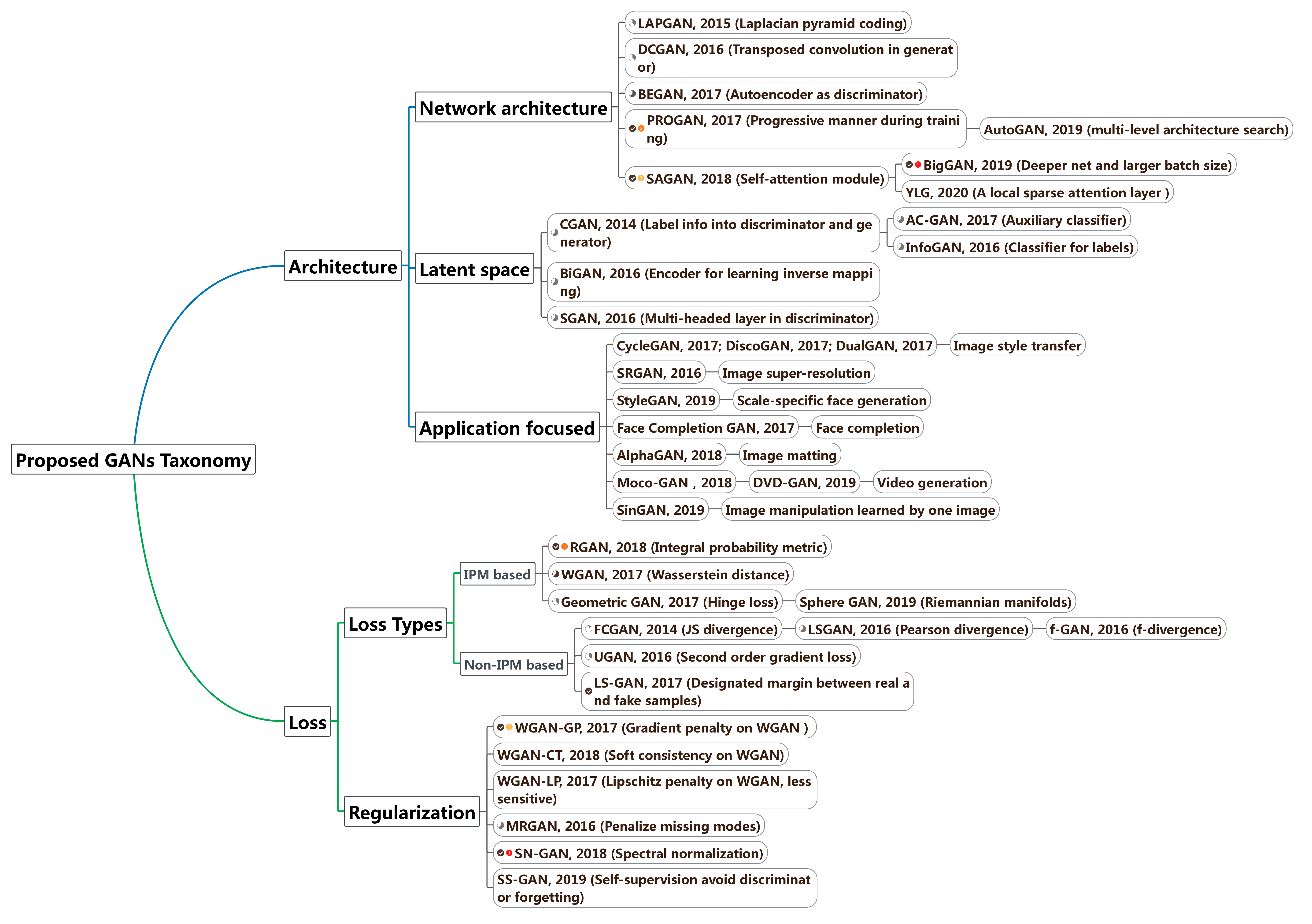}
    \caption{The proposed taxonomy of the recent GANs.}
    \label{chap05-fig:GAN-mindmap}
\end{figure}

Figure~\ref{chap05-fig:GAN-mindmap}
illustrates our proposed taxonomy for the representative GANs present in the literature from 2014 to 2020. We divide current GANs into two main variants i.e., the architecture-variant and the loss-variant. In the architecture-variant, we summarize three categories, which are network architecture, latent space and application focused respectively. The network architecture category refers to improvement or modification made on overall the GAN architecture e.g., progressive mechanism deployed in PROGAN. The latent space category indicates that the architecture modification is made based on different representation of latent space e.g., CGAN involves label information encoded to both generator and discriminator. The last category, application focused, refers to modification made according to different applications e.g., CycleGAN has specific architecture which deals with image style transfer. In terms of the loss-variants, we divide two it into two categories, loss types and regularization. Loss types refers to different loss functions to be optimized for GANs and regularization refers to additional penalization designed to the loss function or any type of normalization operation made to the network. More specifically, we divide the loss function into integral probability metric (IPM) based and non-IPM based. In IPM-based GANs, the \textbf{discriminator} is constrained to a specific class of function~\cite{jolicoeur2018relativistic} e.g., the discriminator in WGAN is constrained to 1-Lipschitz. The discriminator in non-IPM based GANs does not have such constraint.

\section{Related Reviews}
There has been previous GANs review papers for example in terms of reviewing GANs performance ~\cite{kurach2018gan}. That work focuses on the experimental validation across different types of GANs benchmarking on LSUN-BEDROOM~\cite{yu2015lsun}, CELEBA-HQ-128~\cite{liu2018large} and the CIFAR10~\cite{krizhevsky2009learning} image datasets. The results suggest that the original GAN~\cite{goodfellow2014generative} with spectral normalization~\cite{yoshida2017spectral} is a good starting choice when applying GANs to a new dataset. A limitation of that review is that the benchmark datasets do not consider diversity in a significant way. Thus the benchmark results tend to focus more on evaluation of the image quality, which may ignore GANs efficacy in producing diverse images. Work~\cite{hitawala2018comparative} surveys different GANs architectures and their evaluation metrics. A further comparison on different architecture-variants' performance, applications, complexity and so on needs to be explored. Papers~\cite{wang2017generative,creswell2018generative,hong2019generative} focus on the investigation of the newest development treads and the applications of GANs. They compare GAN-variants through different applications. 

Comparing this review to the current review in the literature, we emphasize an introduction to GAN-variants based on their performance including their ability to produce high quality and diverse images, stable training, ability for handling the vanishing gradient problem, etc. This is all done through the taking of a perspective based on architecture and loss function considerations. This perspective is important because it covers fundamental challenges for GANs and it will help researchers how to select a proper architecture and a proper loss function for a typical GAN according to a specific application. It also gives a footprint that how researchers dealed with those problems before to those researchers who will dive into this area. Our search strategy and searched results is presented in Appendix~\ref{app:search}. A detail of searched papers are listed on this link: \textit{\url{https://github.com/sheqi/GAN_Review/blob/master/GAN_CV.csv}.}

In summary, our contributions of this review are three-fold:
\begin{itemize}
    \item We focus on GANs by addressing three important problems: (1) High-quality image generation; (2) Diverse image generation; and (3) Stable training.
    
    \item We propose the novel GAN taxonomy and introduce recent GANs from two perspectives: (1) Architecture of generators and discriminators, e.g., network architecture, latent space, and application driven design; and (2) The objective function for training, e.g., loss design in IPM based and non-IPM based methods, regularization approaches. Compared to other reviews on GANs, this review provides a unique view to different GAN variants.
    
    \item We also provide the comparison and analysis in terms of pros and cons across GAN-variants presented in this paper. 
\end{itemize}

\section{Generative Adversarial Networks}
Figure.~\ref{fig:gan-architecture} demonstrates the architecture of a typical GAN. The architecture comprises two components, one of which is a discriminator ($D$) distinguishing between real images and generated images while the other one is a generator ($G$) creating images to fool the discriminator. Given a distribution $\mathbf{z} \sim p_{\mathbf{z}}$, $G$ defines a probability distribution $p_{g}$ as the distribution of the samples $G(\mathbf{z})$. The objective of a GAN is to learn the generator's distribution $p_{g}$ that approximates the real data distribution $p_{r}$. Optimization of a GAN is performed with respect to a joint loss function for $D$ and $G$
\begin{equation} \label{eq:GAN-formula}
\begin{aligned}
	\min \limits_{G} \max\limits_{D}\hspace{2pt}&\mathbb{E}_{\bm{\mathrm{x}} \sim p_{r}} \mathrm{log}[D(\bm{\mathrm{x}})] + \mathbb{E}_{\bm{\mathrm{z}} \sim p_{\bm{\mathrm{z}}}} \mathrm{log}\left[1 - D(G(\bm{\mathrm{z}}))\right].
\end{aligned}
\end{equation}
GANs, as a member of the deep generative model (DGM) family, has attracted exponentially growing interest in the deep learning community because of some advantages comparing to the tradition DGMs: (1) GANs are able to produce better output than other DGMs. Comparing to the most well-known DGMs---variational autoencoder (VAE), GANs are able to produce any type of probability density while VAE is not able to generate sharp images~\cite{goodfellow2016nips}; (2) The GAN framework can train any type of generator network. Other DGMs may have pre-requirements for the generator e.g., the output layer of generator is Gaussian~\cite{goodfellow2016nips,doersch2016tutorial,kingma2013auto}; (3) There is no restriction on the size of the latent variable. These advantages have led GANs to achieve the state of art performance on producing synthetic data especially for image data. 

\section{\textbf{Architecture-variant GANs}}
\begin{figure*}[!htbp]
    \centering
    \includegraphics[width=0.85\textwidth]{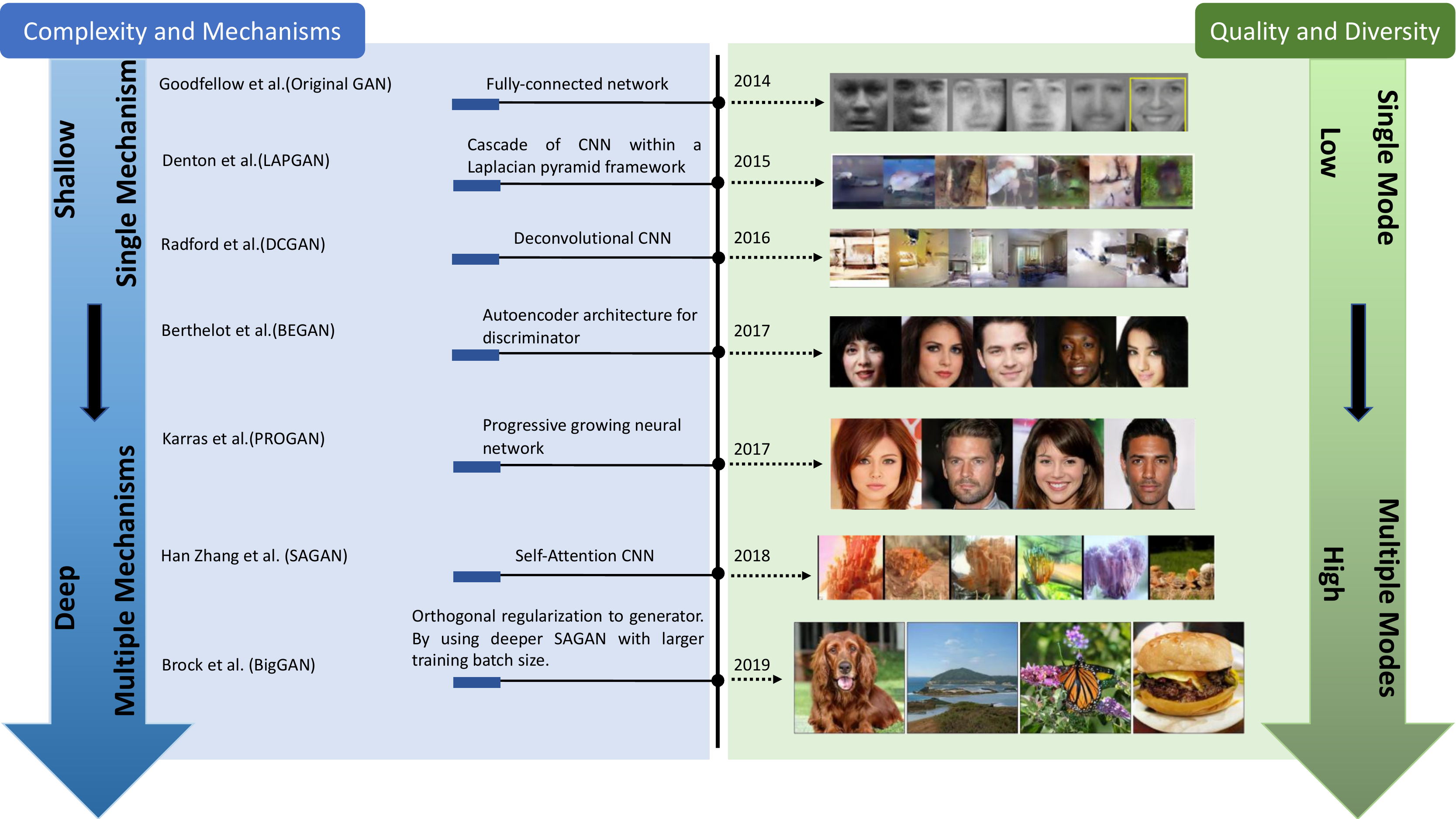}
    \caption{Timeline of part of architecture-variant GANs present in this paper. Complexity in blue stream refers to size of the architecture and computational cost such as batch size. Mechanisms refer to the number of types of models used in the architecture (e.g., BEGAN uses an autoencoder architecture for its discriminator while a deconvolutional neural network is used for the generator. In this case, two mechanisms are used).}
    \label{chap05-fig:GAN-architecture-review}
\end{figure*}

There are many types of architecture-variants proposed in the literature (see Fig.\ref{chap05-fig:GAN-architecture-review})~\cite{radford2015unsupervised,zhang2017stackgan,zhu2017unpaired,berthelot2017began,karras2017progressive}.
Architecture-variant GANs are mainly proposed for the purpose of different applications e.g., image to image transfer~\cite{zhu2017unpaired}, image super resolution~\cite{ledig2017photo}, image completion~\cite{iizuka2017globally}, and text-to-image generation ~\cite{reed2016generative}. In this section, we provide a review on architecture-variants that helps improve the performance for GANs from three aspects mentioned before, namely improving image diversity, improving image quality and more stable training. Review for those architecture-variants for different applications can be referred to work~\cite{hitawala2018comparative,creswell2018generative}.

\subsection{Fully-connected GAN (FCGAN)} 
The original GAN paper~\cite{goodfellow2014generative} uses fully-connected neural networks for both generator and discriminator. This architecture-variant was applied for some simple image datasets i.e., MNIST~\cite{lecun1998gradient}, CIFAR-10~\cite{krizhevsky2009learning} and Toronto face dataset. The authors suggests to do $k$ steps of optimizing $D$ and one step of optimizing $G$ due to overfitting of discriminator if the completion of optimizing $D$ is done in the inner loop of training. In practice, equation~\eqref{eq:GAN-formula} may cause the vanishing gradient issue for optimizing the generator and the authors maximize $\log D(G(z))$ for training $G$. This modification equivalently optimizes Kullback–Leibler (KL) divergence between $p_g$ and $p_r$ for $G$, which also causes the asymmetrical issue and we will revisit this in detail in section~\ref{Loss-variant-GANs}. For the architecture setting, maxout~\cite{glorot2011deep} was deployed for the discriminator while a mixture of ReLU and sigmoid activations were used for the generator. It does not demonstrate good generalization performance for more complex image types.

\subsection{Semi-supervised GAN (SGAN)}
SGAN is proposed in the context of semi-supervised learning~\cite{odena2016semi}. Semi-supervised learning is a promising research field between supervised learning and unsupervised learning. Unlike supervised learning, in which we need a label for every sample, and unsupervised learning, in which no labels are provided, semi-supervised learning has labels for a small subset of example. Compared to FCGAN, SGAN's discriminator is multi-headed i.e., it has softmax and sigmoid for classifying the real data and distinguishing real and fake samples respectively. The authors trained SGAN on the MNIST dataset. Results show that both discriminator and generator in SGAN are improved compared to the original GAN. We think the architecture of the multi-headed discriminator is relatively simple which limits the diversity of the model i.e., the experiment is only carried out on the the MNIST dataset. More complicated architecture for the discriminator may improve the performance for the model.   
\begin{figure*}[!ht]
    \centering
    \includegraphics[width=.8\textwidth]{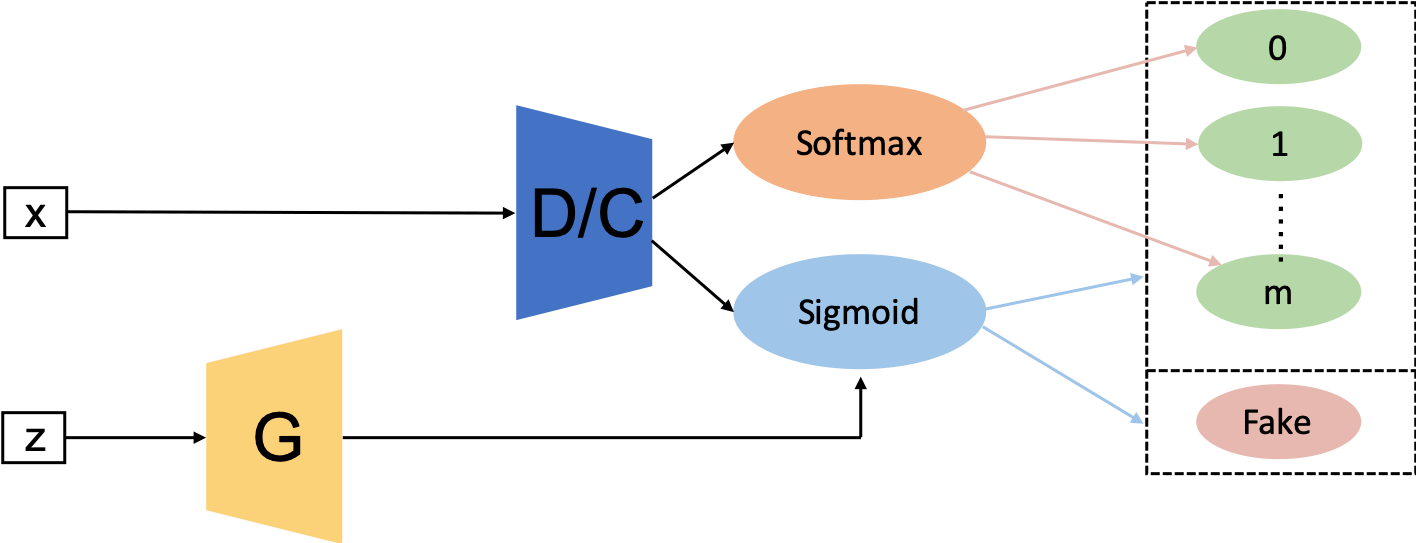}
    \caption{SGAN architecture.}
    \label{fig:SGAN-architecture}
\end{figure*}

\subsection{Bidirectional GAN (BiGAN)}
\begin{figure*}[!ht]
    \centering
    \includegraphics[width=.8\textwidth]{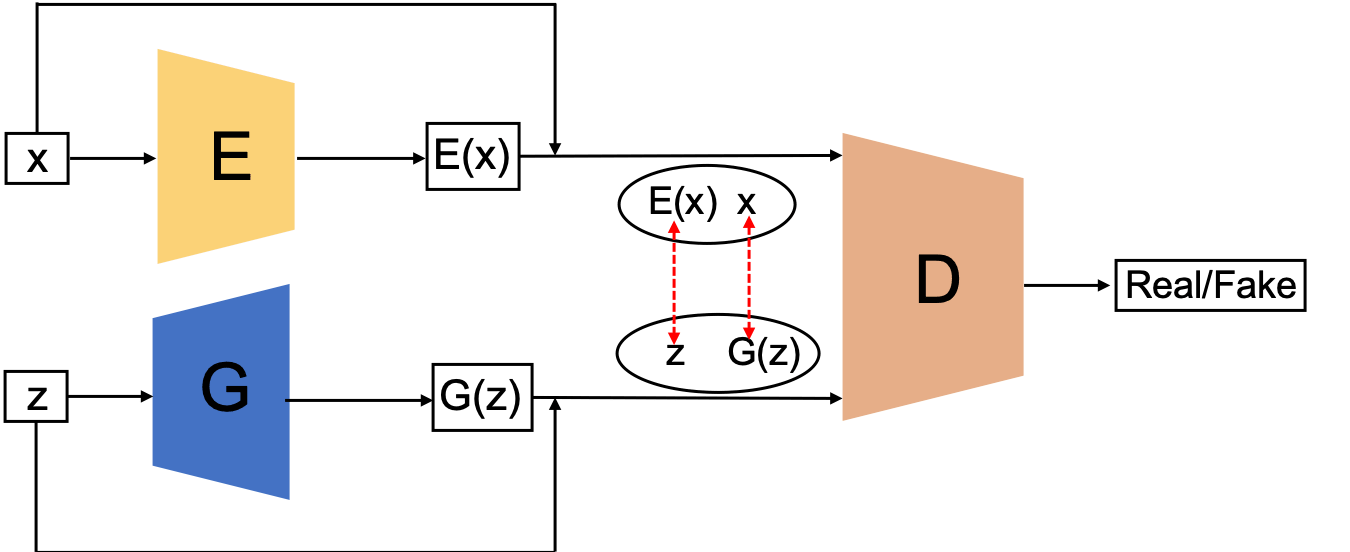}
    \caption{BiGAN architecture.}
    \label{fig:BiGAN-architecture}
\end{figure*}

Traditional GANs have no means of learning the inverse mapping i.e., projecting data back into the latent space. BiGAN is designed for this purpose~\cite{donahue2016adversarial}. As seen in Figure~\ref{fig:BiGAN-architecture}, the overall architecture consists of encoder ($E$), generator ($G$) and discriminator ($D$). $E$ encodes real sample data into $E(\mathbf{x})$ while $G$ decodes $\mathbf{z}$ into $G(\mathbf{z})$. As a result, $D$ aims to evaluate the difference between each pair of $(E(\mathbf{x}), \mathbf{x})$ and $(G(\mathbf{z}), \mathbf{z})$. As $E$ and $G$ do not communicate directly i.e., $E$ never sees $G(\mathbf{z})$ and $G$ never sees $E(\mathbf{x})$. The authors prove that the encoder and decoder must learn to invert one another in order to fool the discriminator in the original paper. It would be interesting to see if such a model is able to deal with adversarial examples in the future work. BiGAN was trained on the MNIST and the ImageNet. Adam optimizer with $\beta_1 = 0.5$ and $\beta_2 = 0.999$ is used for optimization. The batch size is 128 and the weight decay as $2.5e-5$ is applied for all parameters. Batch normalization is also deployed. In the future, it would be interesting to investigate if such a model has some ability to handle the adversarial samples.

\subsection{Conditional GAN (CGAN)} \label{sec:CGAN}
\begin{figure*}[!ht]
    \centering
    \includegraphics[width=.6\textwidth]{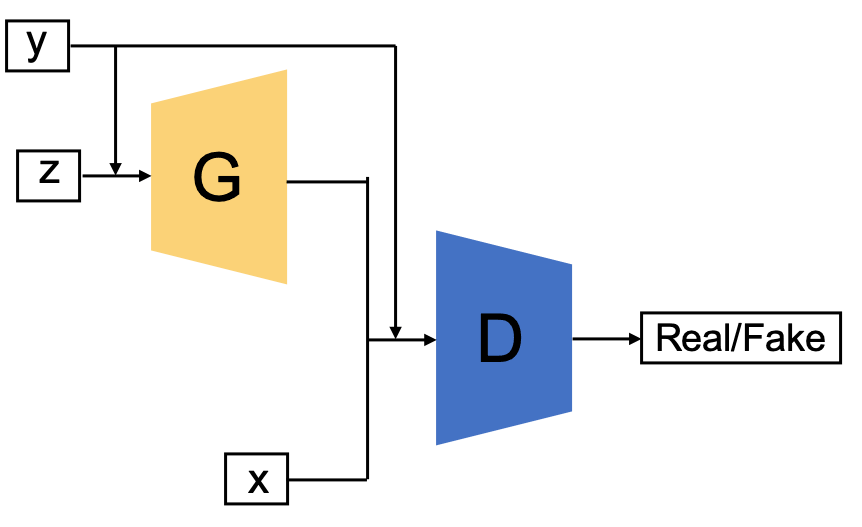}
    \caption{CGAN architecture.}
    \label{fig:CGAN-architecture}
\end{figure*}

CGAN is introduced by conditioning on both discriminator and generator by feeding class label~\cite{mirza2014conditional}. As seen in Figure~\ref{fig:CGAN-architecture}, 
CGAN feeds the extra information $\mathbf{y}$ ($\mathbf{y}$ can be class label or other modal data) to both discriminator and generator. It should be noted that $\mathbf{y}$ is normally encoded inside the generator and discriminator before being concatenated with the encoded $\mathbf{z}$ and encoded $\mathbf{x}$. For example, the MNIST experiment in the original work, both $\mathbf{z}$ and $\mathbf{y}$ are mapped to hidden layers with layer sizes 200 and 1000 respectively before being combined with each other (combined layer dimensionality is $200+1000=1200$) in the generator. By doing this, CGAN enhances the discriminative ability for the discriminator. The loss function of CGAN is slightly different from the FCGAN as seen in equation~\eqref{eq:CGAN-formula}, in which $\mathbf{x}$ and $\mathbf{y}$ are conditioned by $\mathbf{z}$. Benefiting from the extra encoded $y$ information, CGAN is not only able to handle with unimodal image datasets but also multimodal datasets such as Flickr that contains labeled image data with their associated user-generated metadata (UGM) i.e., in particular user-tags, which brings GANs over to the area of multimodal data generation. The authors experiment the CGAN on the MNIST and Yahoo Flickr Creative Common 100M (YFCC 100M). For the MNIST, the model was trained using SGD with mini-batch size of 100 and initial learning rate of 0.1 which was exponentially decreased down to $1e-6$ with the decay factor of 1.00004. Dropout was utilized with probability of 0.5 to both generator and discriminator. The momentum was used with initial value of 0.5 and finally was increased up to 0.7. Class labels were encoded as one-hot vectors and fed to both $G$ and $D$. In terms of the YFCC 100M experiment, training hyperparameters are the same as the set-up in the MNIST experiment. Even CGAN enhances the discriminative ability of the discriminator due to introducing encoded labels, some of the generated labels still lose connection with images.  
\begin{equation} \label{eq:CGAN-formula}
\begin{aligned}
	\min \limits_{G} \max\limits_{D}\hspace{2pt}&\mathbb{E}_{\bm{\mathrm{x}} \sim p_{r}} \mathrm{log}[D(\bm{\mathrm{x}}|\mathbf{y})] + \mathbb{E}_{\bm{\mathrm{z}} \sim p_{\bm{\mathrm{z}}}} \mathrm{log}\left[1 - D(G(\bm{\mathrm{z}}|\mathbf{y}))\right].
\end{aligned}
\end{equation}

\subsection{InfoGAN}
\begin{figure*}[!ht]
    \centering
    \includegraphics[width=.6\textwidth]{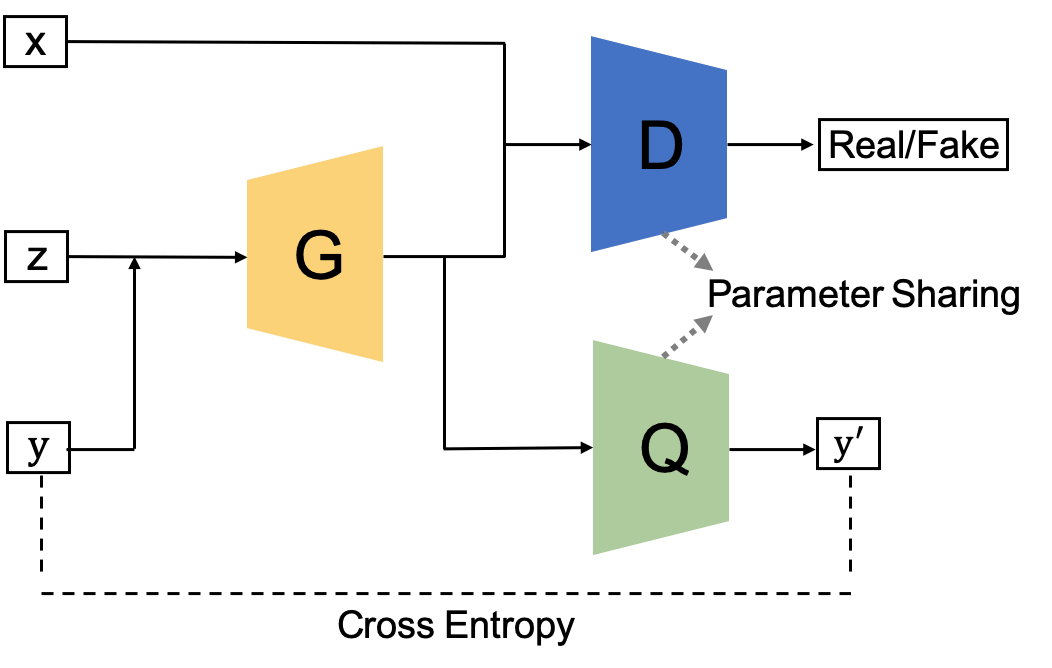}
    \caption{InfoGAN architecture.}
    \label{fig:InfoGAN-architecture}
\end{figure*}

InfoGAN is proposed beyond the CGAN~\cite{chen2016infogan}, which learns the interpretable representations unsupervisedly by maximizing the mutual information between conditional variables and the generative data. In order to achieve this, InfoGAN introduces another classifier $Q$ (see Figure~\ref{fig:InfoGAN-architecture}) 
to predict the $\mathbf{y}$ given by $G(\mathbf{z}|\mathbf{y})$. The combination of $G$ and $Q$ here can be understood as an autoencoder, in which we aim to find the embedding ($G(\mathbf{z}|\mathbf{y})$) minimizing the cross entropy between $\mathbf{y}$ and $\mathbf{y}^{'}$. On the other hand, $D$ does the job as the same as the FCGAN, which distinguishes samples generated from $G$ or from real data. In order to save the computational cost, $Q$ and $D$ share all convolutional layers except the last fully connected layer, which enables the discriminator have the capability of distinguishing real and fake samples and recover the information $\mathbf{y}$. This can improve the discriminative ability for InfoGAN compared to the original architecture. The loss of InfoGAN is a regularization of CGAN's loss
\begin{equation} \label{eq:InfoGAN-formula}
\begin{aligned}
	\min \limits_{G} \max\limits_{D}\hspace{2pt}V(D,G)-\lambda I(G,Q),\hspace{5pt} \lambda > 0
\end{aligned}
\end{equation}
where $V(D,G)$ is the objective of CGAN except that the discriminator does not take $\mathbf{y}$ as input and $I(\cdot)$ is the mutual information. InfoGAN MNIST, 3D face images~\cite{paysan20093d}, 3D chair images~\cite{aubry2014seeing}, SVHN and CelebA. All datasets share the same training setting, in which Adam optimizer is used and batch normalization is applied. Leaky ReLU with 0.1 leaky rate is applied to discriminator and the ReLU is used for the generator. Learning rate $2e-4$ is set for $D$ while $1e-3$ is set for $G$. $\lambda$ is set as 1. Here we think the diversity of the model is very limited due to the parameter of $D$ and $Q$ are shared with each other except the last layer. More complicated set-up $Q$ can be investigated.

\subsection{Auxiliary Classifier GAN (AC-GAN)} \label{sec:AC-GAN}
\begin{figure*}[!ht]
    \centering
    \includegraphics[width=.6\textwidth]{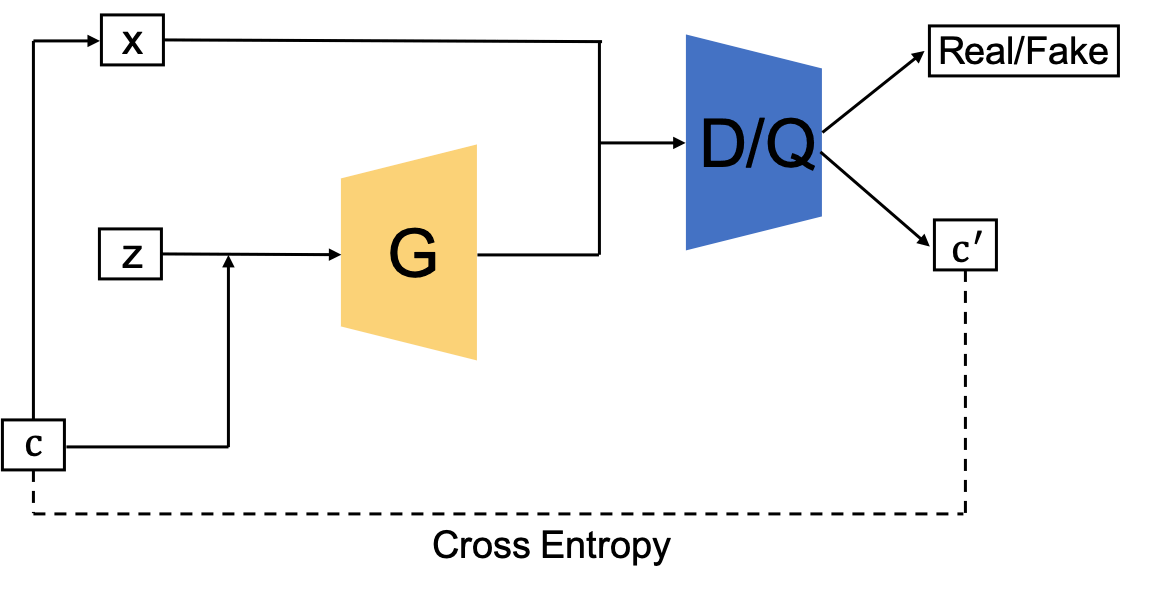}
    \caption{AC-GAN architecture. $\mathbf{x}$ is the real data from class $c$, $\mathbf{z}$ is the input noise and $c^{'}$ is the output of auxiliary classifier.}
    \label{fig:AC-GAN-architecture}
\end{figure*}

AC-GAN~\cite{odena2017conditional} is very similar to CGAN and InfoGAN, which contains an auxiliary classifier in the architecture as seen in Figure~\ref{fig:AC-GAN-architecture}. 
In AC-GAN, each generated sample has a corresponding class label $c$ in addition to $\mathbf{z}$. It should be noted that the difference between AC-GAN and previous two architecture-variants (CGAN and InfoGAN) is the additional information here, which only refers to the class label while the previous two can be other domain data. Thus we use $c$ and $c^{'}$ in AC-GAN in order to be separated from previous two variants. The discriminator in AC-GAN consists of a discriminator $D$ (distinguishes real and fake samples) and a classifier $Q$ (classifies real and fake samples). Similar to InfoGAN, the discriminator and classifier share all weights except the last layer. The loss function of AC-GAN can be constructed by considering the discriminator and classifier, which can be stated as
\begin{equation} \label{eq:AC-GAN-formula}
\begin{aligned}
	&L_S = \mathbb{E}_{\bm{\mathrm{x}} \sim p_{r}} \mathrm{log}[D(\bm{\mathrm{x}}|\mathbf{c})] + \mathbb{E}_{\bm{\mathrm{z}} \sim p_{\bm{\mathrm{z}}}} \mathrm{log}\left[1 - D(G(\bm{\mathrm{z}}|\mathbf{c}))\right] \\
	&L_C = \mathbb{E}_{\bm{\mathrm{x}} \sim p_{r}} \mathrm{log}[Q(\bm{\mathrm{x}}|\mathbf{c})] + \mathbb{E}_{\bm{\mathrm{z}} \sim p_{\bm{\mathrm{z}}}} \mathrm{log}\left[Q(G(\bm{\mathrm{z}}|\mathbf{c}))\right]
\end{aligned}
\end{equation}
where $D$ is trained by maximizing $L_S+L_C$ and $G$ is trained on maximizing $L_C-L_S$. The authors trained AC-GAN on the CIFAR-10 and the ImageNet for all 1000 classes. For both CIFAR-10 and ImageNet, the model was trained by using Adam with $alpha=2e-4$, $\beta_1=0.5$ and $\beta_2=0.999$ for $D$, $G$ and $Q$. Mini-batch size was set to 100. Details of model performance and experiment can be referred to the original paper~\cite{odena2017conditional}. AC-GAN has improved visual quality for the generated images and has high model diversity. However, these improvements depend on large-scale labeled dataset, which may pose some challenges in some real-world application. Combination between AC-GAN and unsupervised or self-supervised manner can be further investigated. We have also introduced a type of GAN, label-noise Robust GANs (rGANs) in section~\ref{sec:rGANs}, which deals with the noisy label issue.

\subsection{Laplacian Pyramid of Adversarial Networks (LAPGAN)}
\begin{figure*}[!ht]
    \centering
    \includegraphics[width=.9\textwidth]{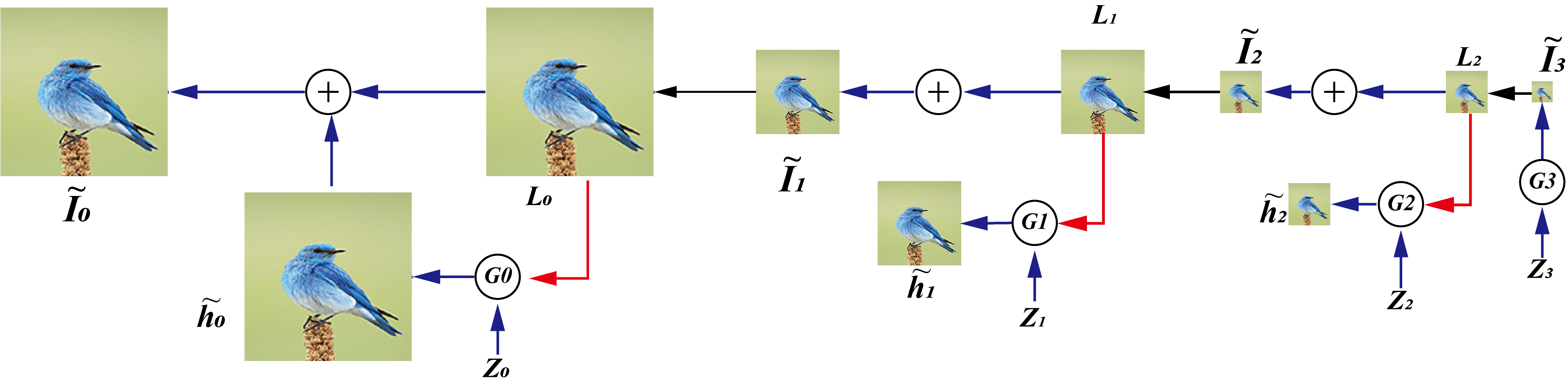}
    \caption{Up-sampling process of generator in LAPGAN (from right to left). The up-sampling process is marked using green arrow and a conditioning process via a conditional GAN (CGAN)~\cite{mirza2014conditional} is marked using the orange arrow. The process initially uses $G_3$ to generate image $\widetilde{I}_3$ and then up-samples the image $\widetilde{I}_3$ to $l_2$. Together with another noise $z_2$, $G_2$ generates a difference image $\widetilde{h}_2$ and adds $\widetilde{h}_2$ to $l_2$ which produces the generated image $\widetilde{I}_2$. The rest can be done in the same manner. LAPGAN contains 3 generators in this work in order to up-sample the image. Figure regenerated from~\cite{denton2015deep}.}
    \label{fig:LAPGAN-architecture}
\end{figure*}

LAPGAN is proposed for the production of higher resolution images from lower resolution input GAN~\cite{denton2015deep}. The Laplacian pyramid~\cite{burt1983laplacian} is an image coding approach, which uses local operators of many scales but identical shape as the basic functions. LAPGAN utilizes a cascade of CNNs within a Laplacian pyramid framework~\cite{burt1983laplacian} to produce high quality images, which can demonstrated in Figure~\ref{fig:LAPGAN-architecture} (from right to left).
Rather than using deconvolutional process (i.e., used in DCGAN) to up-sample the kernel output of the previous layer, LAPGAN uses Laplacian pyramid to up-sample the image. First, LAPGAN use the first generator to produce a very small image, which can alleviate the unstable issue for the generator, and this image is then up-sampled through using Laplacian pyramid. Then the up-sampled image is fed to the next generator for producing the image difference and the summation of the image difference and input image will be the generated image. It can be seen that only the $G_3$ in Figure~\ref{fig:LAPGAN-architecture} is used for generating images but the dimension is very small, which benefits the stable training. For the larger pixel images, the generator is used to generate the image difference, which is much less complicated that the same size raw images. This structure benefits more stable training and high resolution modeling. CIFAR10 ($28 \times 28$ pixel), STL ($96 \times 96$ pixel) and LSUN ($64 \times 64$ pixel) were used for generation. The Laplacian pyramid up-sampling processes for each dataset are $8 \to 14 \to 28$ (CIFAR10), $8 \to 16 \to 32 \to 64 \to 96$ (STL) and $4 \to 8 \to 16 \to 32 \to 64$ (LSUN). The discriminator used 3 hidden layers and a sigmoid output while the generator used 5-layer convnet with ReLU and batch normalization and a linear ouput layer. SGD with an initial learning rate of 0.02, decreased by a factor of $(4e-5)$ at every epoch, was deployed in the experiment. Momentum started at 0.5, increasing by 0.0008 at epoch up to a maximum of 0.8. Current structure includes multiple generators for generating images and the connections between these generators have not been established. In section~\ref{sec:PROGAN}, we introduce a more advanced strategy, which trains the model in a progressive fashion i.e., PROGAN.

\subsection{Deep Convolutional GAN (DCGAN)}
DCGAN is the first work that applied a deconvolutional neural networks architecture for $G$~\cite{radford2015unsupervised}. Deconvolution is proposed to visualize the features for a CNN and has shown good performance for CNNs visualization~\cite{zeiler2014visualizing}. DCGAN deploys the spatial up-sampling ability of the deconvolution operation for $G$, which enables the generation of higher resolution images using GANs. There are some critical modifications in the architecture of DCGAN compared to original FCGAN, which benefits high resolution modeling and more stable training. Firstly, DCGAN replaces any pooling layers with strided convolutions for discriminator and fractional-strided convolutions for generator. Secondly, batch normalization is used for both the discriminator and the generator, which helps locate the generated samples and the real samples centering at zero i.e., \textbf{similar statistics for generated samples and real samples}. Thirdly, ReLU activation is used in generator for all layers except output, which uses Tanh, while LeakyReLU activation is used in the discriminator for all layers. In this case, the LeakyReLU activation will prevent the network stucking in a “dying state” situation (e.g., inputs smaller than 0 in ReLU) as the generator receives gradients from the discriminator. DCGANs are trained on Large-scale Scene Understanding (LSUN)~\cite{yu2015lsun}, ImageNet~\cite{deng2009imagenet} and the customized-assembled face dataset. All models were trained using stochastic gradient descent (SGD) with a mini-batch size of 128. All weights were initialized from a zero-centered Normal distribution with standard deviation 0.02. Adam optimizer was utilized with learning rate of 0.0002 and momentum term of 0.5. The slope of LeakyReLU was set to 0.2 for all models. Models were trained by using $64 \times 64$ pixel image. DCGAN is a very important milestone in the GANs history and the deconvolution becomes the main architecture used in the generator. Due to the limit of the model capacity and the optimization used in DCGAN, it is only successful on low-resolution and less diverse images.

\subsection{Boundary Equilibrium GAN (BEGAN)}
\begin{figure}[!ht]
    \centering
    \includegraphics[width=.7\textwidth]{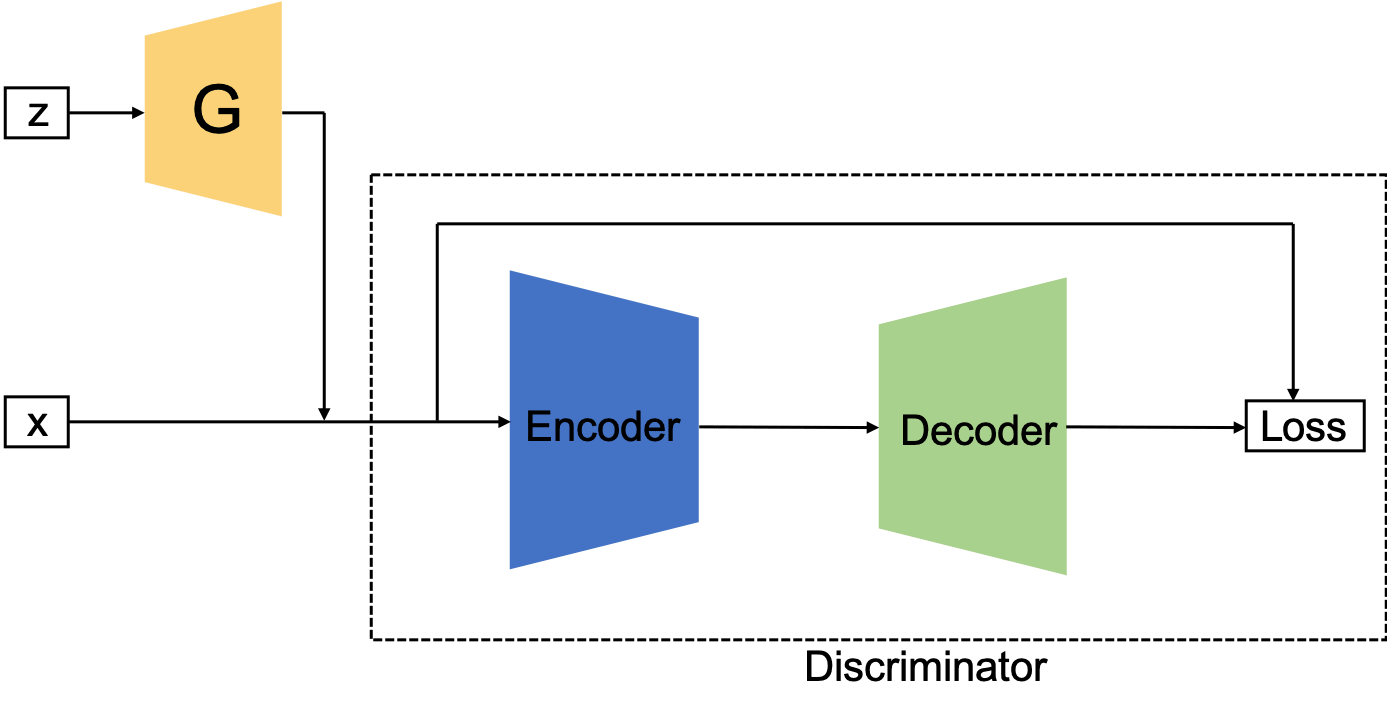}
    \caption{Illustration of BEGAN architecture. $\mathbf{z}$ is the latent variable for $G$ and $\mathbf{x}$ is input image. BEGAN deploys an autoencoder architecture for the discriminator. Loss is calculated using $L_1$ or $L_2$ norm at pixel level.}
    \label{chap05-fig:BEGAN-architecture}
\end{figure}

BEGAN uses an autoencoder architecture for the discriminator which was first proposed in EBGAN~\cite{zhao2016energy} (see Fig.~\ref{chap05-fig:BEGAN-architecture}). 
As seen in Fig.~\ref{chap05-fig:BEGAN-architecture}, the autoencoder loss can be generated for $G$ and $D$ respectively. When training the autoencoder ($D$), the objective is to maximize the reconstruction loss of real images and maximizes the reconstruction loss for generated images i.e., minimize $\mathbb{E}\left[\mathcal{L}\right(x)]-\mathbb{E}\left[\mathcal{L}\right(G(z))]$. When training the $G$, the objective is to minimize $\mathbb{E}\left[\mathcal{L}\right(G(z))]$. By introducing the autoencoder, the authors have proved the optimization of the reconstruction loss above is equivalent to the Wasserstein distance. The authors also propose the use of a hyperparameter $\gamma = \frac{\mathbb{E}\left[\mathcal{L}\right(G(z))]}{\mathbb{E}\left[\mathcal{L}\right(x)]}, \gamma \in \left[0, 1\right]$ to control the balance between the generator and discriminant losses, which allows to balance the effort allocated to the generator and the discriminator i.e., control the variety of generated faces. The experiment in the original paper shows that smaller $\gamma$ makes $G$ generate faces that look overly uniform. Variety of faces increases with a larger value of $\gamma$ but also introduces artifacts. The overall loss function is summarized in equation~\eqref{began-eq}
\begin{equation} \label{began-eq}
\begin{cases}
    \mathcal{L}_D = \mathcal{L}(x) - k_t\mathcal{L}(G(z_D)), & \text{for updating $\theta_D$}  \\
    \mathcal{L}_G = \mathcal{L}(G_{z_G}), & \text{for updating $\theta_G$}\\ 
    k_{t+1} = k_t + \lambda_k(\gamma \mathcal{L}(x)-\mathcal{L}(G(z_G)),& \text{for each training iteration t} \\
\end{cases}
\end{equation}
where $\mathcal{L}(\cdot)$ represents the autoencoder reconstruction loss ($L_2$), $k_t \in [0, 1]$ is a variable that controls how much emphasis of $\mathcal{L}(G(z))$ is penalized for the loss. $k$ is initialized as 0 and is controlled by $\lambda_k$ ($\lambda_k$ can be interpreted as learning rate for $k$, which is set as $1e-3$ in the original paper).

Compared to traditional optimization, the BEGAN matches the autoencoder loss distributions using a loss derived from the Wasserstein distance instead of matching data distributions directly. This modification helps $G$ to generate easy-to-reconstruct data for the autoencoder at the beginning because the generated data is close to 0 and the real data distribution has not been learned accurately yet, which prevents $D$ easily winning $G$ at the early training stage. For encoder and decoder, exponetial linear units (ELUs) were applied at their outputs. The model was trained on CelebA dataset using $128 \times 128$ pixel images. Separate Adam optimizers with initial learning rate of $1e-4$, decaying by a factor of 2 when the measure of convergence stalls, were used for $D$ and $G$. Batch size is set to 16 in the original work. 

\subsection{Progressive GAN (PROGAN)}\label{sec:PROGAN} 
\begin{figure}[!ht]
    \centering
    \includegraphics[width=.8\textwidth]{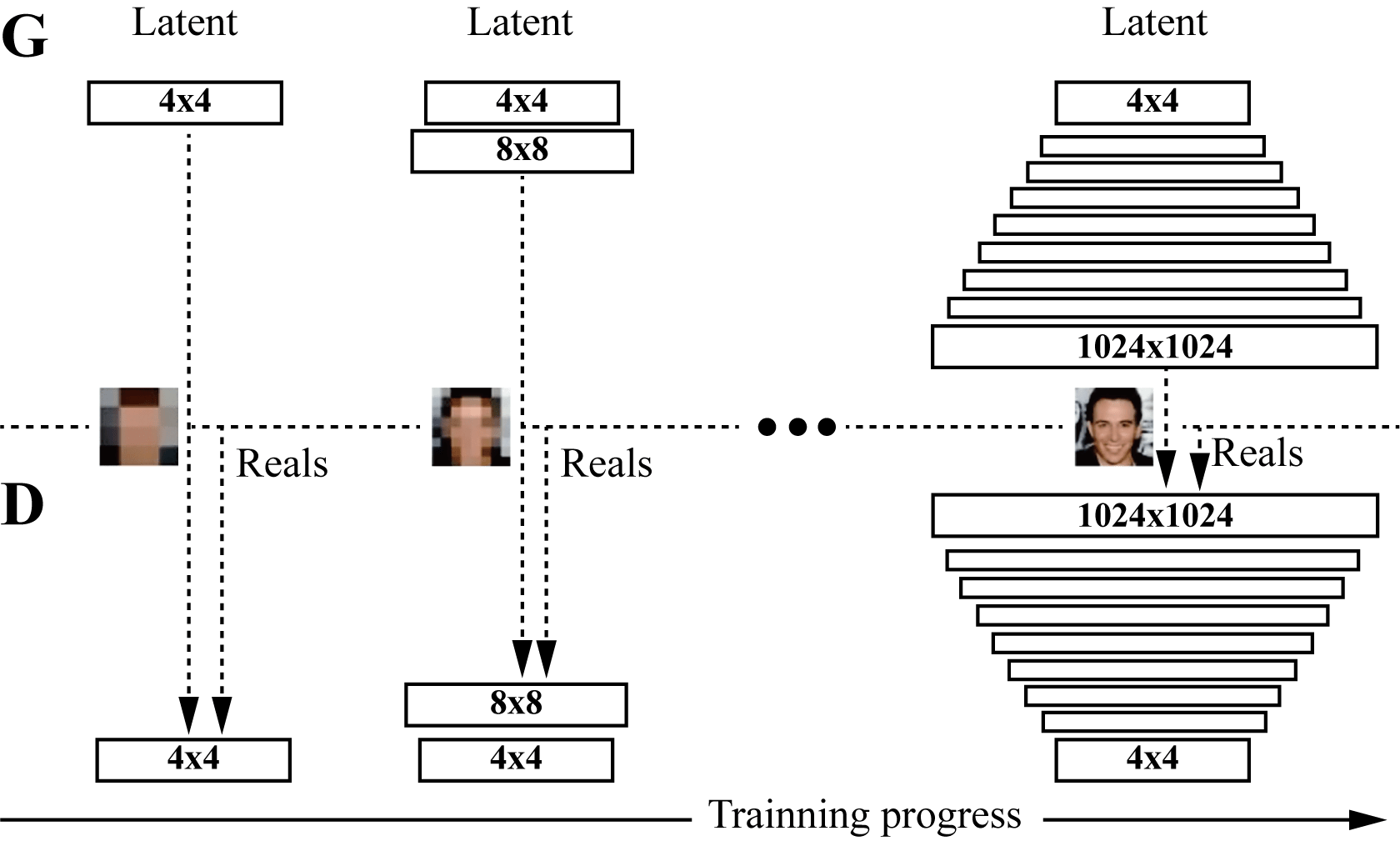}
    \caption{Progressive growing step for PROGAN during the training process. Training starts with $4\times 4$ pixels image resolution. With the training step growing, layers are incrementally added to $G$ and $D$ which increases the resolution for the generated images. All existing layers are trainable throughout the training stage. Figure regenerated from~\cite{karras2017progressive}.}
    \label{chap05-fig:PROGAN-architecture}
\end{figure}

PROGAN involves progressive steps toward the expansion of the network architecture~\cite{karras2017progressive}. This architecture uses the idea of progressive neural networks first proposed in~\cite{rusu2016progressive}. This technology does not suffer from forgetting and is able to deploy prior knowledge via lateral connections to previously learned features. Consequently it is widely applied for learning complex task sequences. Figure.~\ref{chap05-fig:PROGAN-architecture} 
demonstrates the training process for PROGAN. Training starts with low resolution $4\times 4$ pixels image. Both $G$ and $D$ start to grow with the training progressing. Importantly, all variables remain trainable throughout this growing process. This progressive training strategy enables substantially more stable learning for both networks. By increasing the resolution little by little, the networks are continuously asked a much simpler question comparing to the end goal of discovering a mapping from latent vectors. All current state-of-the-art GANs employ this type of training strategy and it has resulted in impressive, plausible images~\cite{karras2017progressive,brock2018large,karras2018style}. The authors start training the PROGAN with $4 \times 4$ pixel images and incrementally add the doubled-sized layers to $G$ and $D$ as seen in Fig.~\ref{chap05-fig:PROGAN-architecture}, in which the new layers are faded smoothly. The multi-scaled training images are produced by using Laplacian pyramid representations~\cite{burt1983laplacian} i.e., similar to LAPGAN. Models were trained on CIFAR10 ($32 \times 32$ pixel images), LSUN ($256 \times 256$ pixel images), and CelebA-HQ ($1024 \times 1024$ pixel images). Leaky ReLU with leakness 0.2 were used for all layers of both $D$ and $G$ excepth last layer (use linear activation). Only pixelwise normalization of the feature vectors after each Conv $3 \times 3$ layer in the generator was deployed i.e., no batch normalization, layer normalization, or weight normalization in either network. The Adam optimizer with $\alpha = 1e-3, \beta_1 = 0, \beta_2 = 0.99$ and $\epsilon = 1e-8$ was utilized for training $D$ and $G$. Mini-batch size was gradually decreased with increasing image pixel for saving the memory budget i.e., batch size 16 for $4\times 4$ to $8\times 8$, $256\times 256 \to 14$, $512\times 512 \to 6$ and $1024\times 1024 \to 3$. The WGAN-GP~\cite{wu2017gp} loss was used for optimizing both $D$ and $G$.

\subsection{Self-attention GAN (SAGAN)}\label{sec:SAGAN}
\begin{figure}[!ht]
    \centering
    \includegraphics[width=.9\textwidth]{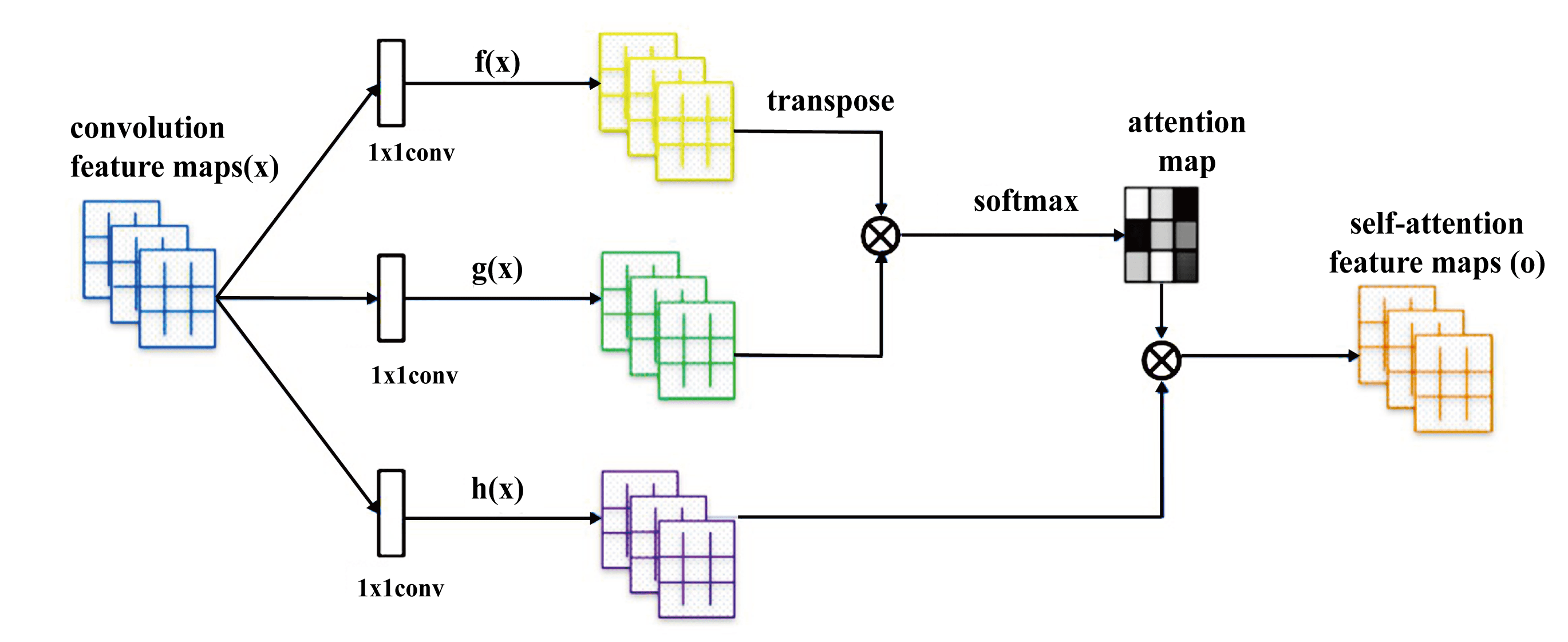}
    \caption{Self-attention mechanism architecture proposed in the paper. $f$, $g$ and $h$ are correspondence with query, key and value in the self-attention mechanism. The attention map indicates the long-range spatial dependencies.
    The $\otimes$ is matrix multiplication. Figure regenerated from~\cite{zhang2018self}.}
    \label{chap05-fig:SAGAN-architecture}
\end{figure}

Traditional CNNs can only capture local spatial information and the receptive field may not cover enough structure, which causes CNN-based GANs to have difficulty in learning multi-class image datasets (e.g., ImageNet) and the key components in generated images may shift e.g., the nose in a face-generated image may not appear in right position. Self-attention mechanism have been proposed to ensure large receptive field and without sacrificing computational efficiency for CNNs~\cite{vaswani2017attention}. SAGAN deploys a self-attention mechanism in the design of the discriminator and generator architectures for GANs~\cite{zhang2018self} (see Fig.~\ref{chap05-fig:SAGAN-architecture}). 
Benefiting from the self-attention mechanism, SAGAN is able to learn global, long-range dependencies for generating images. It has achieved great performance on multi-class image generation based on the ImageNet datasets. The authors trained SAGAN on the ImageNet dataset ($128\times 128$ pixel images). The spectral normalization~\cite{miyato2018spectral} was applied for both $D$ and $G$. Conditional batch normalization was used in the generator while batch projection was used in the discriminator. Adam optimizer with $\beta_1 = 0$ and $\beta_2 = 0.9$ and the learning rate for discriminator was $4e-4$ and for generator was $1e-4$. The authors also demonstrate that the deployment of self-attention mechanism for both the discriminator and the generator at large feature maps is more effective i.e., deployment self-attention mechanism at feature map with $32\times 32$ size achieves the best performance using FID score and deployment self-attention mechanism at feature map with $64\times 64$ size achieves the best performance using Inception score, which indicates the self-attention mechanism is complementary to convolution for large feature maps. Thus the self-attention mechanism is suggested to be applied for large feature maps in order to improve the diversity for GANs.

\subsection{BigGAN}\label{sec:BigGAN}
BigGAN~\cite{brock2018large} has also achieved state-of-the-art performance on the ImageNet datasets. Its design is based on SAGAN and it has demonstrated that the performance can be benefited by scaling up GAN training i.e., increase the number of channels for each layer and increase the batch size. The authors train the model on ImageNet with $128\times 128$, $256\times 256$ and $512\times 512$ resolutions. The training setting in this work follows the SAGAN, in which the learning rate was halved and train two $D$ steps per $G$ step. Different selections of latent variables $z$ are explored and the authors state that Bernoulli $\{0,1\}$ and Censored Normal $\text{max}(\mathcal{N}(0, I), 0)$ work best without truncation. The truncation trick involves using a different distribution for the generator's latent space during training than during inference or image synthesis. In BigGAN, a Gaussian distribution is used during training, and a truncated Gaussian is used during inference. This truncation trick provides a trade-off between image quality or fidelity and image variety. A more narrow sampling range results in better quality, whereas a larger sampling range results in more variety in sampled images. We summarize following operations on BigGAN that make BigGAN scale-up the architecture: (1) \textbf{Self-attention module} and \textbf{Hinge loss}: the BigGAN uses the model architecture with attention modules from SAGAN and is trained via hinge loss, in which self-attention contributes to the model diversity and hinge loss enables more stable training; (2) \textbf{Class conditional information}: the class information is provided to the generator model via class-conditional batch normalization; (3) \textbf{Update discriminator more than generator}: the BigGAN slightly modifies this and updates the discriminator model twice before updating the generator model in each training iteration; (4) \textbf{Moving average of model weights}: before images are generated for evaluation, the model weights are averaged across prior training iterations using a moving average; (5) Some operations on the network: \textbf{orthogonal weight initialization}, \textbf{larger batch size}, \textbf{skip-z connections} (skip connections from the latent to multiple layers), and \textbf{shared embeddings} i.e., the authors show these operations are all able to help improve the performance. The authors also characterize the analysis of instability specific to such large scale. More details can be referred to the original paper.

\subsection{Label-noise Robust GANs (rGANs)} \label{sec:rGANs}
\begin{figure}[!ht]
    \centering
    \includegraphics[width=1.\textwidth]{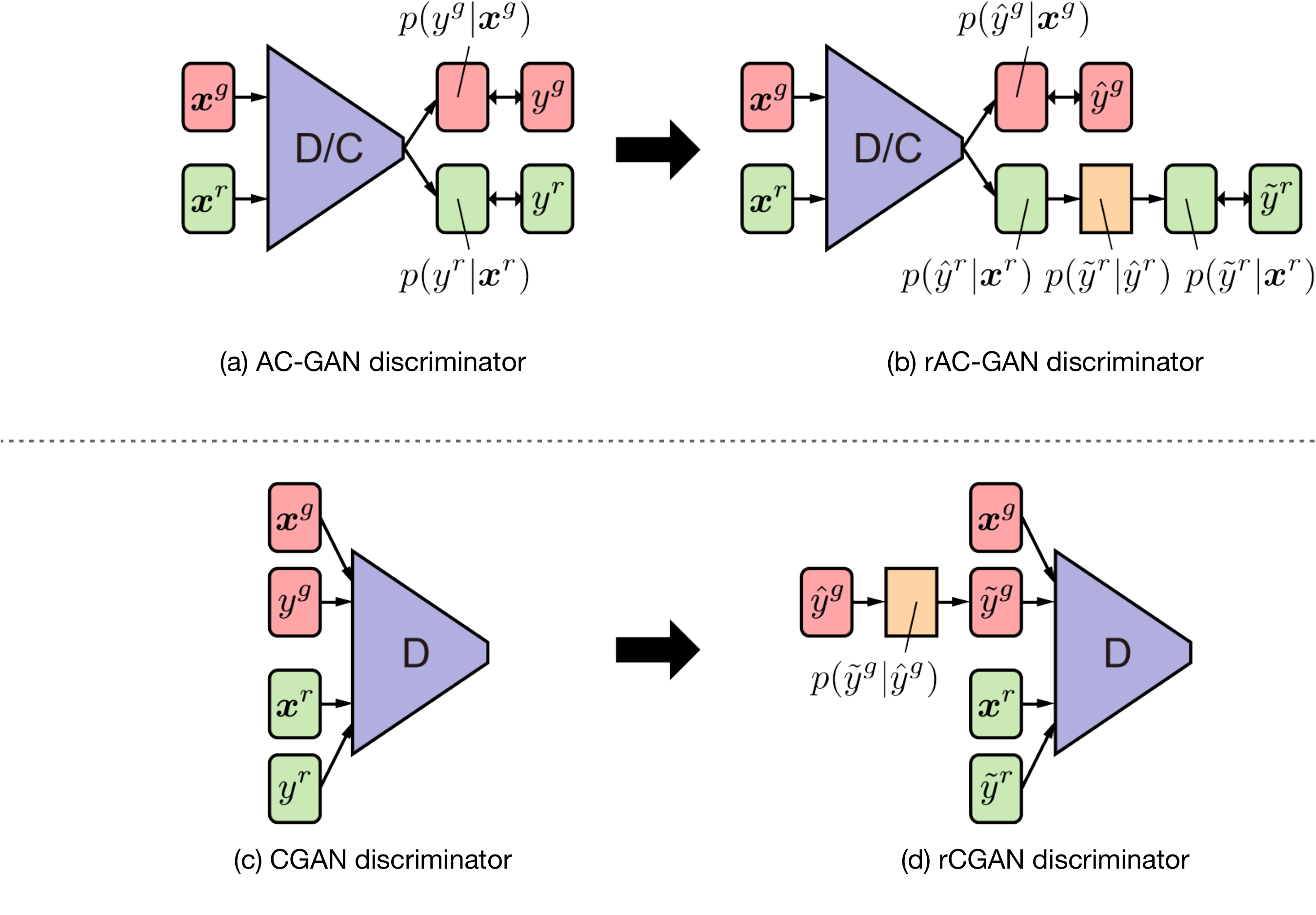}
    \caption{Discriminators of (a) AC-GAN, (b) rAC-GAN, (c) CGAN and (d) rCGAN. A noise
transition model is denoted as an orange rectangle. Generators in rAC-GAN and rCGAN remain the same as in AC-GAN and CGAN. Figure regenerated and reorganized from~\cite{kaneko2019label}.}
    \label{fig:rGANs}
\end{figure}

We have discussed CGAN in section~\ref{sec:CGAN} and AC-GAN in section~\ref{sec:AC-GAN} respectively, which have ability to learn the disentangled representation and improve the discriminative ability for GANs. However, large-scale labeled datasets are required for training models, which poses some challenges in real-world scenario. Kaneko \textit{et al.}~\cite{kaneko2019label} propose a family of GANs named label-noise robust GANs (rGANs), which incorporates a noise transition model that is able to learn a clean label conditional generative distribution even when provided training labels are noisy. Two variants are discussed, which are an extension for AC-GAN (rAC-GAN) and an extension for CGAN (rCGAN) as seen in Figure~\ref{fig:rGANs}. 
The core part of rGANs is a noise transition module $p(\tilde{y}|\hat{y})$ ($\tilde{y}$ is the noisy label and $\hat{y}$ is the clean label) introduced to the discriminator, in which $p(\tilde{y}=j|\hat{y}=i)$ = $T_{(i,j)}$ as $T$ is a noise transition matrix $T_{(i,j)} \in [0, 1]^{(c\times c)}$ ($\sum_i T_{i,j}=1$, $c$ is the number of classes). The authors trained rGANs on CIFAR-10 and CIFAR-100. The authors demonstrate that rAC-GAN and rCGAN perform better than original architectures in CIFAR-10 and also show the robustness to label noise. However, in CIFAR-100, when high noise introduced to labels, their performance drops. We think such a framework is still somewhat limited when encountering more complicated datasets e.g., ImageNet and it needs to be investigated in the future.

\subsection{Your Local GAN (YLG)}\label{sec:YLG}
This work~\cite{Daras_2020_CVPR} introduces a new local sparse attention layer that preserves the two-dimensional geometry and locality. To show the applicability of the idea, they replace the dense attention layer of SAGAN~\cite{vaswani2017attention} with new construction. The key innovations are 1) the attention patterns well supported by information theoretic framework of Information Flow Graphs;
2) YLG-SAGAN has been introduced and achieves superior performance with reducing the training time by approximately $40\%$; 3) they made the natural inversion process of performing gradient descent on the loss work on bigger models rather than previous works on small GANs. One specific trick the author utilizes is called ESA (Enumerate, Shift, Apply). They modify one dimensional
sparsifications to become aware of two-dimensional locality via \textbf{enumerating} pixels of the
image based on their Manhattan distance from the pixel at
location (0, 0) (breaking ties using row priority), \textbf{shifting} the indices of any given one-dimensional sparsification to match the Manhattan distance enumeration instead of the reshape enumeration, and \textbf{applying} this new one dimensional sparsification pattern, that respects two-dimensional locality, to the one-dimensional reshaped version of the image.

However, we think two conflicting objectives exist in this work. On one hand, this method intends to make the networks as sparse as possible for computational and statistical efficiency, on the other hand they still need to support good and full information flow. 

\subsection{AutoGAN}\label{sec:autoGAN}
\begin{figure}[!htbp]
    \centering
    \includegraphics[width=.9\textwidth]{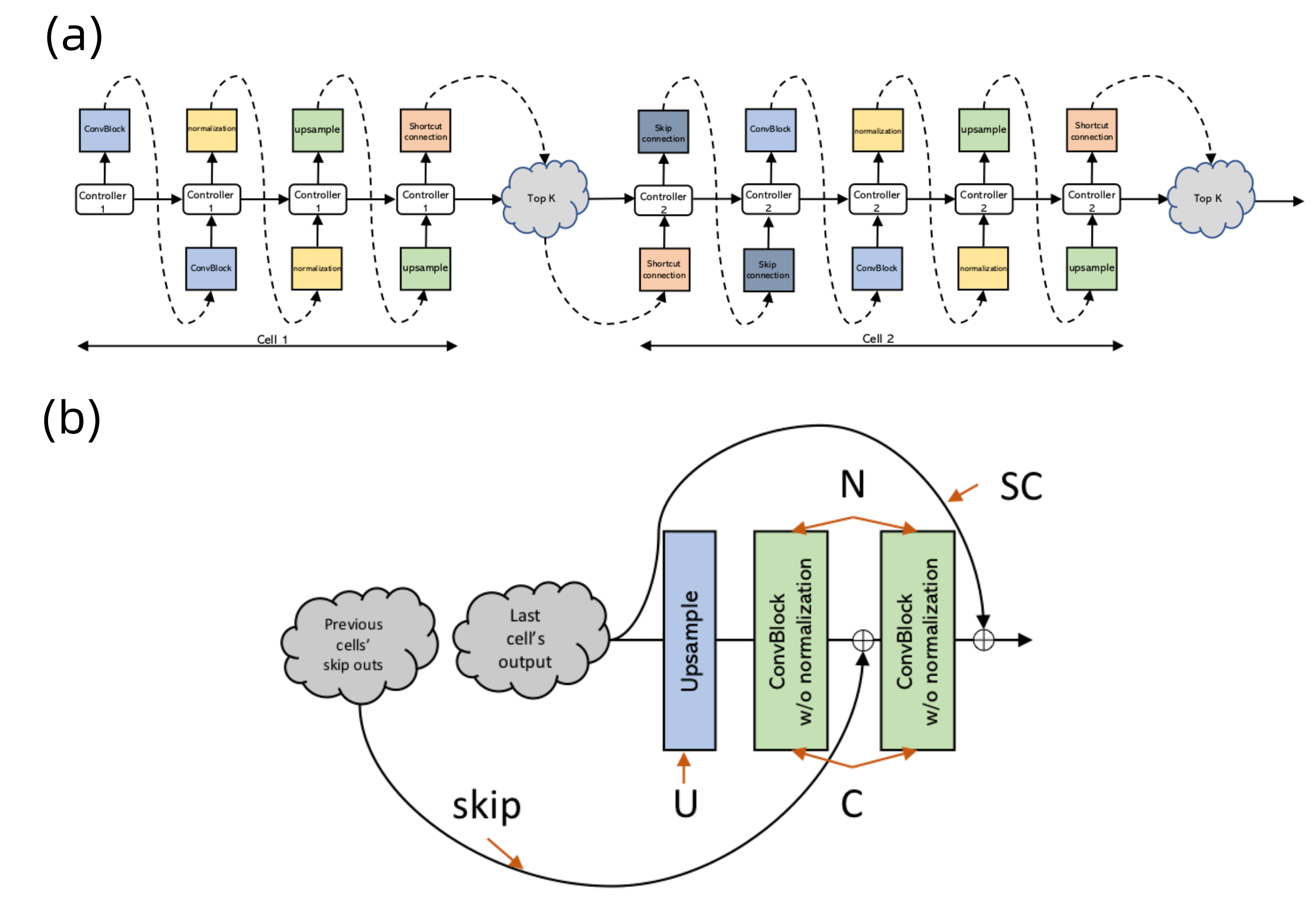}
    \caption{(a) The running scheme of the RNN controller. At each time step, the controller outputs a hidden vector to be decoded into an operation, with its corresponding softmax classifier. (b) The search space of a generator cell in AutoGAN. Figure regenerated and reorganized from~\cite{Gong_2019_ICCV}.}
    \label{chap05-fig:AutoGAN-architecture}
\end{figure}

AutoGAN~\cite{Gong_2019_ICCV} studies on introducing the neural architecture search (NAS) algorithm to generative adversarial networks (GANs) shown in Fig.~\ref{chap05-fig:AutoGAN-architecture}. 
The search space of the \textbf{generator} architectural variations in Fig.~\ref{chap05-fig:AutoGAN-architecture} (b) are guided via an RNN controller as shown in Fig.~\ref{chap05-fig:AutoGAN-architecture} (a) together with parameter sharing and dynamic-resetting to accelerate the process. They use the Inception score as the reward, and introduce a multi-level search strategy to perform NAS in a progressive way. The authors use hinge loss for training the shared GAN, following the training setting of spectral
normalization GAN.

The whole pipeline is insightful but also poses novel challenges w.r.t the marriage of NAS and GANs. NAS remains to be optimized in vanilla classification problem, not to say the unstable training problems brought by GANs. Although in the paper AutoGAN shows promising results with NAS for GAN architecture, which is quit unique from manual design GAN architectures introduced above. We think it still has two critical issues to be solved:
\begin{itemize}
    \item The search space for the generator is limited and the search strategy for the discriminator is not discussed. 
    \item AutoGAN has not yet been tested on high-resolution image generation datasets. Thus, we can not have an intuitive estimation of applicability of this methods. The current image generation task is preliminary.
\end{itemize}

\subsection{MSG-GAN}\label{sec:MSG-GAN}
\begin{figure}[!htbp]
    \centering
    \includegraphics[width=1.\textwidth]{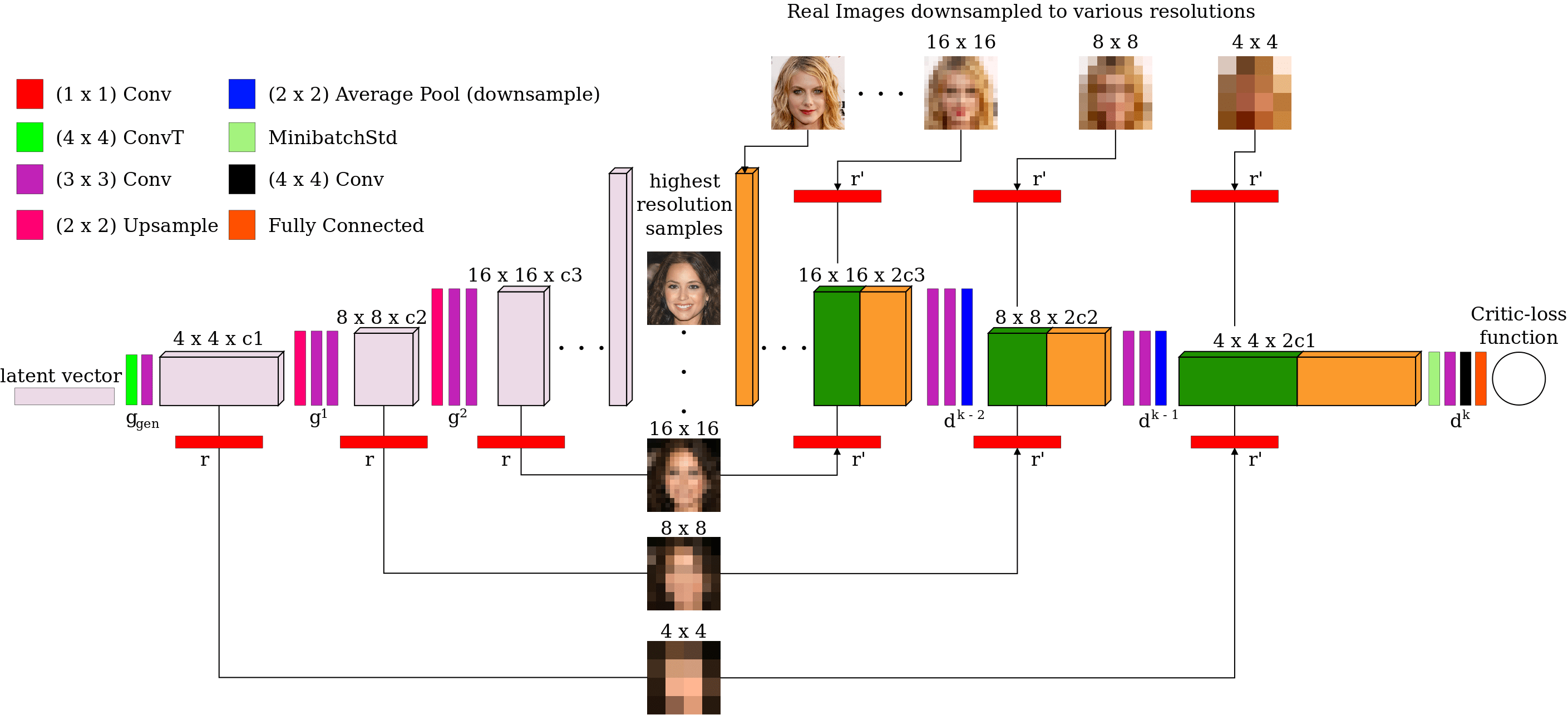}
    \caption{Architecture of MSG-GAN. Similar progressive training in PROGAN is deployed here. MSG-GAN includes connections from the intermediate layers of the generator to the intermediate layers of the discriminator. Multi-scale images are sent to the discriminator, which are concatenated with the corresponding main path i.e., green in the discriminator represents images generated from generator while orange represents original features in the main path of the discriminator. Figure regenerated and reorganized from~\cite{karnewar2020msg}.}
    \label{fig:MSG-GAN}
\end{figure}

It is known to all that GANs are extremely difficult to adapt to different datasets. Karnewar \textit{et al.} argue that one of the reason causing this issue is gradients passing from the discriminator to the generator become uninformative when there is not enough overlap in the supports of the real and fake distributions and propose MSG-GAN~\cite{karnewar2020msg} to handle such a problem. As seen in Figure~\ref{fig:MSG-GAN}, 
latent space of the generator and the discriminator is connected with each other, in which more information will be shared between the generator and the discriminator. More specifically, activations in each transpose convolutional steps (3 steps in Figure~\ref{fig:MSG-GAN}) in $G$ are mapped to an image at different scales by operation $r$ i.e., $1\times 1$ convolution in this case. Similarly, the mapped image is then encoded by $r^{'}$ to activations, which is concatenated with activations encoded by a real image. This connection enables more information shared between $D$ and $G$ and experimental results demonstrate benefits from this. The authors trained MSG-GAN on multiple datasets i.e., CIFAR10, Oxford flowers, LSUN, Indian Celebs, CelebA-HQ ($1024\times 1024$) and FFHQ ($1024 \times 1024$). The hyperparameter setting is almost same for all datasets. Specifically, $\mathbf{z}\in \mathbb{R}^{512\times 1}$ is drawn from a standard normal distribution. The RMSprop with a learning rate of 0.003 is used for both $D$ and $G$. WGAN-GP loss is used for training the network. Although MSG-GAN has achieved very good results on several image datasets, the ability of MSG-GAN for generating diverse images has not been tested yet and we found the result on CIFAR10 is not as good as other datasets carried out in the study. We guess this might be caused by the connection between $G$ and $D$ may constrain the diversity on $G$ as activations from $G$ and $D$ will be concatenated to with each other. The diversity on images might cause inconsistent matched activations, which pose negative impact on the training. More work can be investigated on this side such as adding self-attention module to improve the model diversity.

\subsection{\textbf{Summary}}
\begin{figure}[!ht]
    \centering
    \includegraphics[width=1.\textwidth]{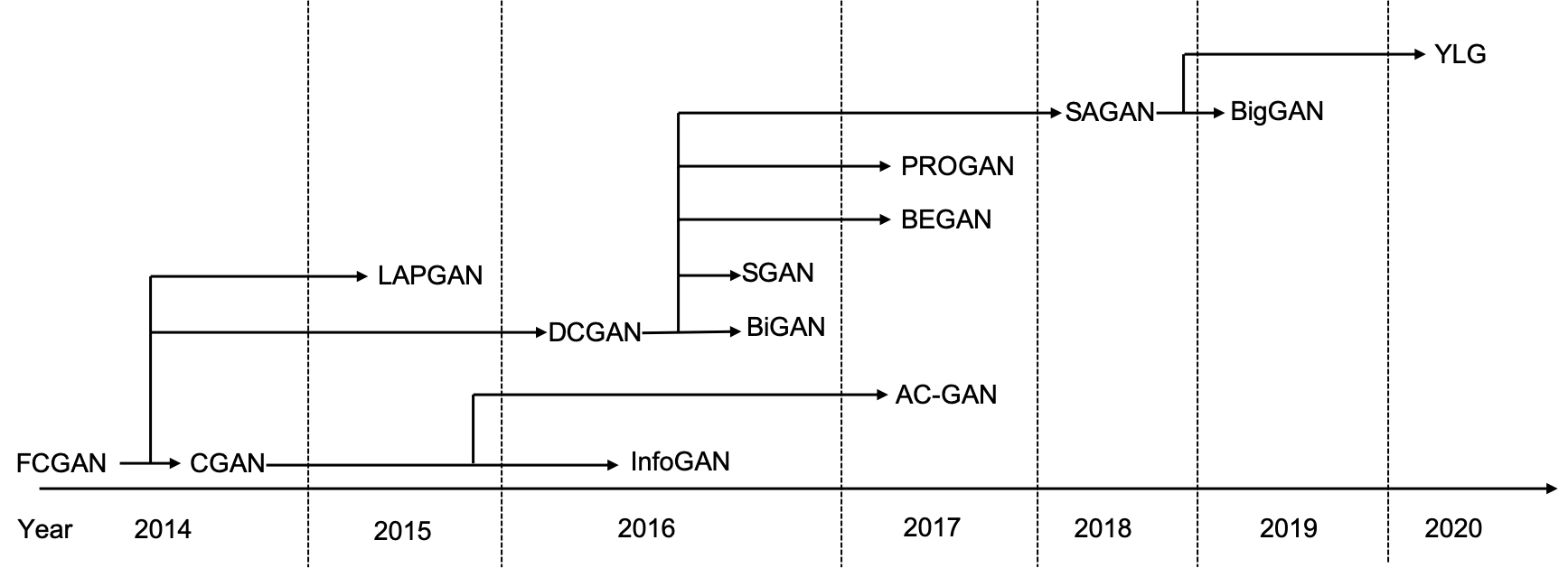}
    \caption{An overview of the footprint for architecture-variant GANs discussed in this section.}
    \label{fig:architecture_roadmap}
\end{figure}

\begin{figure}[!ht]
    \centering
    \includegraphics[width=.7\textwidth]{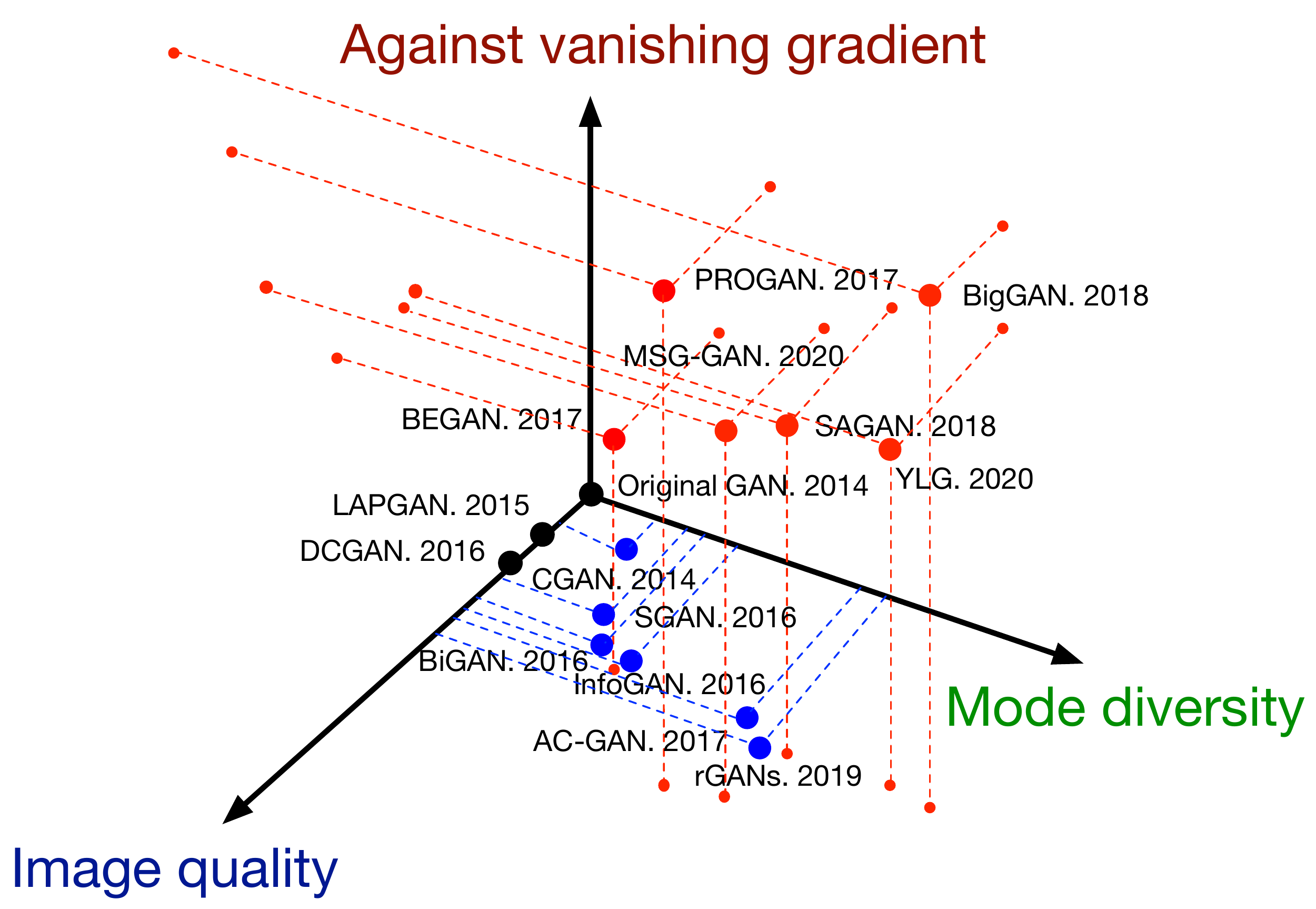}
    \caption{Summary of recent architecture-variant GANs for solving the three challenges. The challenges are categorized by three orthogonal axes. A larger value for each axis indicates better performance. Red points indicate GAN-variants which cover all three challenges, blue points cover two, and black points cover only one challenge. Quantitative results can be referred to Table~\ref{tab:performance} in section~\ref{sec:discussion}.}
    \label{chap05-fig:architecture_three_dimension}
\end{figure}

We have provided an overview of architecture-variant GANs which aim to improve performance based on the three key challenges: (1) Image quality; (2) Mode diversity; and (3) Vanishing gradient. Figure~\ref{fig:architecture_roadmap} 
illustrates a footprint for architecture-variant GANs from 2014 to 2020 that discussed in this section. It can be seen that there are lots of interconnections in different GAN variants. Fig.~\ref{chap05-fig:architecture_three_dimension}
illustrates relative performance related to three challenges. It is difficult to summarize quantitive results for each architecture-variant GAN because different GANs are designed for generating different types of images e.g., PROGAN and BEGAN are designed for generating face images, in which these two models may not have the capacity for generating diverse natural images. Here we summarize the performance according to different challenges based on the literature. It might be subjective for this summary because different GANs use different image datasets and different evaluation metrics. We suggest readers to follow the original articles in order to get deeper insights on the theory and the performance of each GAN variant.

All proposed architecture-variants are able to improve image quality. SAGAN is proposed for improving the capacity of multi-class learning in GANs, the goal of which is to produce more diverse images. Benefiting from the SAGAN architecture, BigGAN is designed for improving both image quality and image diversity. It should be noted that both PROGAN and BigGAN are able to produce high resolution images. BigGAN realizes this higher resolution by increasing the batch size and the authors mention that a progressive growing~\cite{karras2017progressive} operation is unnecessary when the batch size is large enough (2048 used in the original paper~\cite{brock2018large}). However, a progressive growing operation is still needed when GPU memory is limited (a large batch size is hungry for GPU memory). Benefiting from spectrum normalization (SN), which will be discussed in loss-variant GANs part, both SAGAN and BigGAN is effective for the vanishing gradient challenge. These milestone architecture-variants indicate a strong advantage of GANs --- compatibility, where a GAN is open to any type of neural architecture~\cite{goodfellow2016nips}. This property enables GANs to be applied to many different applications~\cite{mahapatra2019image,choi2018stargan,qiu2017deep}.

Regarding the improvements achieved by different architecture-variant GANs, we next present an analysis on the interconnections and comparisons between the architecture-variants presented here. Starting with the FCGAN described in the original GAN literature, this architecture-variant can only generate simple image datasets. Such a limitation is caused by the network architecture, where the capacity of FC networks is very limited. Research on improving the performance of GANs starts from designing more complex architectures for GANs. A more complex image datasets (e.g., ImageNet) has higher resolution and diversity comparing to simple image datasets (e.g., MNIST) and needs accordingly more sophisticated approaches. 

In the context of producing higher resolution images, one obvious approach is to increase the size of generator. LAPGAN and DCGAN up-sample the generator based on such a perspective. Benefiting from the concise deconvolutional up-sampling process and easy generalization of DCGAN, the architecture in DCGAN is more widely used in the GANs literature. It should be noticed that most GANs in the computer vision area use the deconvolutional neural network as the generator, which is first used in DCGAN. Therefore, DCGAN is one of the classical GAN-variants in the literature. 

SGAN, BiGAN, CGAN, InfoGAN and AC-GAN share similar properties that have been added to the GANs, which improve the discriminative ability for GANs. CGAN is the first work that introduces adding encoded labels together with images to the discriminator and noise input to the generator, in which the input noise and the input image are now encoded with labels. Based on CGAN, apart from distinguishing real and fake samples, InfoGAN adds another classifier $Q$ to estimate that input images (including generated images) belong to which class. AC-GAN is very similar to InfoGAN. The difference between AC-GAN and InfoGAN is that real images are also conditioned by labels and the classifier predict that both generated images and real images belong to which class. Similar to this modification, SGAN adds a multi-headed last layer to the discriminator, which contains softmax and sigmoid so the discriminator is able to distinguish real and fake and classify images at the same time. BiGAN introduces learning the inverse mapping, which also shows the improvement on the quality of generated images. The architecture-variants mentioned here all have similary properties, which is to add more encoding information to GANs compared to the original GAN. The performance of these types depend on the dataset that should be well-labeled, which may pose challenges on some real-world applications e.g., the dataset is not labeled, is partially labeled or contains noisy labels (rGANs in section~\ref{sec:rGANs}). Extension of such an architecture can be a combination with self-supervised learning manner.    

The ability to produce high quality images is an important aspect of GANs clearly. This can be improved through judicious choice of architecture. BEGAN and PROGAN demonstrate approaches from this perspective. With the same architecture used for the generator in DCGAN, BEGAN redesigns the discriminator by including encoder and decoder, where the discriminator tries to distinguish the difference between the generated and autoencoded images in pixel space. Image quality has been improved in this case. Based on DCGAN, PROGAN demonstrates a progressive approach that incrementally trains an architecture similar to DCGAN. This novel approach cannot only improve image quality but also produce higher resolution images.       

Producing diverse images is the most challenging task for GANs and it is very difficult for GANs to successfully produce images such as those represented in the ImageNet sets. It is difficult for traditional CNNs to learn global and long-range dependencies from images. Thanks to self-attention mechanism though, approaches such as those in SAGAN integrate self-mechanisms to both discriminator and generator, which helps GANs a lot in terms of learning multi-class images. Moreover, BigGAN, which can be considered an extension of SAGAN, introduces a deeper GAN architecture with a very large batch size, which produces high quality and diverse images as in ImageNet and is the current state-of-the-art.  

Here we give a recap on how architecture-variant GANs remedy each challenge:

\vspace{10pt}
\noindent
\textbf{Image Quality}\hspace{10pt} One of the basic objectives of GANs is to produce more realistic images which requires high image quality. The original GAN (FCGAN) is only applied to MNIST, Toronto face dataset and CIFAR-10 because of its limited capacity of the architecture. DCGAN and LAPGAN introduce the deconvolutional process and the up-sampling process to the architecture respectively. These two processes enable the model have larger capacity to produce higher resolution images. The rest of architecture variants (i.e., BEGAN, PROGAN, SAGAN and BigGAN) all have some modifications on the loss function e.g., use the Wasserstein distance as loss function and we will address this aspect in the later of the paper, which are also beneficial to the image quality. Regarding the architecture only, BEGAN uses an autoencoder architecture for the discriminator, which compares generated images and real images in pixel level. This helps generator produce easy-to-reconstruct data. PROGAN utilizes a deeper architecture and the model is growing with the training progressing. This progressive training strategy improves the learning stability for discriminator and generator thus it is easier for the model to learn how to produce high resolution images. SAGAN mainly benefits from the SN which we will address in the next section. BigGAN demonstrates that high resolution image generation can benefit from a deeper model with larger batch size.

\vspace{10pt}
\noindent
\textbf{Vanishing Gradient}\hspace{10pt}  Changing the loss function is the only way to remedy such a problem. Some architecture variants here avoid the vanishing gradient issue because of using different loss functions where we will revisit this in the next section.

\vspace{10pt}
\noindent
\textbf{Mode Dversity}\hspace{10pt}  This is the most challenging problem for GANs. It is very difficult for GANs to produce realistic diverse images such as natural images. In terms of architecture-variant GANs, only SAGAN and BigGAN address such kind of issue. Benefiting from self-attention mechanism, CNNs in SAGAN and BigGAN can process large receptive field which overcomes the components shitting problems in generated images. This enables such type of GANs are able to produce diverse images.

\section{\textbf{Loss-variant GANs}} \label{Loss-variant-GANs}
\begin{figure*}[!ht]
    \centering
    \includegraphics[width=.9\textwidth]{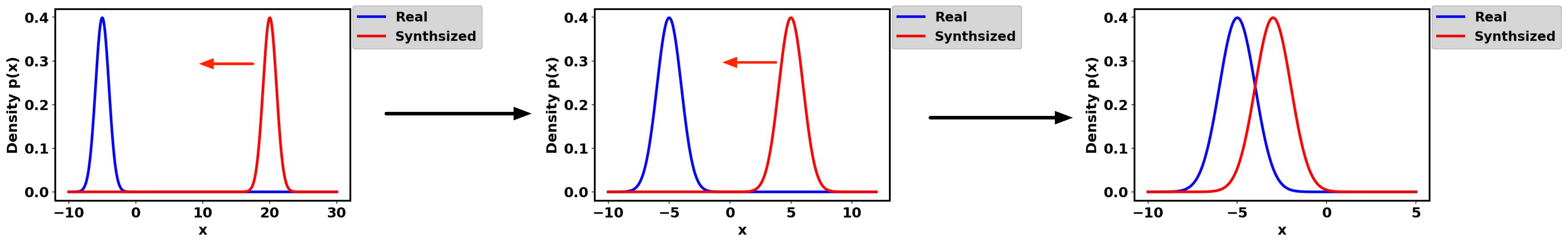}
    \caption{Illustration of training progress for a GAN. Two normal distributions are used here for visualization. Given an optimal $D$, the objective of GANs is to update $G$ in order to move the generated distribution $p_{g}$ (red) towards the real distribution $p_{r}$ (blue) ($G$ is updated from left to right in this figure. Left: initial state, middle: during training, right: training converging). However, JS divergence for the left two figures are both 0.693 and the figure on the right is 0.336, indicating that JS divergence does not provide sufficient gradient at the initial state.}
    \label{chap05-fig:JSD-pg-pr}
\end{figure*}

\begin{figure}[!ht]
    \centering
    \subfigure[JS divergence changes with distance.]{
    \centering
    \includegraphics[width=.4\textwidth]{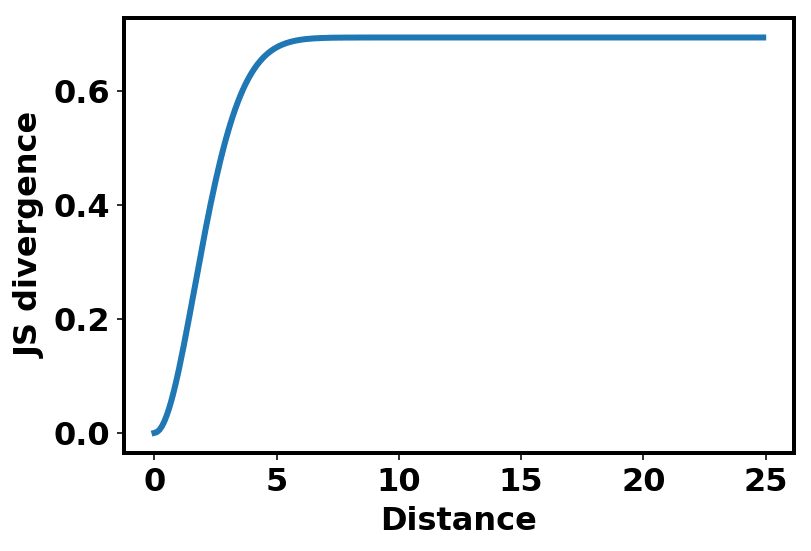}
    \label{chap05-fig:JS-curve}
    }
    \subfigure[Gradient JS divergence changes with distance.]{
    \centering
    \includegraphics[width=.4\textwidth]{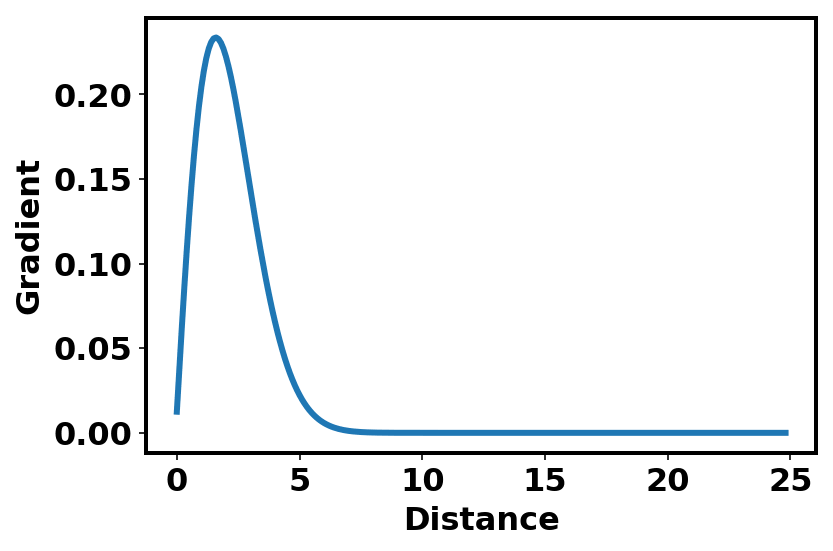}
    \label{chap05-fig:JS-curve-gradient}
    }    
    \caption{JS divergence and gradient change with the distance between $p_{r}$ and $p_{g}$. The distance is the difference between two distribution means.}
    \label{chap05-fig:JS-curve-summary}
\end{figure}

Another design decision in GANs which significantly impacts performance is the choice of loss function in equation~\eqref{eq:GAN-formula}. While the original GAN work~\cite{goodfellow2014generative} has already proved global optimality and the convergence of GANs training. It still highlights the instability problem which can arise when training a GAN. The problem is caused by the global optimality criterion as stated in~\cite{goodfellow2014generative}. Global optimality is achieved when an optimal $D$ is reached for any $G$. So the optimal $D$ is achieved when the derivative of $D$ for the loss in equation~\eqref{eq:GAN-formula} equals 0. So we have 
\begin{equation}
\begin{aligned}
    -\frac{p_{r}(\mathbf{x})}{D(\mathbf{x})}+\frac{p_g(\mathbf{x})}{1-D(\mathbf{x})}=0,\\
    D^{*}(\mathbf{x})=\frac{p_{r}(\mathbf{x})}{p_{r}(\mathbf{x})+p_{g}(\mathbf{x})},
\end{aligned}
\end{equation}
where $\mathbf{x}$ represents the real data and generated data,  $D^{*}(\mathbf{x})$ is the optimal discriminator, $p_r(\mathbf{x})$ is the real data distribution and $p_g(\mathbf{x})$ is the generated data distribution. We have got the optimal discriminator $D$ so far. When we have the optimal $D$, the loss for $G$ can be visualized by substituting $D^{*}(\mathbf{x})$ into equation~\eqref{eq:GAN-formula}
\begin{equation}
\begin{aligned}
    \mathcal{L}_{G}=&\mathbb{E}_{\bm{\mathrm{x}} \sim p_{r}} \mathrm{log} \frac{p_{r}(\bm{\mathrm{x}})}{\frac{1}{2}\left[p_{r}(\bm{\mathrm{x}})+p_{g}(\bm{\mathrm{x}})\right]} + \mathbb{E}_{\bm{\mathrm{x}} \sim p_{g}} \mathrm{log} \frac{p_{g}(\bm{\mathrm{x}})}{\frac{1}{2}\left[p_{r}(\bm{\mathrm{x}})+p_{g}(\bm{\mathrm{x}})\right]} - 2 \cdot\mathrm{log}2.
    \label{chap05-eq:GAN-loss-G}
\end{aligned}
\end{equation}
Equation~\eqref{chap05-eq:GAN-loss-G} demonstrates the loss function for a GAN when discriminator is optimized and it is related to two important probability measurement metrics. One is Kullback–Leibler (KL) divergence which is defined as
\begin{equation}
    KL(p_{1}\|p_{2})=\mathbb{E}_{\bm{\mathrm{x}} \sim p_{1}} \mathrm{log} \frac{p_{1}}{p_{2}},
\end{equation}
and the other is Jensen-Shannon (JS) divergence which is stated as
\begin{equation}
\begin{aligned}
    JS(p_{1}\|p_{2})=&\frac{1}{2}KL(p_{1}\|\frac{p_{1}+p_{2}}{2})+\frac{1}{2}KL(p_{2}\|\frac{p_{1}+p_{2}}{2}).
    \label{chap05-eq:JS}
\end{aligned}
\end{equation}
Thus the loss for $G$ regarding the optimal $D$ in equation~\eqref{chap05-eq:GAN-loss-G} can be reformulated as 
\begin{equation}
    \mathcal{L}_{G}=2\cdot JS(p_{r}\|p_{g}) - 2 \cdot\mathrm{log}2,
    \label{chap05-eq:JS-loss-G}
\end{equation}
which indicates that the loss for $G$ now equally becomes the minimization of the JS divergence between $p_{r}$ and $p_{g}$. With the training $D$ step by step, the optimization of $G$ will be closer to the minimization of JS divergence between $p_{r}$ and $p_{g}$. We now start to explain the unstable training problem, where $D$ often easily wins $G$. This unstable training problem is actually caused by the JS divergence in equation~\eqref{chap05-eq:JS}. Give an optimal $D$, the objective of optimization for equation~\eqref{chap05-eq:JS-loss-G} is to move $p_{g}$ toward $p_{r}$ (see Fig.~\ref{chap05-fig:JSD-pg-pr}). 
JS divergence for the three plots from left to right are 0.693, 0.693 and 0.336, which indicates that JS divergence stays constant (log2=0.693) if there is no overlap between $p_{r}$ and $p_{g}$. Figure~\ref{chap05-fig:JS-curve-summary} 
demonstrates the change of JS divergence and its gradient corresponding to the distance between $p_{r}$ and $p_{g}$. It can be seen that JS divergence is constant and its gradient is almost 0 when the distance is greater than 5, which indicates that training process does not have any effect on $G$. The gradient of JS divergence for training the $G$ is non-zero only when $p_g$ and $p_{r}$ have substantial overlap i.e., the vanishing gradient will arise for $G$ when $D$ is close to optimal. In practice, the possibility that $p_{r}$ and $p_{g}$ do not overlap or have negligible overlap is very high~\cite{MAtowards}. 

\begin{figure}[!ht]
    \centering
    \subfigure[KL divergence.]{
    \centering
    \includegraphics[width=.4\textwidth]{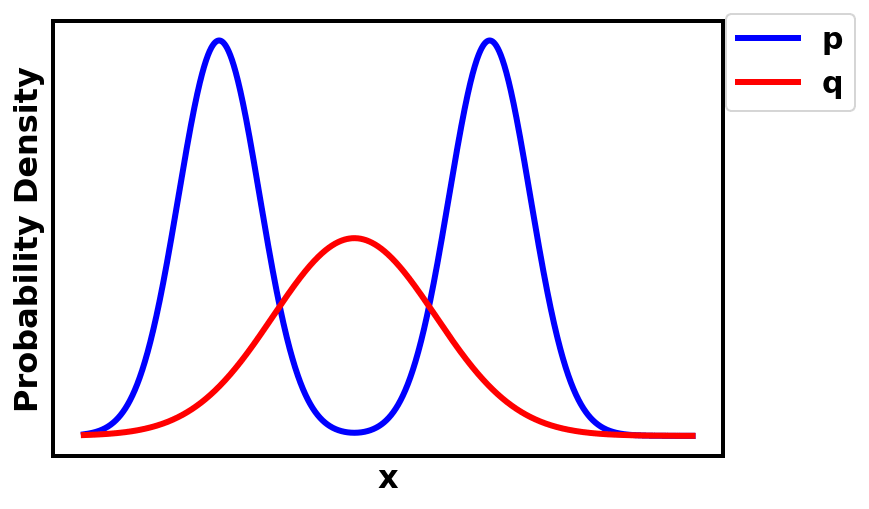}
    \label{chap05-fig:forward_KL}
    }
    \subfigure[Reverse KL divergence.]{
    \centering
    \includegraphics[width=.4\textwidth]{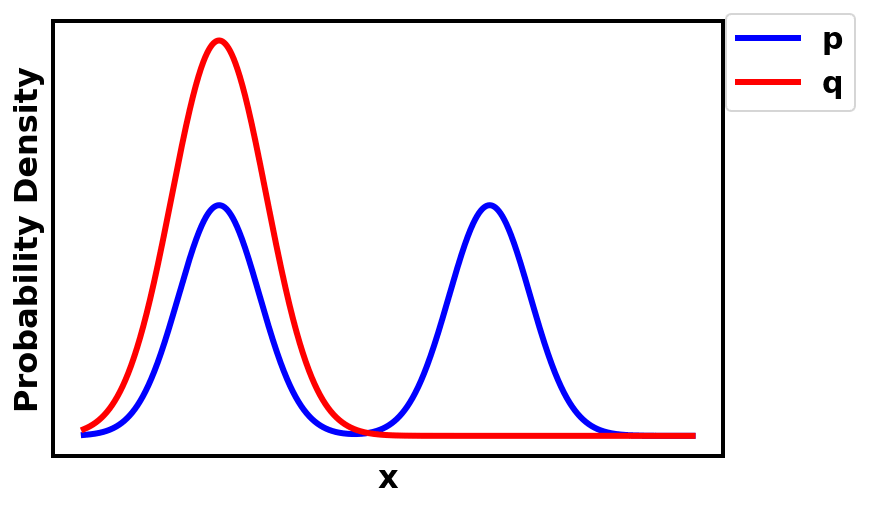}
    \label{chap05-fig:reverse_KL}
    }    
    \caption{Optimized $q$ (in red) when minimizing KL divergence $KL(p\|q)$ and reverse KL divergence $KL(q\|p)$.}
    \label{chap05-fig:KL_summary}
\end{figure}

The original GANs work~\cite{goodfellow2014generative} also highlights the minimization of $-\mathbb{E}_{\mathbf{x}\sim p_{g}}\mathrm{log}[D(\mathbf{x})]$ for training $G$ to avoid a vanishing gradient. However, this training strategy will lead to another problem called mode dropping. First, let us examine $KL(p_{g}\|p_{r})=\mathbb{E}_{\mathbf{x}\sim p_{g}}\mathrm{log}\frac{p_{g}}{p_r}$. With an optimal discriminator $D^*$, $KL(p_{g}\|p_{r})$ can be reformulated as
\begin{equation}
\begin{aligned}
    KL(p_{g}\|p_{r})&=\mathbb{E}_{\mathbf{x}\sim p_{g}}\mathrm{log}\frac{p_g(\mathbf{x})/(p_r(\mathbf{x})+p_g(\mathbf{x}))}{p_r(\mathbf{x})/(p_r(\mathbf{x})+p_g(\mathbf{x}))}, \\
    &=\mathbb{E}_{\mathbf{x}\sim p_{g}}\mathrm{log}\frac{1-D^{*}(\mathbf{x})}{D^{*}(\mathbf{x})}, \\
    &=\mathbb{E}_{\mathbf{x}\sim p_{g}}\mathrm{log}[1-D^*(\mathbf{x})]-\mathbb{E}_{\mathbf{x}\sim p_{g}}\mathrm{log[D^*(\mathbf{x})]}.
\end{aligned}    
\label{chap05-eq:KL-pg-pr}
\end{equation}
The alternative loss form for $G$ now can be stated by switching the order of the two sides in equation~\eqref{chap05-eq:KL-pg-pr}
\begin{equation}
\begin{aligned}
   &-\mathbb{E}_{\mathbf{x}\sim p_{g}}\mathrm{log}[D^*(\mathbf{x})]\\
   &=KL(p_{g}\|p_{r})-\mathbb{E}_{\mathbf{x}\sim p_{g}}\mathrm{log}[1-D^*(\mathbf{x})],\\
   &=KL(p_{g}\|p_{r})-2\cdot JS(p_{r}\|p_{g})+2\cdot \mathrm{log}2+\mathbb{E}_{\mathbf{x}\sim p_{x}}\mathrm{log}[D^*(\mathbf{x})],
\end{aligned}
\label{chap05-eq:G-loss-alternative}
\end{equation}
where the alternative loss for $G$ in equation~\eqref{chap05-eq:G-loss-alternative} is only affected by the first two terms (the last two terms are constant), however, this loss function is dominated by $KL(p_{g}\|p_{r})$ since $JS(p_{r}\|p_{g})$ is bounded in $[0, \log2]$ as illustrated in Figure~\ref{chap05-fig:JS-curve}. It can be noticed that the first term in equation~\eqref{chap05-eq:G-loss-alternative} is reverse KL divergence, in which the $p_g$ optimized by the reverse is totally different from the $p_g$ optimized by KL divergence. Figure~\ref{chap05-fig:KL_summary} illustrates the this difference by using a mixture of two Gaussians for $p$ and a single Gaussian for $q$. When $p$ has multiple modes, $q$ tries to blur all modes together in order to put high probability mass on all of them as seen in Figure~\ref{chap05-fig:forward_KL}. However, Figure~\ref{chap05-fig:reverse_KL} shows that $q$ chooses recover a single Gaussian in order to avoid putting probability mass in the low-probability areas at the middle of two Gaussians.
The optimization on reverse KL divergence therefore will cause the mode collapse during training GANs, which is highlighted below
\begin{itemize}
    \item \textit{When $p_g(\mathbf{x})\to 0$, $p_r(\mathbf{x})\to 1$, $KL(p_{g}\|p_{r}) \to 0$.}
    \item \textit{When $p_g(\mathbf{x})\to 1$, $p_r(\mathbf{x})\to 0$, $KL(p_{g}\|p_{r}) \to + \infty$.}
\end{itemize}
The penalization for two instances of poor performance made by $G$ are totally different. The first instance of poor performance is that $G$ is not producing a reasonable range of samples and yet incurs a very small penalization. The second instance of poor performance concerns $G$ producing implausible samples but has very large penalization. The first example concerns the fact that the generated samples lack diversity while the second concerns that fact that the generated samples are not accurate. Considering this first case, $G$ generates repeated but ``safe'' samples instead of taking risk to generate diverse but ``unsafe'' samples, which leads to the mode collapse problem. In summary, using the original loss in equation~\eqref{eq:GAN-formula} will result in the vanishing gradient for training $G$ and using the alternative loss in equation~\eqref{chap05-eq:G-loss-alternative} will incur the mode collapse problem. These kind of problems cannot be solved by changing the GANs architectures. Therefore, it could be argued that ultimate GANs problem stems from the design of the loss function and that innovative ideas for this redesign of the loss function may solve the problem.

Loss-variant GANs have been researched extensively to improve the stability of training GANs. 

\subsection{Wasserstein GAN (WGAN)}
WGAN~\cite{arjovsky2017wasserstein} has successfully solved the two problems for the original GAN by using the Earth mover (EM) or Wasserstein-1~\cite{rubner2000earth} distance as the loss measure for optimization. The EM distance is defined as
\begin{equation}
    W(p_r, p_g)=\inf_{\gamma \in \prod(p_r,p_g)} \mathbb{E}_{(\mathbf{x},\mathbf{y})\sim \gamma}\|\mathbf{x}-\mathbf{y}\|,
    \label{chap05-eq:wasserstein_distance}
\end{equation}
where $\prod(p_r,p_g)$ denotes the set of all joint distributions and  $\gamma(\mathbf{x},\mathbf{y})$ whose marginals are $p_r$ and $p_g$. Compared with KL and JS divergence, EM is able to reflect distance even when $p_r$ and $p_g$ do not overlap and it is also continuous and thus able to provide meaningful gradient for training the generator. Figure~\ref{chap05-fig:G-loss-gradient}
illustrates the gradient of WGAN comparing to the original GAN. It is noticeable that WGAN has a smooth gradient for training the generator spanning the complete space. However, the infimum in equation~\eqref{chap05-eq:wasserstein_distance} is intractable but the creators demonstrate that instead the Wasserstein distance can be estimated as
\begin{equation}
    \max_{w\sim\mathcal{W}}\mathbb{E}_{\mathbf{x}_{p_r}}[f_w(\mathbf{x})]-\mathbb{E}_{\mathbf{z}\sim p_z}[f_w(G(\mathbf{z}))],
    \label{chap05-eq:WGAN-D-loss}
\end{equation}
where $f_w$ can be realized by $D$ but has some constraints (for details the interested reader can refer to the original work~\cite{arjovsky2017wasserstein}) and $\mathbf{z}$ is the input noise for $G$. So $w$ here is the parameters in $D$ and $D$ aims to maximize equation~\eqref{chap05-eq:WGAN-D-loss} in order to make the optimization distance equivalent to Wasserstein distance. When $D$ is optimized, equation~\eqref{chap05-eq:wasserstein_distance} will become the Wasserstein distance and $G$ aims to minimize it. So the loss for $G$ is
\begin{equation}
    -\min_G\mathbb{E}_{\mathbf{z}\sim p_z}[f_w(G(\mathbf{z}))]
    \label{chap05-eq:WGAN-G-loss}
\end{equation}
An important difference between WGAN and the original GAN is the function of $D$. The $D$ in the original work is used as a binary classifier but $D$ used in WGAN is to fit the Wasserstein distance, which is a regression task. Thus, the sigmoid in the last layer of $D$ is removed in the WGAN. The authors train WGAN on LSUN dataset with $64\times 64$ resolution. \textit{Importantly, training of WGAN will be unstable at times when a momentum based optimizer such as Adam ($\beta_1 > 0$ is used)}. Therefore, RMSProp is utilized for training WGAN.

\subsection{WGAN-GP}
\begin{figure}[!ht]
    \centering
    \subfigure[Weights of WGAN]{\includegraphics[width=.4\textwidth]{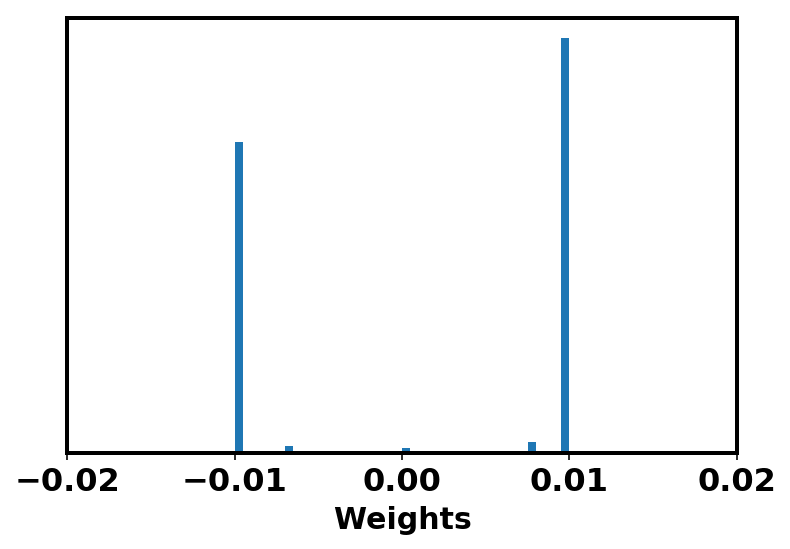}}
    \subfigure[Weights of WGAN-GP]{\includegraphics[width=.4\textwidth]{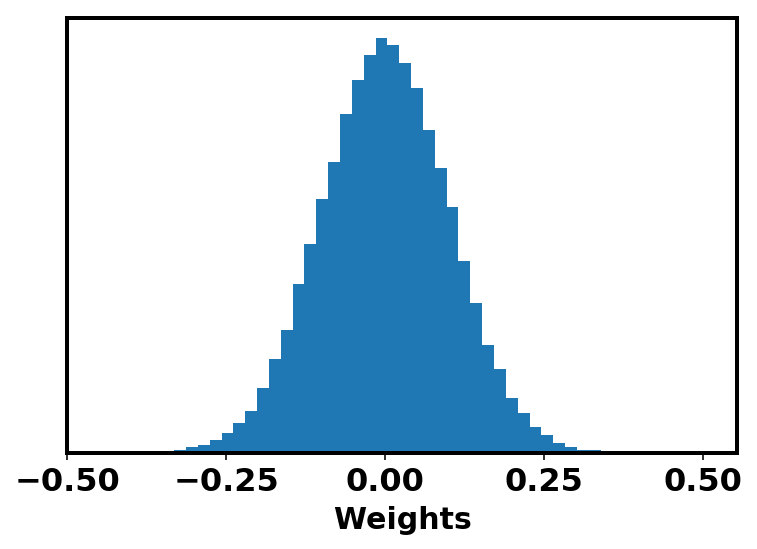}}    
    \caption{Comparison of parameter distribution between WGAN and WGAN-GP. Top is WGAN and bottom is WGAN-GP. Figure from~\cite{gulrajani2017improved}.}
    \label{chap05-fig:WGAN-WGANGP-para}
\end{figure}

Even though WGAN has been shown to be successful in improving the stability of GAN training, it is not well generalized for a deeper model. Experimentally it has been determined that most WGAN parameters are localized at -0.01 and 0.01 because of parameter clipping. This will dramatically reduce the modeling capacity of $D$. WGAN-GP has been proposed using gradient penalty for restricting $\|f \|_L \leq K$ for the discriminator ~\cite{gulrajani2017improved} and the modified loss for discriminator now becomes
\begin{equation}
\begin{aligned}
    \mathcal{L}_D=&\mathbb{E}_{\mathbf{x}_g\sim p_g}[D(\mathbf{x}_g)]-\mathbb{E}_{\mathbf{x}_r\sim p_r}[D(\mathbf{x}_r)] + \lambda \mathbb{E}_{\hat{\mathbf{x}}\sim p_{\hat{x}}}[(\| \triangledown
    _{\hat{\mathbf{x}}}D(\hat{\mathbf{x}})\|_2-1)^2],
\end{aligned}
\end{equation}
where $\mathbf{x}_r$ is sample data drawn from the real data distribution $p_r$,  $\mathbf{x}_g$ is sample data drawn from the generated data distribution $p_g$ and $p_{\hat{\mathbf{x}}}$ is sampled uniformly along the straight lines between those pairs of points, which are sampled from the real data distribution $p_r$ and the generated data distribution $p_g$. The first two terms are original loss in WGAN and the last term is the gradient penalty. WGAN-GP demonstrates a better distribution of trained parameters compared to WGAN (Fig.~\ref{chap05-fig:WGAN-WGANGP-para}) and better stability performance during training of GANs. Before WGAN-GP, successful training on GANs only took place on those models consisting of few layers both in the discriminator and generator i.e., DCGAN uses 4 convolutional layers in $D$ and 4 deconvolutional layers in $G$. WGAN-GP successfully demonstrates stable training of WGAN-GP by using the ResNet-101 architecture as the backbone, which has the impact on the GANs research on large-scale image generation i.e., PROGAN, BigGAN. As mentioned in the previous section, WGAN has unstable issues when using the momentum based optimizer such as Adam. WGAN-GP shows the stable training by using the Adam optimizer and even faster convergence using the same training settings. WGAN-GP was experimented on ImageNet with $32\times 32$ image resolution, LSUN dataset with $128\times 128$ image resolution and CIFAR-10 with $32\times 32$ image resolution. Adam optimizer with $\alpha=1e-4$, $\beta_1=0$, $\beta_2=0.9$ was utilized in the experiment. Learning rate was $2e-4$ and batch size was 64. The authors find piecewise linear activation functions e.g., ReLU and leaky ReLu and smooth activation functions e.g., Tanh both can train WGAN-GP stably.   

\subsection{Least Square GAN (LSGAN)}
\begin{figure*}[ht!]
    \centering
    \subfigure[]{
        \includegraphics[width=.3\textwidth]{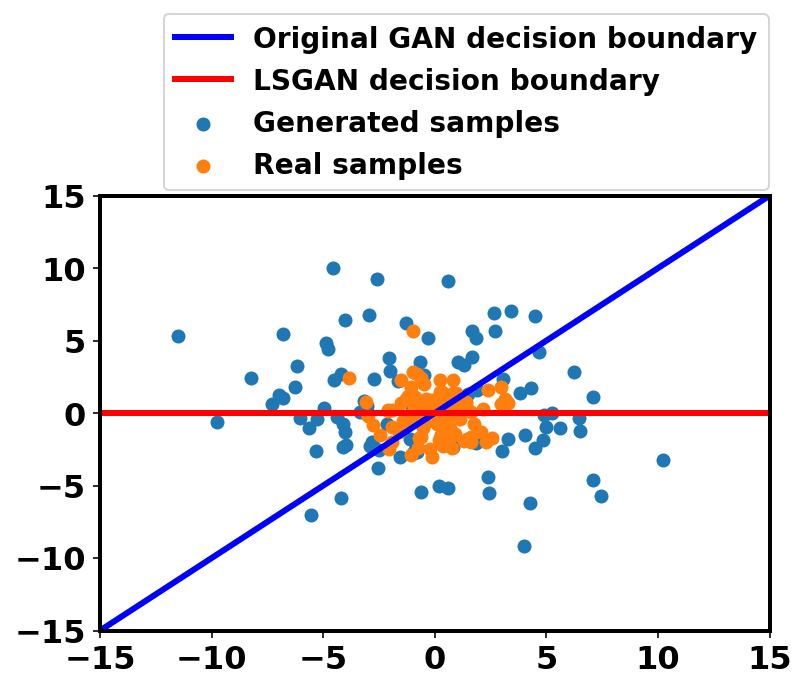}
    }
    \subfigure[]{
        \includegraphics[width=.3\textwidth]{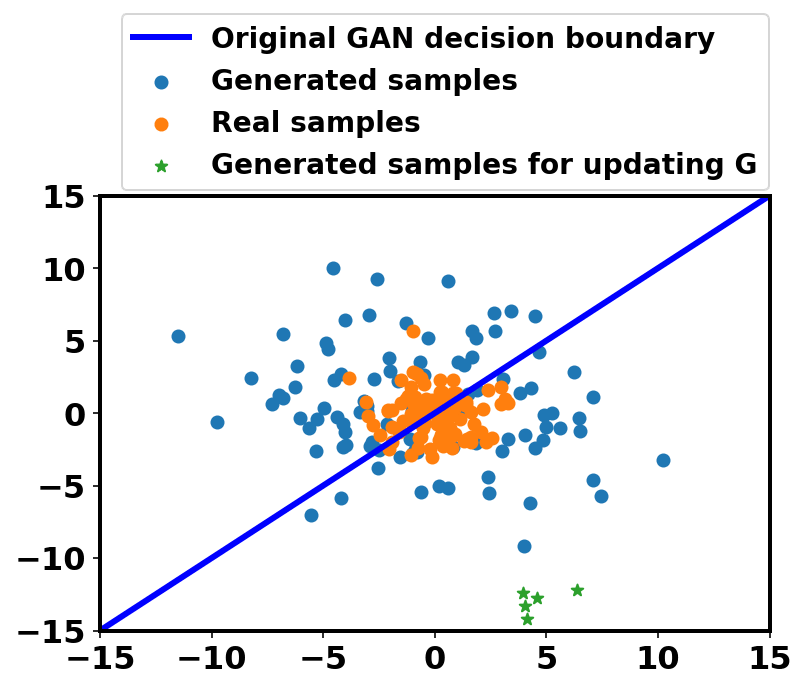}
    }
    \subfigure[]{
        \includegraphics[width=.3\textwidth]{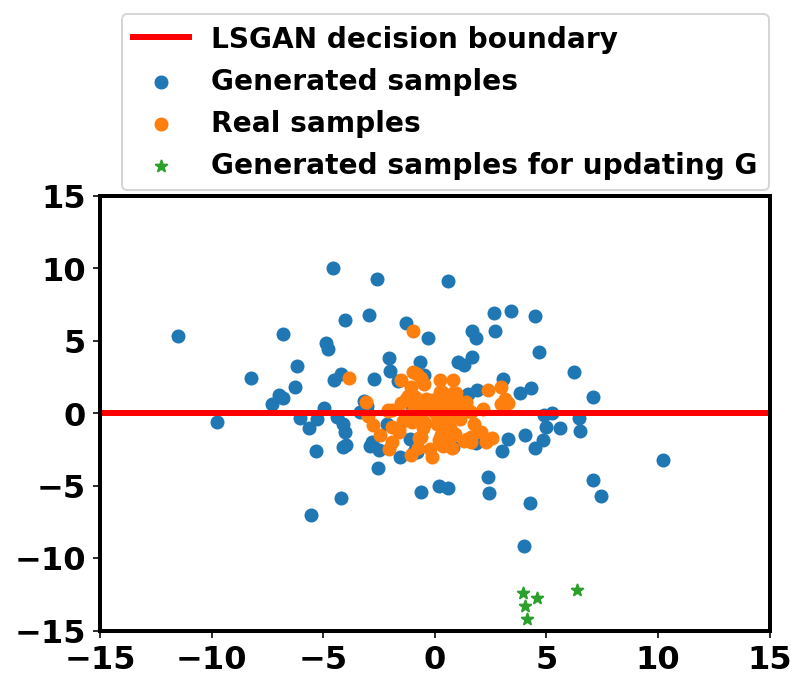}
    }
    \caption{Decision boundary illustration of original GAN and LSGAN. (a). Decision boundaries for $D$ of original GAN and LSGAN. (b). Decision boundary of $D$ for the original GAN. It has small errors for the generated samples, which is far away from the decision boundary (in green), for updating $G$. (c). Decision boundary for $D$ of LSGAN. It penalizes the large error for any generated sample that is far away from the boundary (in green). Thus it pushes generated samples (in green) toward the boundary~\cite{mao2017least}.}
    \label{chap05-fig:LS-boundary}
\end{figure*}

The LSGAN is a new approach proposed in~\cite{mao2017least} to remedy the vanishing gradient problem for $G$ from the perspective of the decision boundary determined by the discriminator. This work argues that the decision boundary for $D$ of the original GAN penalizes very small error to update $G$ for those generated samples that are far away from the decision boundary. The author proposes using a least square loss for $D$ instead of sigmoid cross entropy loss stated in the original paper~\cite{goodfellow2014generative}. The proposed loss function is defined as
\begin{equation}
\begin{aligned}
    &\min_{D}\mathcal{L}_D=\frac{1}{2}\mathbb{E}_{\mathbf{x}\sim p_r}[(D(\mathbf{x})-b)^2]+\frac{1}{2}\mathbb{E}_{\mathbf{z}\sim p_z}[(D(G(\mathbf{z}))-a)^2],\\
    &\min_{G}\mathcal{L}_G=\frac{1}{2}\mathbb{E}_{\mathbf{z}\sim p_z}[(D(G(\mathbf{z}))-c)^2],
\end{aligned}
\label{chap05-eq:LS-eq}
\end{equation}
where $a$ is the label for the generated samples, $b$ is the label for the real samples and $c$ is the hyperparameter that $G$ wants $D$ to recognize the generated samples as the real samples by mistake. This modified change has two benefits: (1) The new decision boundary made by $D$ penalizes large error arising from those generated samples that are far away from the decision boundary, which pushes those ``bad'' generated samples towards the decision boundary. This is beneficial in terms of generating improved image quality; (2) Penalizing the generated samples that are far away from the decision boundary is able to provide sufficient gradient when updating the $G$, which remedies the vanishing gradient problems for training $G$. Figure~\ref{chap05-fig:LS-boundary} 
demonstrates the comparison of decision boundaries for LSGAN and the original GAN. The decision boundaries for $D$ that have been trained by the original sigmoid cross entropy loss and the proposed least square loss are different.
The work~\cite{mao2017least} has proven that the optimization of LSGAN is equivalent to minimizing the Pearson $\chi^2$ divergence between $p_r+p_g$ and $2p_g$ when $a$, $b$ and $c$ subsject to $b-c=1$ and $b-a=2$. Similar to WGAN, $D$ here behaves as regression and the sigmoid is also removed. LSGAN was evaluated on LSUN and HWDB1.0~\cite{liu2013online} with $112\times 112$ image resolution. The Adam optimizer with $\beta_1=0.5$ was used and the learning rate was $1e-3$ and $2e-4$ for LSUN and HWDB1.0 respectively. Similary to DCGAN, ReLU activations and Leaky ReLU activations were used for the generator and the discriminator respectively.

\subsection{$f$-GAN}
$f$-GAN summarizes that GANs can be trained by using an $f$-divergence~\cite{nowozin2016f}. $f$-divergence $D_f(p_r\|p_g)$ measures the difference between two probability distributions ($p_r$ and $p_g$ regarding GANs) e.g., KL divergence, JS divergence and Pearson $\chi^2$ as mentioned before, which can be summarized as
\begin{equation}\label{eq:f-GAN}
    D_f(p_r\|p_g) = \int_\mathcal{X}p_g(x)f\left(\frac{p_r(x)}{p_g(x)}\right)dx
\end{equation}
where $f$ is a convex function and $f(1) = 0$. It should be noted that $f$ is termed as generator function in the original paper~\cite{nowozin2016f}, which is totally different from the concept generator $G$ in GANs. Thus we use $f$ or $f$-divergence function in this section instead of generator function in the original paper in order to avoid confusion with generator $G$ in this paper. $f$-GAN generalizes the loss function of GANs according to $f$-divergence function presented in equation~\eqref{eq:f-GAN}. List of $f$-divergence with $f$-divergence function is shown in Table~\ref{tab:f-GAN}.
\begin{table}[!ht]
    \centering
    \begin{tabular}{c|c|c}
        \toprule
        Divergence &  $D_f(p_g\|p_q)$ & f-divergence function\\ \midrule
        KL divergence & $\int p_r(x)\log\frac{p_r(x)}{p_g(x)}dx$& $t\log t$\\ 
        Reverse KL& $\int p_g(x)\log\frac{p_g(x)}{p_r(x)}dx$& $-\log t$\\
        FCGAN ($2\cdot JS-2\cdot \log 2$)~\cite{goodfellow2014generative} & $\int p_r(x)\log\frac{2p_r(x)}{p_r(x)+p_g(x)}+p_g(x)\log\frac{2p_g(x)}{p_r(x)+p_g(x)}dx-2\cdot \log 2$&$t\log t - (t+1)\log(t+1)$\\
        LSGAN (Pearson $\mathcal{X}^2$)~\cite{mao2017least} & $\int \frac{(p_g(x)-p_r(x))^2}{p_r(x)}dx$ & $(t-1)^2$ \\ 
        EBGAN~\cite{zhao2016energy} & $\int |p_r(x)-p_g(x)|dx$ & $|t-1|$ \\
        \bottomrule
    \end{tabular}
    \caption{Examples of $D_f(p_r\|p_g)$ with $f$-divergence function based on~\cite{nielsen2013chi,nowozin2016f}.} 
    \label{tab:f-GAN}
\end{table}
However, equation~\eqref{eq:f-GAN} is intractable thus it requires being estimated as a tractable form such as expectation form. By using the convex conjugate (Fenchel conjugate) $f(u) = \mathrm{sup}_{t\in \mathrm{dom}_{f^*}}\{tu-f^*(t)\}$~\cite{hiriart2012fundamentals}, $f$-divergence can be represented as a lower bound on the divergence
\begin{equation}
\begin{aligned}
    D_f(p_r\|p_g) = & \int_\mathcal{X} p_g(x) \underset{t\in \mathrm{dom}_{f^*}}{\mathrm{sup}}\left(t\frac{p_r(x)}{p_g(x)}-f^*(t)\right)dx \\
    & \geq \underset{T\in \mathcal{T}}{\mathrm{sup}}\left(\int_{\mathcal{X}}T(x) p_r(x) - f^*(T(x))p_g(x)) \right)dx \\
    & = \underset{T\in \mathcal{T}}{\mathrm{sup}} \left(\mathbb{E}_{x\in p_r}[T(x)] - \mathbb{E}_{x\in p_g}[f^*(T(x))] \right)
\end{aligned}
\label{eq:f-GAN-lower}
\end{equation}
where $\mathcal{T}$ is an arbitrary function class of $T$ that satisfies $\mathcal{X} \to \mathbb{R}$ (e.g., parameterized discriminator with a specific activation function such as sigmoid). The derivation above yields a lower bound for $D_f(p_r\|p_g)$ that is tractable, thus this can be directly calculated. The optimization for $f$-GAN firstly is characterized by maximizing the lower bound (last line in equation~\eqref{eq:f-GAN-lower}) with respect to discriminator, which aims to make the lower bound to be the estimation of $f$-divergence, and then minimizes the $f$-divergence regarding the generator in order to make $p_g$ close to $p_r$. This optimization is known as variational divergence minimization (VDM). The authors train generative neural samplers based on VDM on MNIST ($28\times 28$ pixel images) and LSUN ($96\times 96$ pixel images). The model architecture and training settings are the same as proposed in DCGAN.

\subsection{Unrolled GAN (UGAN)}
\begin{figure*}[!ht]
    \centering
    \includegraphics[width=.9\textwidth]{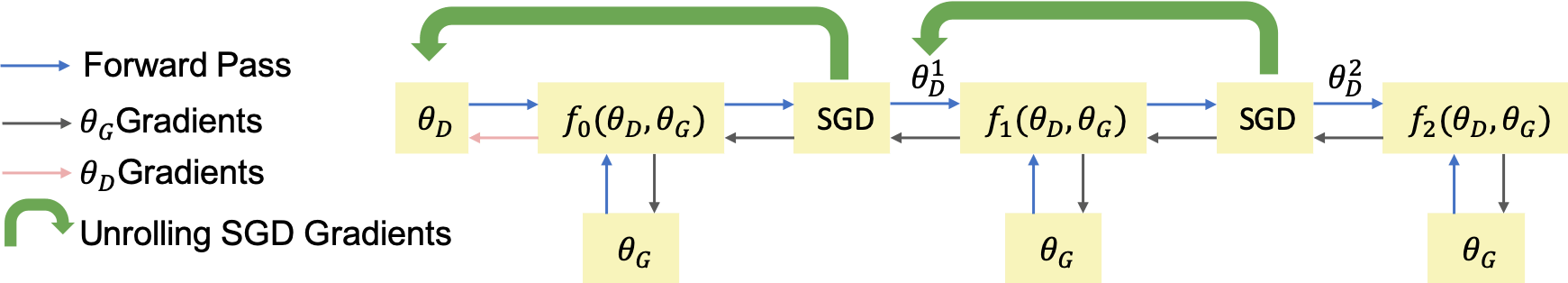}
    \caption{An example of computation for an unrolled GAN with 3 unrolling steps. $G$ and $D$ update using equation~\eqref{chap05-eq:UGAN-update}. Each step $k$ uses the gradients of $f_k$ regarding $\bm{\mathrm{\theta}}_D^k$ stated in the equation~\eqref{chap05-eq:UGAN-theta-update}. }
    \label{chap05-fig:UGAN_update}
\end{figure*}

UGAN is a design proposed to solve the problem of mode collapse for GANs during training~\cite{metz2016unrolled}. The core design innovation of UGAN is the addition of a gradient term for updating $G$, which has an ability of capturing responses of the discriminator to a change in the generator. The optimal parameter for $D$ can be expressed as an iterative optimization procedure as below
\begin{equation}
\begin{aligned}
    \bm{\mathrm{\theta}}_D^0 =&\bm{\mathrm{\theta}}_D,\\
    \bm{\mathrm{\theta}}_D^{k+1}=&\bm{\mathrm{\theta}}_D^k+\eta^k\frac{\mathrm{d}f(\bm{\mathrm{\theta}}_G,\bm{\mathrm{\theta}}_D^k)}{\mathrm{d}\bm{\mathrm{\theta}}_D^k},\\
    \bm{\mathrm{\theta}}_D^*(\bm{\mathrm{\theta}}_G)=&\lim_{k \to \infty}\bm{\mathrm{\theta}}_D^k,
\end{aligned}
\label{chap05-eq:UGAN-theta-update}
\end{equation}
where $\eta^k$ is the learning rate, $\bm{\mathrm{\theta}}_D$ represents parameters for $D$ and $\bm{\mathrm{\theta}}_G$ represents parameters for $G$. The surrogate loss by unrolling for $K$ steps can be expressed as
\begin{equation}
    f_{K}(\bm{\mathrm{\theta}}_G,\bm{\mathrm{\theta}}_D)=f(\bm{\mathrm{\theta}}_G,\bm{\mathrm{\theta}}_D^K(\bm{\mathrm{\theta}}_G,\bm{\mathrm{\theta}}_D)).
    \label{chap05-eq:UGAN-G-loss}
\end{equation}
This surrogate loss is then used for updating parameters for $D$ and $G$
\begin{equation}
\begin{aligned}
    &\bm{\mathrm{\theta}}_G \gets \bm{\mathrm{\theta}}_G - \eta \frac{\mathrm{d}f_K(\bm{\mathrm{\theta}}_G,\bm{\mathrm{\theta}}_D)}{\mathrm{d}\bm{\mathrm{\theta}}_G},\\
    &\bm{\mathrm{\theta}}_D \gets \bm{\mathrm{\theta}}_D + \eta \frac{\mathrm{d}f(\bm{\mathrm{\theta}}_G,\bm{\mathrm{\theta}}_D)}{\mathrm{d}\bm{\mathrm{\theta}}_D}    
    \label{chap05-eq:UGAN-update}
\end{aligned}
\end{equation}
Figure~\ref{chap05-fig:UGAN_update}
illustrates the computational diagram for an unrolled GAN with three unrolling steps. Equation~\eqref{chap05-eq:UGAN-G-gradient} illustrates the gradient for updating $G$. 
\begin{equation}
\begin{aligned}
    &\frac{\mathrm{d}f_{K}(\bm{\mathrm{\theta}}_G,\bm{\mathrm{\theta}}_D)}{\mathrm{d}\bm{\mathrm{\theta}}_G}=\frac{\partial f(\bm{\mathrm{\theta}}_G,\bm{\mathrm{\theta}}^K_D(\bm{\mathrm{\theta}}_G,\bm{\mathrm{\theta}}_D))}{\bm{\mathrm{\theta}}_G}+\frac{\partial f(\bm{\mathrm{\theta}}_G,\bm{\mathrm{\theta}}^K_D(\bm{\mathrm{\theta}}_G,\bm{\mathrm{\theta}}_D))}{\partial \bm{\mathrm{\theta}}^K_D(\bm{\mathrm{\theta}}_G,\bm{\mathrm{\theta}}_D)}\frac{\mathrm{d}\bm{\mathrm{\theta}}^K_D(\bm{\mathrm{\theta}}_G,\bm{\mathrm{\theta}}_D)}{\mathrm{d}\bm{\mathrm{\theta}}_G}
    \label{chap05-eq:UGAN-G-gradient}
\end{aligned}
\end{equation}
It should be noted that the first term in equation~\eqref{chap05-eq:UGAN-G-gradient} is the gradient for the original GAN. The second term here reflects how $D$ reacts to changes in $G$. If $G$ tends to collapse to one mode, $D$ will increase the loss for $G$. Thus, this unrolled approach is able to prevent the mode collapse problem for GANs. The author train UGAN on MNIST and CIFAR10 datasets. All convolutions have kernel size of $3\times 3$ with batch normalization. The discriminator used leaky ReLU with a 0.3 leakness and the generator uses ReLU. The generator consisted of 5 layers, which are fully connected layer, 3 deconvolutional layers and 1 convolutional layer. The discriminator had 4 layers, which are 3 convolutional layers and 1 fully connected layer. The Adam optimizer with a generator learning rate of $1e-4$ and a discriminator learning rate as $2e-4$ was utilized in the experiment.

\subsection{Loss Sensitive GAN (LS-GAN)}
\begin{figure}[!ht]
    \centering
    \includegraphics[width=.7\textwidth]{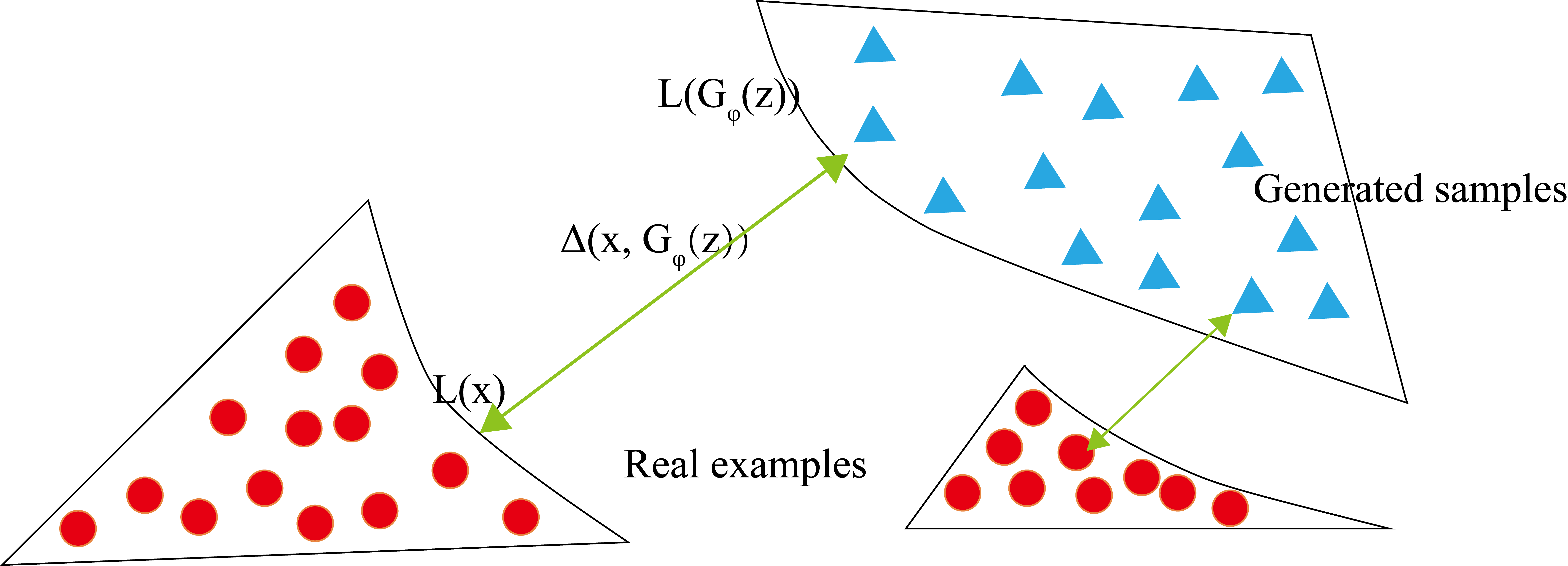}
    \caption{Demonstration of the loss in equation~\eqref{chap05-eq:LS-GAN-loss}. $\Delta(\mathbf{x},G(\mathbf{z}))$ is used to separate real samples and generated samples. If some generated samples are close enough to the real samples, LS-GAN will focus on other generated samples that are far away from the real samples. This optimization loss puts restriction on $D$ to prevent it from separating generated and real samples too well. Thus, it solves the vanishing gradient problem which arises in original GANs. ($G_\phi(\mathbf{z})$ here is equivalent to $G(\mathbf{z})$ where $\phi$ represents the parameters for generator). Figure regenerated from~\cite{qi2017loss}.}
    \label{chap05-fig:LS-GAN-loss-demo}
\end{figure}

LS-GAN was introduced to train the generator to produce realistic samples by minimizing the designated margins between real and generated samples~\cite{qi2017loss}. This work argues that the problems such as the vanishing gradient and mode collapse as appearing in the original GAN is caused by a non-parametric hypothesis that the discriminator is able to distinguish any type of probability distribution between real samples and generated samples. As mentioned before, it is very normal for the overlap between the real samples distribution and the generated samples distribution to be negligible. Moreover, $D$ is also able to separate real samples and generated samples. The JS divergence will become a constant under this situation, where the vanishing gradient arises for $G$. In LS-GAN, the classification ability of $D$ is restricted and is learned by a loss function $L_\theta(\mathbf{x})$ parameterized with $\theta$, which assumed that a real sample ought to have smaller loss than a generated sample. The loss function can be trained as the following constraint
\begin{equation}
    L_\theta(\mathbf{x})\leq L_\theta(G(\mathbf{z})) - \Delta(\mathbf{x},G(\mathbf{z})), 
\end{equation}
where $\Delta(\mathbf{x},G(\mathbf{z}))$ is the margin measuring the difference between real samples and generated samples. This constraint indicates that a real sample is separated from a generated sample by at least a margin of $\Delta(\mathbf{x},G(\mathbf{z}))$. The optimization for the LS-GAN is then stated as
\begin{equation}
\begin{aligned}
    &\min_D\mathcal{L}_D = \mathbb{E}_{\mathbf{x}\sim p_r}L_\theta(\mathbf{x})+\lambda \mathbb{E}_{\substack{\mathbf{x}\sim p_r\\ \mathbf{z}\sim p_z}}(\Delta(\mathbf{x},G(\mathbf{z}))+L_\theta(\mathbf{x})-L_\theta(G(\mathbf{z})))_+,\\
    &\min_G\mathcal{L}_G = \mathbb{E}_{\mathbf{z}\sim p_z}L_{\theta}(G(\mathbf{z})),
    \label{chap05-eq:LS-GAN-loss}
\end{aligned}
\end{equation}
where $\lambda$ is a positive balancing parameter, $(a)_+=\max(a,0)$ and $\bm\theta$ are the parameters in $D$. From the second term in $\mathcal{L}_D$ in the equation~\eqref{chap05-eq:LS-GAN-loss}, $\Delta(\mathbf{x},G(\mathbf{z}))$ is added as a regularization term for optimizing $D$ in order to prevent $D$ from overfitting the real samples and the generated samples. Figure~\ref{chap05-fig:LS-GAN-loss-demo}
demonstrates the efficacy of equation~\eqref{chap05-eq:LS-GAN-loss}. The loss for $D$ puts a restriction on the ability of $D$ i.e., It challenges the ability of $D$ for to separate well generated samples from real samples, which is the original cause for the vanishing gradient. More formally, LS-GAN assumes that $p_r$ lies in a set of Lipschitz densities with a compact support. 

The models were trained CIFAR-10, SVHN~\cite{netzer2011reading} and CelebA with a mini-batch of 64 images. All weights were initialized from a zero-mean Gaussian distribution with a standard deviation of 0.02. The Adam optimizer was used to train the model with initial learning rate as $1e-3$ and $\beta_1$ as 0.5 while the learning rate was decreased every 25 epochs by a factor of 0.8.

\subsection{Mode Regularized GAN (MRGAN)}
MRGAN proposes a metric regularization to penalize missing modes~\cite{che2016mode}, which is then used to solve the mode collapse problem. The key idea behind this work is the use of an encoder $E(\mathbf{x})$: $\mathbf{x}\to \mathbf{z}$ to produce the latent variable $\mathbf{z}$ for $G$ instead of using noise. This procedure has two benefits: (1) The encoder reconstruction can add more information to $G$ so that is not that easy for $D$ to distinguish between generated samples and real samples; and (2) the encoder ensures correspondence between $\mathbf{x}$ and $\mathbf{z}$ ($E(\mathbf{x})$), which means $G$ can cover different modes in the $\mathbf{x}$ space. So it prevents the mode collapse problem. The loss function for this mode regularized GAN is
\begin{equation}
\begin{aligned}
    &\mathcal{L}_G=-\mathbb{E}_\mathbf{z}[\mathrm{log}[D(G(\mathbf{z}))]]+\mathbb{E}_{\mathbf{x}\sim p_r}[\lambda_1 d(\mathbf{x},G\circ E(\mathbf{x}))+\lambda_2 \mathrm{log}[D(G(\mathbf{x}))]], \\
    &\mathcal{L}_E=\mathbb{E}_{\mathbf{x}\sim p_r}[\lambda_1 d(\mathbf{x},G\circ E(\mathbf{x}))+\lambda_2 \mathrm{log}[D(G(\mathbf{x}))]],
\end{aligned}
\end{equation}
where $d$ is a geometric measurement which can be chosen from many options e.g., pixel-wise $L^2$ and distance of extracted features. The authors evaluate the performance of including the mode regularization on the MNIST, CelebA ($64\times 64$) datasets.

\subsection{Geometric GAN}
Geometric GAN~\cite{lim2017geometric}
\begin{figure}[!ht]
    \centering
    \includegraphics[width=.6\textwidth]{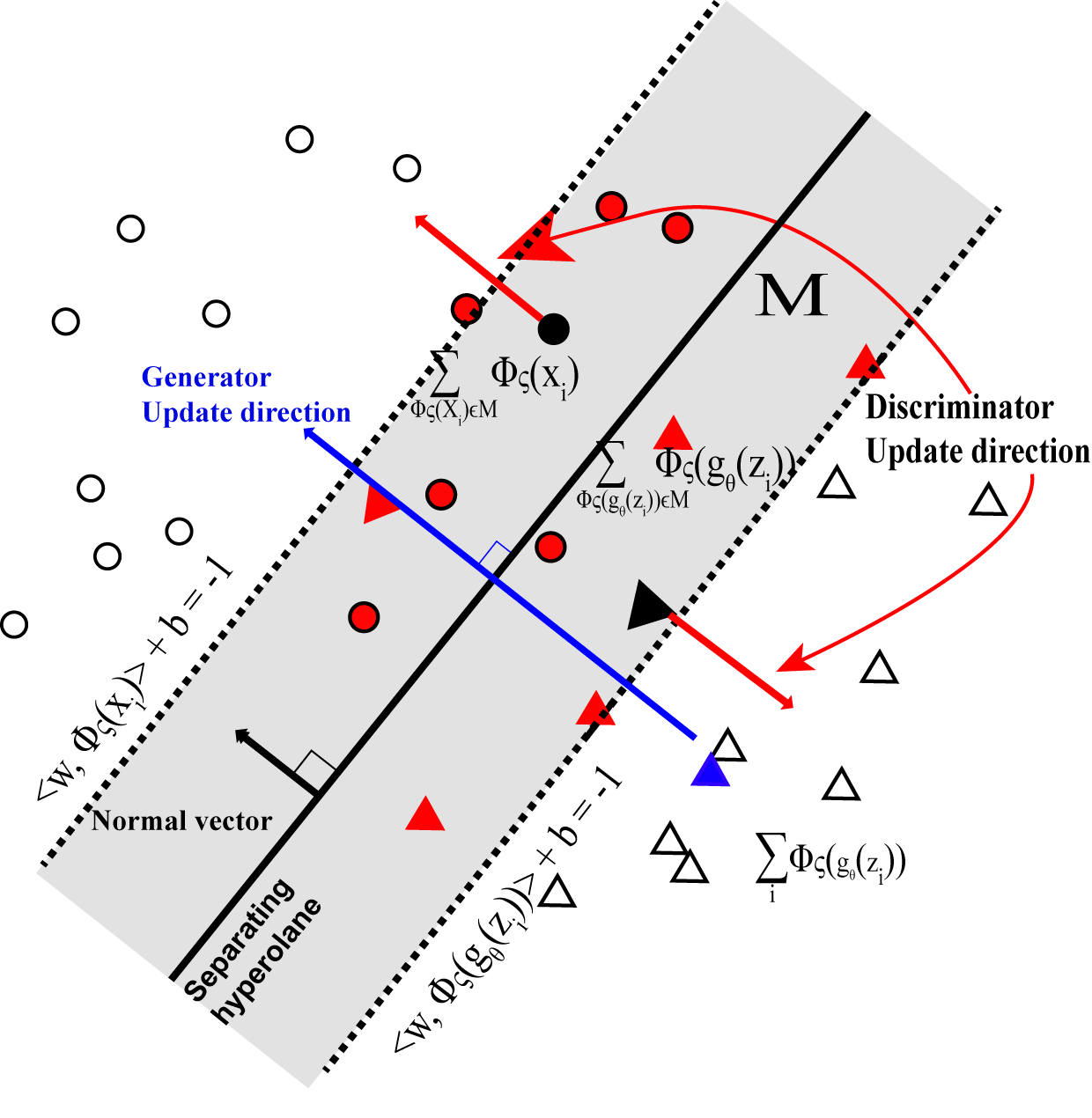}
    \caption{SVM hyperplane used in Geometric GAN. Discriminator updates by pushing real data samples and generated data samples away from the hyperplane while generator updates by pushing generated data samples towards the hyperplane. Figure regenerated from~\cite{lim2017geometric}.}
    \label{chap05-fig:Geo-GAN-hyperplane}
\end{figure}
is proposed using SVM separating hyperplane, which has the maximal margins between the two classes. Figure~\ref{chap05-fig:Geo-GAN-hyperplane} demonstrates the update rule for the discriminator and generator based on the SVM hyperplane. The loss function of geometric GAN can be derived as an alternative fashion by minimizing the hinge loss as
\begin{equation}
\begin{aligned}
    &\mathcal{L}_D = -\mathbb{E}_{(x,y)\sim p_r}[min(0, -1+D(x,y))] - \mathbb{E}_{z\sim p_z, y\sim p_r}[min(0,-1-D(G(z),y))],\\
    &\mathcal{L}_G = -\mathbb{E}_{z\sim p_z, y\sim p_r}D(G(z),y),
\end{aligned}
\label{eq:GGAN_loss}
\end{equation}
This hinge loss fashion is also deployed in the SAGAN mentioned in section~\ref{sec:SAGAN} and BigGAN mentioned in section~\ref{sec:BigGAN}. Compared to the other loss functions, the authors demonstrate the efficacy of hinge loss for dealing with the high-dimension low-sample size (HDLSS) problem~\cite{marron2007distance,carmichael2017geometric,ahn2010maximal}, which is a classification problem caused by the mini-batch size is much smaller than the dimension of the feature space. In this paper, the geometric GAN is designed based on soft-margin SVM linear classifier rather than hard-margin SVM linear classifier.

The networks were trained on MNIST ($64\times 64$ resolution), CelebA ($64\times 64$ resolution) and LSUN ($64\times 64$ resolution) datasets. The DCGAN architecture trained by using RMSprop optimizer with learning rate $2e-4$ and mini-batch size 64 was deployed in this work. The authors demonstrate that geometric GAN is more stable for training and less prone to mode collapse.

\subsection{Relativistic GAN (RGAN)}
\begin{figure}[!ht]
    \centering
    \includegraphics[width=.7\textwidth]{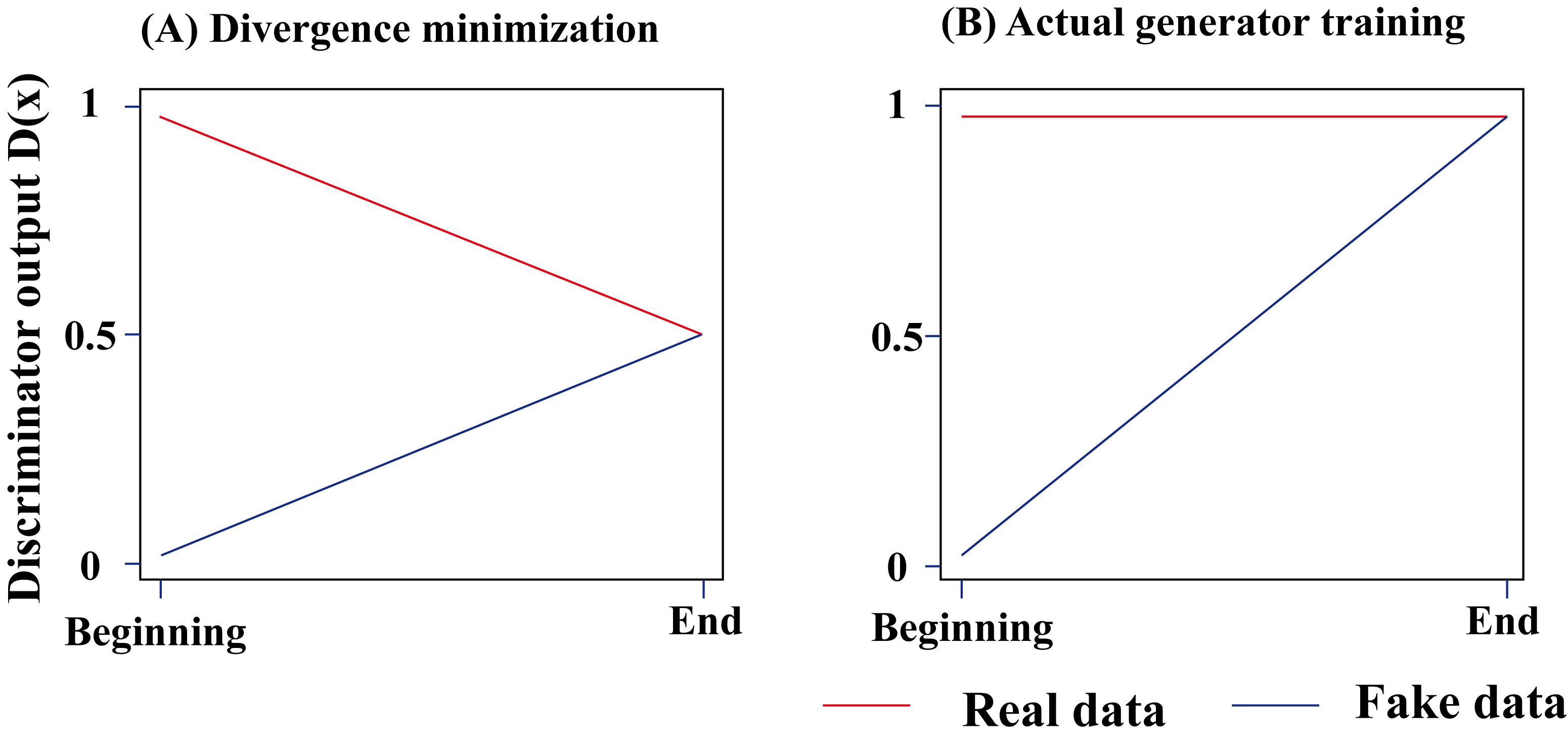}
    \caption{$D$ output comparison between RGAN and original GAN. (a) $D$ output in RGAN; (b) $D$ output in original GAN when training the $G$. Figure regenerated from~\cite{jolicoeur2018relativistic}.}
    \label{chap05-fig:RGAN-loss-demo}
\end{figure}

RGAN~\cite{jolicoeur2018relativistic} is proposed as a general approach to devising new cost functions from the existing one i.e., it can be generalized for all integral probability metric (IPM)~\cite{sriperumbudur2009integral,muller1997integral} GANs. The discriminator in the original GAN measures \textit{the probability for a given real sample or a generated sample}. The author argues that key relative discriminant information between real data and generated data is missing in original GAN. The discriminator in RGAN takes into account that \textit{how a given real sample is more realistic compared to a given random generated sample}. Loss function of RGAN applied to original GAN is stated as
\begin{equation}
\begin{aligned}
    &\min_D \mathbb{E}_{\substack{\mathbf{x}_r\sim p_r\\\mathbf{x}_g\sim p_g}}[\mathrm{log}(\mathrm{sigmoid}(C(\mathbf{x}_r)-C(\mathbf{x}_g)))],\\
    &\min_G \mathbb{E}_{\substack{\mathbf{x}_r\sim p_r\\\mathbf{x}_g\sim p_g}}[\mathrm{log}(\mathrm{sigmoid}(C(\mathbf{x}_g)-C(\mathbf{x}_r)))],
\end{aligned}
\end{equation}
where $C(\mathbf{x})$ is the non-transformed layer. Figure~\ref{chap05-fig:RGAN-loss-demo} 
demonstrates the effect on $D$ of using the RGAN approach compared to the original GAN. In terms of the original GAN, the optimization aims to push the $D(\mathbf{x})$ to 1 (right one). For RGAN, the optimization aims to push $D(\mathbf{x})$ to 0.5 (left one), which is more stable compared to the original GAN. The author also claims that RGAN can be generalized to other types of loss-variant GANs if those loss functions belong to IPMs. The generalization loss is stated as 
\begin{equation}
\begin{aligned}
    &\mathcal{L}_D=\mathbb{E}_{\substack{\mathbf{x}_r\sim p_r\\\mathbf{x}_g\sim p_g}}[f_1(C(x_r)-C(x_g))] + \mathbb{E}_{\substack{\mathbf{x}_r\sim p_r\\\mathbf{x}_g\sim p_g}}[f_2(C(x_g)-C(x_r))],\\
    &\mathcal{L}_G=\mathbb{E}_{\substack{\mathbf{x}_r\sim p_r\\\mathbf{x}_g\sim p_g}}[g_1(C(x_r)-C(x_g))] +\mathbb{E}_{\substack{\mathbf{x}_r\sim p_r\\\mathbf{x}_g\sim p_g}}[g_2(C(x_g)-C(x_r))],
\end{aligned}
\end{equation}
where $f_1(y)=g_2(y)=-y$ and $f_2(y)=g_1(y)=y$. Details of loss generalization for other GANs refers to the original paper~\cite{jolicoeur2018relativistic}.

The authors trained the networks on the CIFAR-10 and the CAT dataset with various image sizes i.e., $64\times 64$, $128\times 128$ and $256\times 256$. The DCGAN architecture with the Adam optimizer was used. Various training settings have been explored and more details can be referred to the original paper~\cite{jolicoeur2018relativistic}. The author successfully demonstrates that the relativistic discriminator offers a way to fix and improve on standard GAN and is able to achieve better performance with other tricks e.g., spectral normalization, gradient penalty. More importantly, the author demonstrates the generability of this approach, in which any type of GAN can be trained through a RGAN fashion.

\subsection{Spectral normalization GAN (SN-GAN)}
SN-GAN~\cite{miyato2018spectral} proposes the use of weight normalization to train the discriminator more stably. This technique is computationally light and easily applied to existing GANs. Previous work for stabilizing the training of GANs ~\cite{qi2017loss,arjovsky2017wasserstein,gulrajani2017improved} emphasizes the importance that $D$ should be from the set of K-Lipshitz continuous functions. Popularly speaking, Lipschitz continuity~\cite{donchev1998stability,armijo1966minimization,goldstein1977optimization} is more strict than the continuity, which describes that the function does not change rapidly. This smooth $D$ is of benefit in stabilizing the training of GANs. The work mentioned previously focused on the control of the Lipschitz constant of the discriminator function. This work demonstrates an alternative simpler way to control the Lipschitz constant through spectral normalization of each layer for $D$. Spectral normalization is performed as
\begin{equation}
    \mathbf{\bar{W}}_{SN}(\mathbf{W})=\frac{\mathbf{W}}{\sigma({\mathbf{W}})},
    \label{chap05-eq:spectral-norm}
\end{equation}
where $\mathbf{W}$ represents weights on each layer for $D$ and $\sigma({\mathbf{W}})$ is the $L_2$ matrix norm of $\mathbf{W}$. The paper proves this will make $\|f \|\leq 1$. The fast approximation for the $\sigma({\mathbf{W}})$ is also demonstrated in the original paper.

The authors evaluated the performance of SN-GAN on the CIFAR-10 ($32\times 32$ resolution), the STL-10 ($48\times 48$ resolution)~\cite{coates2011analysis} and the ImageNet ($128\times 128$ resolution) by comparing to the existing regularization/normalization techniques including weight clipping~\cite{arjovsky2017wasserstein}, gradient penalty~\cite{wu2017gp}, batch normalization~\cite{ioffe2015batch}, weight normalization~\cite{salimans2016weight}, layer normalization~\cite{ba2016layer} and orthonormal regularization~\cite{brock2016neural}. Several training settings has been carried out for a comprehensive comparison. The authors demonstrate the efficacy of spectral normalization on the diversity and the quality of generated images compared to previously proposed approaches.

\subsection{RealnessGAN}
Currently, the discriminator in GANs can only output 0 and 1 i.e., real and fake instead of a continuous distribution as the measure of realness. Xiangli \textit{et al.}~\cite{xiangli2020real} propose the RealnessGAN to tackle this new perspective, which treats realness as a random variable that can be estimated from multiple angles. Traditional GANs adopt a single scalar (discriminator output) as the measure of realness. The authors argue that the realness is more complicated and covers multiple factors such as texture and overall configuration in the case of images. Following this observation, the discriminator is re-designed to learn a realness distribution instead of a single scalar. To achieve this, RealnessGAN replaces the single scalar by a distribution $p_{\mathrm{realness}}$ so that $D(\mathbf{x})=\{p_{\mathrm{realness}}(\mathbf{x}, u); u\in \Omega\}$ given by an input sample, where $\Omega$ is the set of outcomes of $p_{\mathrm{realness}}$ and each outcome $u$ can be viewed as a potential realness measure by a chosen realness measuring criteria. In the original paper, the discriminator returns $N$ probabilities on these $N$ outcomes $\Omega = \{u_0, u_1, \cdots, u_{N-1}\}$ as
\begin{equation}
    p_{\mathrm{realness}(\mathbf{x}, u_i)} = \frac{e^{\theta_i(\mathbf{x})}}{\sum_j e^{\theta_j(\mathbf{x})}}
    \label{eq:realnessGAN-outcomes}
\end{equation}
where $\theta = \{\theta_0, \theta_1, \cdots, \theta_{N-1}\}$ are the parameters of $D$. Apart from the outcomes $\Omega$, two distributions $\mathcal{A}_1$ for real and $\mathcal{A}_0$ for fake are also defined on $\Omega$. In practical implementation, given a mini-batch $\{x_0, x_1, \cdots, x_{m-1}\}$ i.e., logits computed by the discriminator on the $i$-th outcome, a Gaussian distribution $\mathcal{N}(\mu_i, \sigma_i)$ is fitted on $\{\theta_i(\mathbf{x}_0), \theta_i(\mathbf{x}_1), \cdots, \theta_i(\mathbf{x}_{m-1})\}$ and new logits is re-computed as $\{\theta^{'}_i(\mathbf{x}_0), \theta^{'}_i(\mathbf{x}_1), \cdots, \theta^{'}_i(\mathbf{x}_{m-1}); \theta^{'}_i \sim \mathcal{N}(\mu_i, \sigma_i)\}$. Increasing number of outcomes will make $D$ more rigorous and put more constraints on $G$. In another word, larger number of outcomes is suggested for a more complicated dataset. The minmax loss can finally be represented as
\begin{equation}
	\min \limits_{G} \max\limits_{D}\hspace{2pt} \mathbb{E}_{\bm{\mathrm{x}} \sim p_{r}} \mathrm{KL}(\mathcal{A}_1\|D(\bm{\mathrm{x}})) + \mathbb{E}_{\bm{\mathrm{z}} \sim p_{\bm{\mathrm{z}}}} \mathrm{KL}(\mathcal{A}_0\|D(G(\bm{\mathrm{z}}))), \hspace{5pt}  \text{KL refers to KL divergence}
	\label{eq:realnessGAN-loss}
\end{equation}

The authors trained RealnessGAN on CIFAR10, and CelebA by using the Adam optimizer. The network architecture of RealnessGAN is identical to the DCGAN architecture with $\mathbf{z} \sim \mathcal{N}(0, I)$. Batch normalization is deployed for $G$ and spectrum normalization is applied for $D$. The number of outcomes are set to 51 for CelebA and 3 for CIFAR10 datasets respectively.  

\subsection{Sphere GAN}
\label{Sphere-GAN}
Sphere GAN~\cite{Park_2019_CVPR} is a novel integral probability metric (IPM)-based GAN, which uses the hypersphere to bound IPMs in the objective function, thereby it can enable stable training. Via exploiting the information of higher-order statistics of data using geometric moment matching, the GAN model can provide more accurate results. The objective function of sphere GAN is defined as
\begin{equation}
 \min _{G} \max _{D} \sum_{r} E_{x}\left[d_{s}^{r}(\mathbf{N}), D(x)\right]-\sum_{r} E_{z}\left[d_{s}^{r}(\mathbf{N}, D(G(z)))\right].  
\end{equation}

\begin{figure}[htbp]
    \centering
    \includegraphics[width=.8\textwidth]{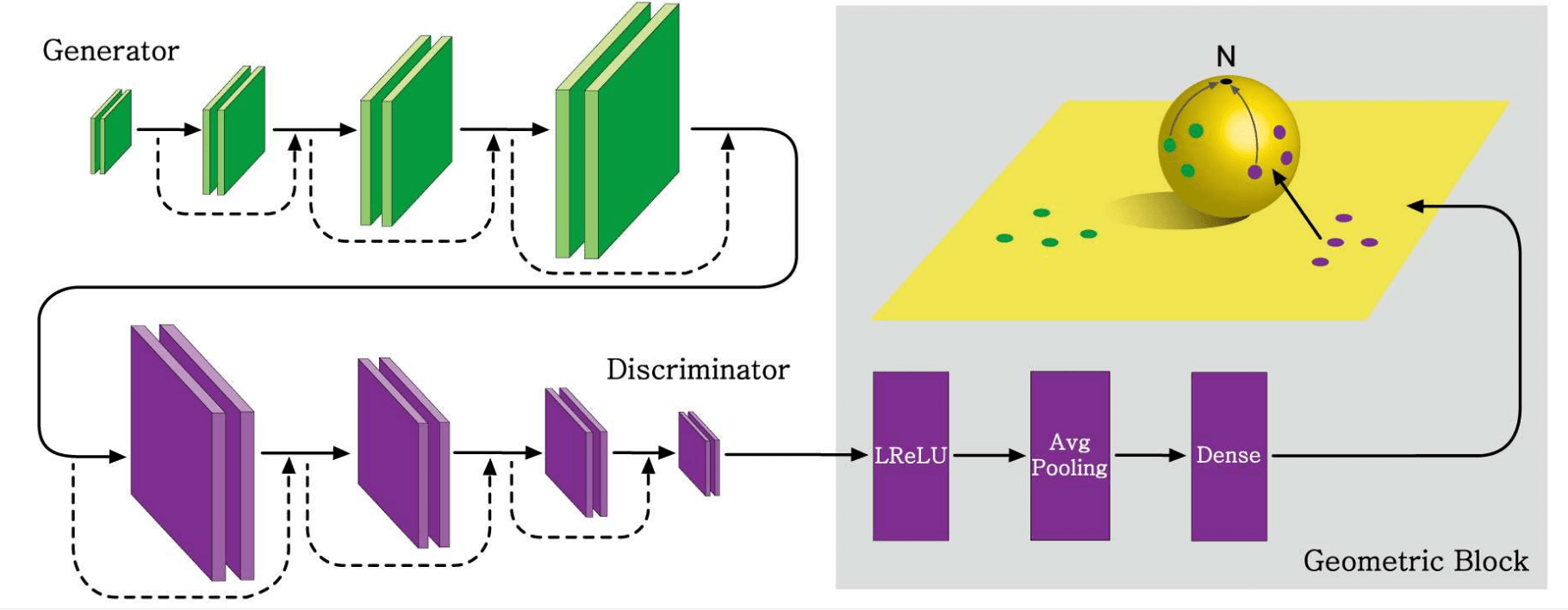}
    \caption{Pipeline of sphere GAN. The generator outputs the real and fake data (generated via a noisy inputs) into the discriminator. The final output is an $n-$dimensional Euclidean feature space (yellow plane). The green and purple circles are the feature points of the fake and real samples. \textbf{The key idea of sphere GAN is remapping the feature points into the $n-$dimensional hypersphere} (i.e., yellow sphere). After the geometry transformation, the mapped points can be used for calculating the geometric moments centered at the north pole of the hypersphere. At the same time, the discriminator tries to maximize the moment differences of real and fake samples, while the generator tries to interfere with the discriminator by minimizing the moment differences. Figure regenerated from~\cite{Park_2019_CVPR}.}
    \label{chap05-fig:sphereGAN-architecture}
\end{figure}

For $r=1, \cdots, R$ where the function $d_{s}^{r}$ measures the $r$ -th moment distance between each sample and the north pole of the hypersphere, $\mathbf{N}$. Note that the subscript $s$ indicates that $d_{s}^{r}$ is defined on $\mathbb{S}^{n}$. Different from the conventional discriminators based on the Wesserstein distance require Lipschitz constraints, which forces the discriminators to be a member of 1-Lipschitz functions. The sphere GAN alleviate the constraints by defining IPMs on the hypersphere. Figure~\ref{chap05-fig:sphereGAN-architecture} shows the pipeline of sphere GAN.

Unlike in conventional approaches like WGAN-GP, WGAN-CT, and WGANL, sphere GAN does not need any additional constraints that forces discriminators to lie in a desired function space. By using geometric transformation, sphere GAN ensures that distance functions lie in a desired function space with no additional constraint term. 

\subsection{Self-supervised GAN (SS-GAN)}
\label{self-supervised GAN}
Although the conditional GAN has achieved great success in natural image synthesis. The main drawback of conditional GANs is the necessity for labeled data. Self-Supervised GANs~\cite{Chen_2019_CVPR} exploit adversarial training and \textbf{self-supervision} for bridging the gap between conditional and unconditional GANs. 

This work imbues the discriminator with a mechanism to learn useful representations, independently of the quality of the current generator. In a self-supervised manner, they train a model on predicting rotation angle for extracting representations from the resulting networks, and then propose to add a self-supervised task (a rotation-based loss) to the discriminator, as
\begin{align}
L_{G}&=-V(G, D)-\alpha \mathbb{E}_{\boldsymbol{x} \sim P_{G}} \mathbb{E}_{r \sim \mathcal{R}}\left[\log Q_{D}\left(R=r | \boldsymbol{x}^{r}\right)\right] \\
L_{D}&=V(G, D)-\beta \mathbb{E}_{\boldsymbol{x} \sim P_{\text {data }}} \mathbb{E}_{r \sim \mathcal{R}}\left[\log Q_{D}\left(R=r | \boldsymbol{x}^{r}\right)\right]
\end{align}
where $V(G, D)$ is the original value function~\cite{goodfellow2014generative}, $r \in \mathcal{R}$ is a rotation selected from a set of possible rotations ($\mathcal{R}=\left\{0^{\circ}, 90^{\circ}, 180^{\circ}, 270^{\circ}\right\}$). Image $\boldsymbol{x}$ rotated by $r$ degrees is denoted as $\boldsymbol{x}^{r},$ and $Q\left(R | \boldsymbol{x}^{r}\right)$ is the discriminator's predictive distribution over the angles of rotation of the sample. The implementation trick is they use output of the second last layer of discriminator added with a linear layer to predict the rotation type. This work tries to enforce the discriminator to learn good representation via learning the rotation information. 

\subsection{\textbf{Summary}}
\begin{figure}[!ht]
    \centering
    \includegraphics[width=1.\textwidth]{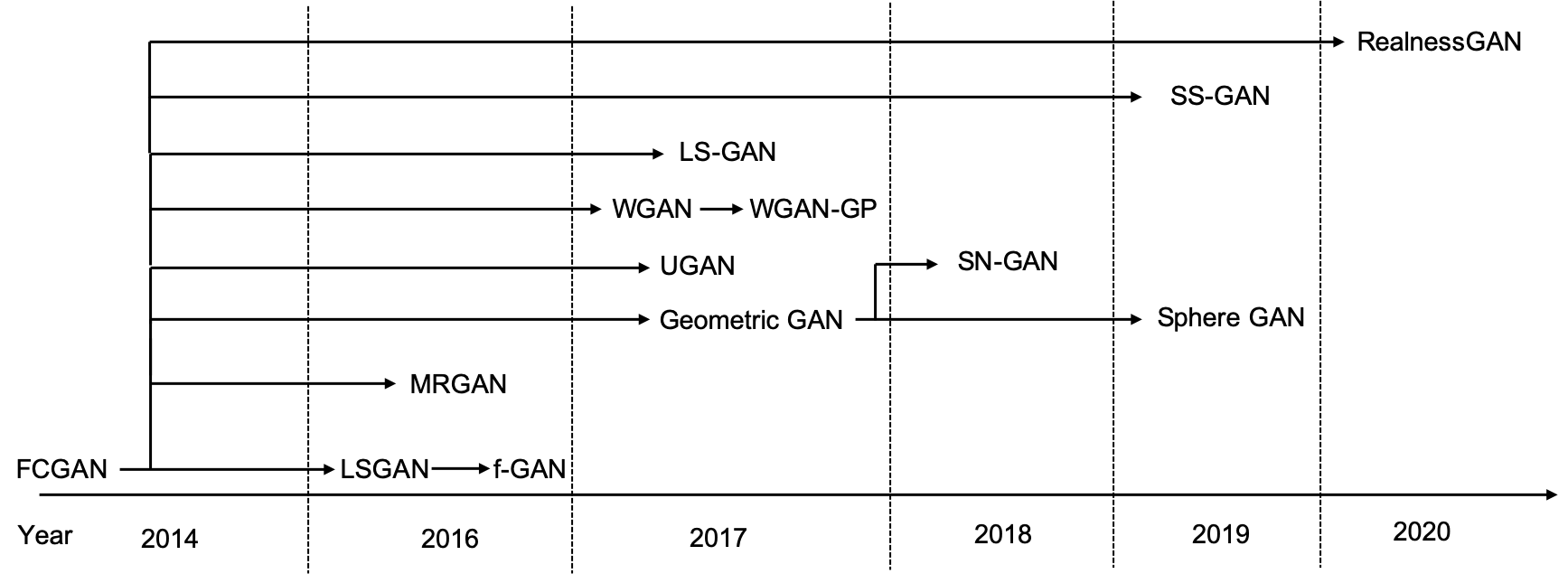}
    \caption{An overview of the footprint for loss-variant GANs discussed in this section.}
    \label{fig:loss_roadmap}
\end{figure}

\begin{figure}[!ht]
    \centering
    \includegraphics[width=.7\textwidth]{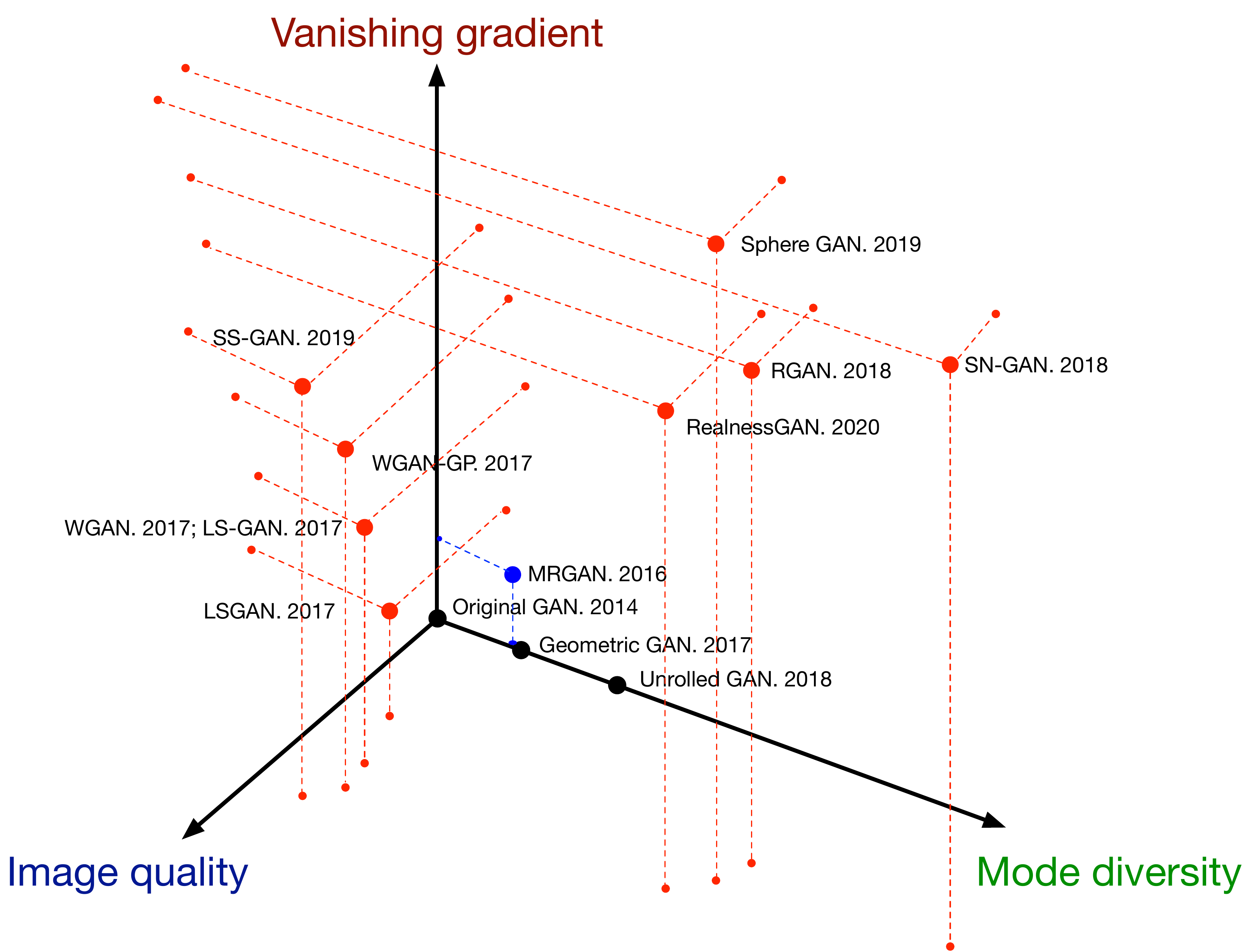}
    \caption{Current loss-variants for solving the challenges. Challenges are categorized in terms of three independent axes. Red points indicate the GAN-variant covers all three challenges, blue points cover two, and black points cover only one challenge. Larger value for each axis indicates better performance.}
    \label{chap05-fig:loss_three_dimension}
\end{figure}

\begin{figure}[ht!]
	\centering
	\subfigure{\includegraphics[width=.4\textwidth]{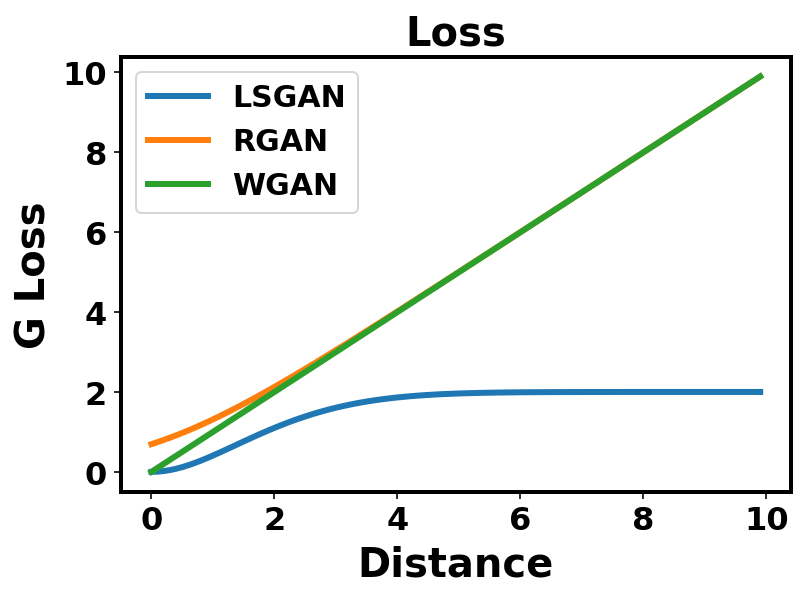}
		\includegraphics[width=.405\textwidth]{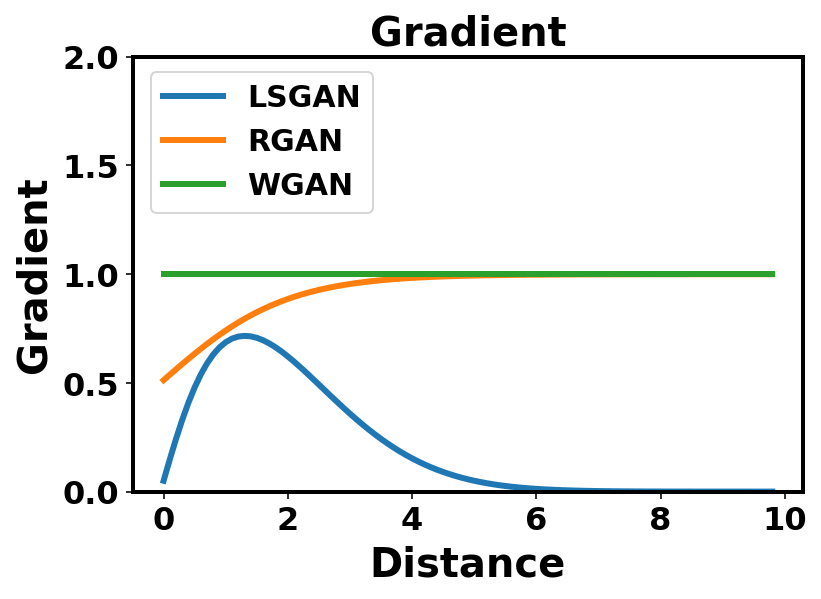}}
	
	\caption{Loss and gradient for generator of different loss-variant GANs.}
	\label{chap05-fig:G-loss-gradient}
\end{figure}

We explain the training problems (mode collapse and vanishing gradient for $G$) in the original GAN and we have introduced loss-variant GANs in the literature, which are mainly proposed for improving the performance of GANs in terms of three key aspects. Figure~\ref{fig:loss_roadmap}
illustrates the footprint of loss-variants that discussed in this section.
Figure~\ref{chap05-fig:loss_three_dimension}
summarizes the efficacy of loss-variant GANs for the challenges. More details of quantitative results are provided in section~\ref{sec:Notes_on_evaluation}. Losses of LSGAN, RGAN and WGAN are very similar to the original GAN loss. We use a toy example (i.e., two distributions used in Fig.~\ref{chap05-fig:JSD-pg-pr}) to demonstrate the $G$ loss regarding the distance between real data distribution and generated data distribution in Fig.~\ref{chap05-fig:G-loss-gradient}. 
It can be seen that RGAN and WGAN are able to inherently solve the vanishing gradient problems for the generator when the discriminator is optimized. LSGAN on the contrary still suffers from a vanishing gradient for the generator, however, it is able to provide a better gradient compared to the original GAN in Fig.~\ref{chap05-fig:JS-curve-summary} when the distance between the real data distribution and the generated data distribution is relatively small. This is demonstrated in the original paper~\cite{mao2017least} where LSGAN is shown to be better easier to push generated samples to the boundary made by discriminator.

Table~\ref{chap05-tab:GAN-loss_sum}
\begin{table*}[!ht]
    \centering
    \footnotesize
    \begin{tabularx}{\linewidth}{l|X|X}
        \toprule
        {\textbf{GAN type}} & {\textbf{Pros}} & {\textbf{Cons}} \\ \hline
        
        {FCGAN~\cite{goodfellow2014generative}, 2014} & {1) Generates samples very fast. 2) Able to deal with a sharp probability distribution. } & {1) Vanishing gradient for $G$. 2) Mode collapse. 3) Resolution of generated images is very low. } \\ \hline

        {MRGAN~\cite{che2016mode}, 2016} & {1) Improves mode diversity. 2) Stabilizes GAN training.}&{1) Generated image quality is low. 2) Does not solve vanishing gradient problem for $G$. 3) Only tested on CelebA dataset. Has not been tested on more diverse image datasets e.g., CIFAR and ImageNet.} \\ \hline
        
        {f-GAN~\cite{nowozin2016f}, 2016} & {1) Provides a unified framework based on $f$-divergence.}&{1) Has not specified stability for different $f$-divergence functions.} \\ \hline
        
        {WGAN~\cite{arjovsky2017wasserstein}, 2017} & {1) Solves the vanishing gradient problem. 2) Improves image quality. 3) Solves the mode collapse problem.}&{1) Large weight clipping causes longer time of convergence and small weights clipping causes a vanishing gradient. 2) Weight clipping reduces the capacity of the model, which limits the model's capability to learn more complex functions. 3) A very deep WGAN does not converge easily.} \\ \hline
        
        {WGAN-GP~\cite{gulrajani2017improved}, 2017} & {1) Converges much faster than WGAN. 2) Model is more stable during training. 3) Able to use deeper GAN to model more complex function.}&{1) Cannot use batch normalization because gradient penalization is done for each sample in the batch.} \\ \hline
        
        {LSGAN~\cite{mao2017least}, 2017} & {1) Remedies the vanishing gradient and stabilized GAN training. 2) Improves the mode diversity for the model. 3) Easy to implement.}&{1) Generated samples are pushed to decision boundary instead of real data, which  may affect the generated image quality.} \\ \hline
        
        {LS-GAN~\cite{qi2017loss}, 2017} & {1) Solves the vanishing gradient problem. 2) Solves the mode collapse problem.}&{1) Difficult to implement. Lots of components have to be carefully designed for the loss function. 2) The quality of the generated images might be affected by the margin which has been added between real samples and generated samples.} \\ \hline
        
        {Geometric GAN~\cite{lim2017geometric}, 2017} & {1) Less mode collapse. 2) More stable training. 3) Converges to the Nash equilibrium between the discriminator and generator.} &{1) Has not demonstrated capability in the face of the vanishing gradient problem. 2) Experimental tests have to be done on more complex datasets 
        e.g., ImageNet.} \\ \hline
        
        {Unrolled GAN~\cite{metz2016unrolled}, 2018} & {1) Solves the mode collapse problem. 2) Demonstrates that high order gradient information can help in training. 3) Improves GAN training stability.}&{1) The quality of the generated images are low.} \\ \hline
        
        {RGAN~\cite{jolicoeur2018relativistic}, 2018} & {1) Solves the vanishing gradient problem. 2) Unified framework for IPM-based GANs. 3) Solves the mode collapse problem.}&{1) Lack of mathematical implications of adding relativism to GANs. 2) Has not done a survey on which IPM-based GAN will achieve the best performance through the addition of this relativism.}\\ \hline
        
        {SN-GAN~\cite{miyato2018spectral}, 2018} & {1) Computationally light and easy to implement on existing GANs. 2) Improves image quality and solves mode collapse. 3) Stabilizes GAN training and solves the vanishing gradient problem.}&{1) Requires testing on more complex image datasets.}\\ \hline
        
        {Sphere GAN~\cite{Park_2019_CVPR}, 2019} & {1) Enable stable training. 2) More accurate results generated as high-order statistical information has been exploited. 3) Sphere GAN does not need any additional constraints that forces discriminators lie in a desired function space.}&{1) The diversity of sphere GAN needs to be investigated in the future.}\\ \hline
        
        {SS-GAN~\cite{Chen_2019_CVPR}, 2020} & {1) Enables the ability of self-supervision. 2) Results are competitive compared to current GAN models.}&{1)More advanced self-supervised architecture for the discriminator could be included in the future.}\\ \hline

        {RealnessGAN~\cite{xiangli2020real}, 2020} & {1) Discriminator outputs distribution as a measure of realness. 2) Image quality has been improved.}&{1) Model diversity needs to be improved in the future.}\\
        \bottomrule 
    \end{tabularx}
    \caption{Summary of loss-variant for GANs.}
    \label{chap05-tab:GAN-loss_sum}
\end{table*}
gives details of the properties of each loss-variant GAN. WGAN, LSGAN, LS-GAN, RGAN and SN-GAN are proposed to overcome the vanishing gradient for $G$. LSGAN argues that the vanishing gradient is mainly caused by the sigmoid function in the discriminator so it uses a least squares loss to optimize the GAN. LSGAN turns out to be the optimization on Pearson $\chi^2$ divergence and remedies the vanishing gradient problem. WGAN uses Wasserstein (or Earth mover) distance as the loss. Compared to JS divergence, Wasserstein distance is smoother and there is no sudden change with respect to the distance between real samples and generated samples. To be able to use Wasserstein distance as the loss, the discriminator must be Lipschitz continuous (constraint on the discriminator i.e., \textbf{IPM-based}), where WGAN deploys the parameter clipping to force discriminator satisfy the Lipschitz continuity. However, it causes problems such as most of parameters in the discriminator locates to the edges of clipping range, which leads to the \textbf{low capacity} of discriminator. WGAN-GP is proposed the use of gradient penalty to make discriminator is Lipschitz continuous, which successfully solves the problems in WGAN. LS-GAN proposes to use a margin that is enforced to separate real samples from generated samples, which restricts the modelling capability of discriminator. It solves the vanishing gradient problem for the generator because this problem arises when the discriminator is optimized. RGAN is a unified framework that is suitable for all IPM-based GANs e.g., WGAN. RGAN adds discriminant information to GANs for better learning. SN-GAN proposes an elegant way for optimizing a GAN. As mentioned before, a Lipschitz continuous discriminator is important for stable learning, vanishing gradient and so on. SN-GAN proposes spectral normalization~\cite{yoshida2017spectral} to constrain the discriminator under the Lipschitz continuous requirement. SN-GAN is the first GAN (we do not consider AC-GANs~\cite{odena2017conditional} because an ensemble of 100 AC-GANs is used for ImageNet datasets~\cite{deng2009imagenet}.) that has been successfully applied to ImageNet datasets. In theory, spectral normalization as demonstrated in the SN-GAN could be applied to every GAN type. SAGAN and BigGAN~\cite{brock2018large,zhang2018self} both deploy spectral normalization and achieve good results with ImageNet.

Loss-variant GANs are able to be applied to architecture-variants. However, SN-GAN and RGAN show stronger generalization abilities compared to other loss-variants, where these two loss-variants can be deployed by other types of loss-variants. Spectral normalization can be applied to any GAN-variant~\cite{miyato2018spectral} while the RGAN concept can be applied to any IPM-based GAN~\cite{jolicoeur2018relativistic}. We strongly recommend the use of spectral normalization for all GANs applications as described here. There are a number of loss-variant GANs mentioned in this paper which is able to solve the mode collapse and unstable training problem. Details are given in Table~\ref{chap05-tab:GAN-loss_sum}.

\section{Applications}
\subsection{Data Augmentation}\label{sec:data-augmentation}
\begin{figure}[!ht]
    \centering
    \includegraphics[width=.7\textwidth]{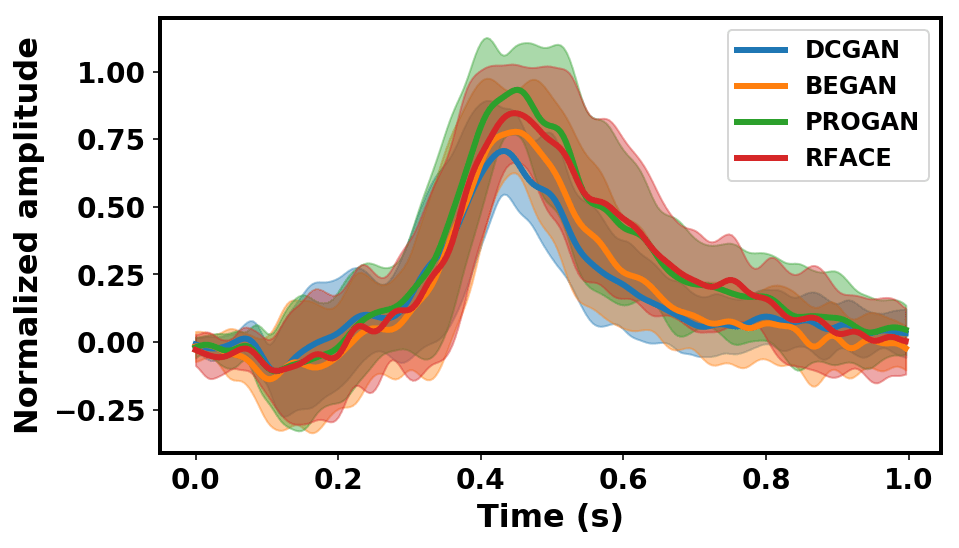}
    \caption{P300 (postive ongoing peak between 0.4 and 0.6 s) signal across 12 participants with respect to different types of GANs and real face from the study~\cite{wang2018use,wang2019neuroscore}.}
    \label{fig:EEG}
\end{figure}

\begin{figure}[!ht]
    \centering
    \includegraphics[width=.9\textwidth]{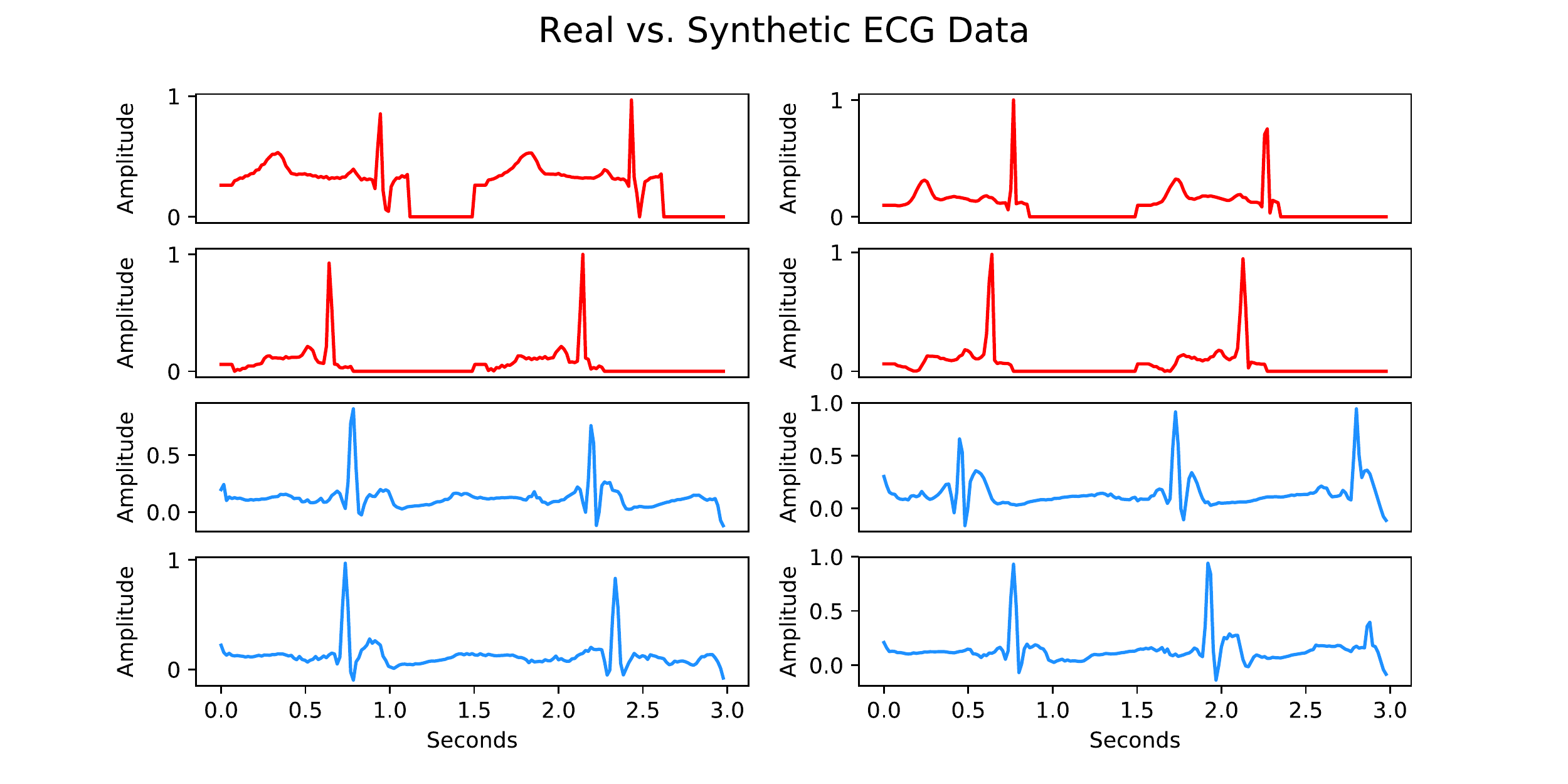}
    \caption{Examples of real (red) and synthetic (blue) ECG signals from the study~\cite{delaney2019synthesis}.}
    \label{fig:ECG}
\end{figure}

A well-known application for unsupervised learning approaches is augmenting the data. In reality, there are three circumstances requiring data augmentation: (1) Training a large deep neural network but the labeled data is very limited, (2) current data lacks variations e.g., do not cover diverse illuminations and appearance variations, and (3) access to a database is strictly restricted e.g., the database contains sensitive information. The first two situations can have alternative solutions without using unsupervised fashion but they will cost lots of extra hours and need lots of people engaged e.g., collect data and label data. GANs can be reliable approach to help us augment the data. Semi-supervised GAN (SGAN)~\cite{souly2017semi} provides such a GAN architecture that is able to automatically generate annotated images, which can be a solution to those situations that lack training data. The second case might have more breadth in real world. Lots of psychology or neuroscience experiments need broad variational data as stimulus. For example, human's EEG is sensitive to different types of face images such as a happy face and a sad face~\cite{mavratzakis2016emotional}. It takes lots of time for researchers to prepare for the stimulus in traditional psychology and neuroscience experiments. Architectures such as a style-based generator~\cite{karras2018style} can produce abroad different types of face images as stimulus. More importantly, these generated images can be generated with the level of variation i.e., different levels of happiness, which can be used for quantitative study in the experiment. Fortunately, we are excited to see researchers from neuroscience field begin to deploy GANs to generate the stimulus for the experiment e.g., Wang \textit{et al.}~\cite{wang2018use,wang2019neuroscore} used GANs for generating different types faces to study the P300 component~\cite{polich2007updating} in their EEG experiment as seen in Figure~\ref{fig:EEG}.
Compared to the first two perspectives, which can be solved without unsupervised fashion, the last aspect does not have a solution without borrowing unsupervised learning approaches, which imposes significant difficulty on some research such as seizure detection via EEG i.e., the seizure EEG data can only be accessed by very few researchers because of the privacy issue. Benefiting from data generation through GANs, it now becomes possible to synthesize those sensitive data. We have already seen some researchers applied GANs to those fields~\cite{hartmann2018eeg,brophy2019quick,delaney2019synthesis}. It is exciting to see some research areas could benefit from deep learning research. Figure~\ref{fig:ECG} illustrates examples of real ECG (in red) and synthetic ECG (in blue) generated by GANs, in which it can be seen that the generated ECG is very similar to the real ECG.            

\subsection{Image Synthesis}
Image synthesis is still a main focused area currently, in which lots of GAN variants have been proposed. In this section, we classify all applications related to images under the image synthesis category such as image super-resolution, image-to-image translation and image matting. 

\vspace{10pt}
\noindent
\textbf{Image Super-Resolution}\hspace{10pt} Image super-resolution enables a high-resolution image generated from a low-resolution image by upsampling. SRGAN~\cite{ledig2017photo} is a representative framework for image super-resolution by using GANs. Apart from general adversarial loss in GANs, SRGAN extends the loss by adding content loss (e.g., a pixel-wise MSE loss) in the area of super-resolution, which lead to a perceptual loss presented as below
\begin{equation}
    \mathcal{L}^{\mathrm{SR}} = \mathcal{L}_{\mathrm{X}}^{\mathrm{SR}} + 10^{-3}\mathcal{L}^{\mathrm{SR}}_{\mathrm{GAN}}
    \label{eq:SRGAN}
\end{equation}
where $\mathcal{L}_{X}^{\mathrm{SR}}$ is the content loss and $\mathcal{L}^{\mathrm{SR}}_{\mathrm{GAN}}$ is the GAN loss. In practice, the content loss $\mathcal{L}_{X}^{\mathrm{SR}}$ is chosen depending on applications. SRGAN presents three content losses (1) standard pixel-wise MSE loss $\mathcal{L}_{MSE}^{\mathrm{SR}}$, (2) a loss defined on feature maps representing lower-level features $\mathcal{L}_{VGG22}^{\mathrm{SR}}$ and (3) a loss defined on feature maps representing higher-level features $\mathcal{L}_{VGG54}^{\mathrm{SR}}$. The authors show that different content losses perform differently according to different evaluation metrics. The generator in SRGAN is conditioned by low-resolution images, which are inferred with $4\times$ upscaling factors. The authors show the superior perceptual performance of SRGAN compared to traditional approaches.     

\begin{figure}[ht!]
    \centering
    \includegraphics[width=1.\textwidth]{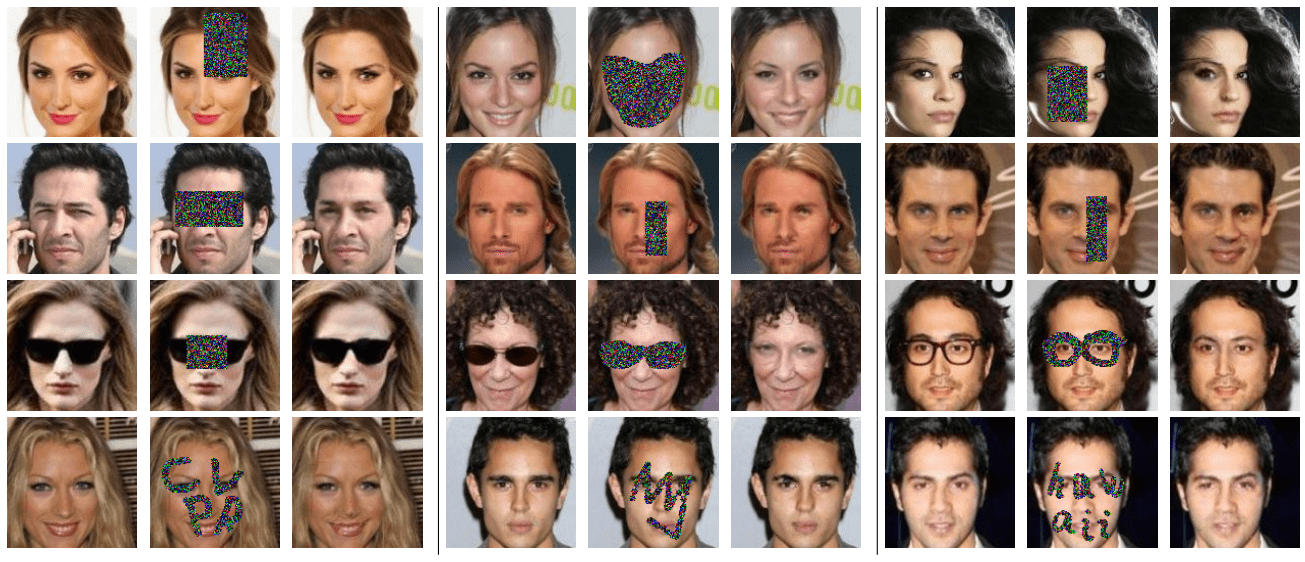}
    \caption{Face completion results using the CelebA dataset. From left to right: original images, masked inputs and completion results~\cite{li2017generative}.}
    \label{fig:face-completion}
\end{figure}

\vspace{10pt}
\noindent
\textbf{Image Completion/Repair}\hspace{10pt} Image completion/repair is a common image editing operation, which aims to fill the missing or masked regions in images with synthesized content. Most efficient traditional completion algorithms~\cite{barnes2009patchmatch,huang2014image} depend on low-level cues, which are used to search for patches from known regions of the same image, and synthesize the contents that locally appear similarly to the matched patches. These approaches perform well for background completion as patterns from the background are similarly to each other. The assumption of similar patterns for a missing part with other parts in an image can be violated in some situations e.g., filling missing parts for a face image, in which many objects have unique patterns. Li \textit{et al.}~\cite{li2017generative} propose to use 
autoencoder incorporating with GANs. The network architecture is shown in Figure~\label{fig:face-completion-arch}. Two discriminators $D_G$ and $D_L$ are deployed (one for global image content while the other for local image content), which turns out two adversarial losses are used for optimization. The overall loss function is represented as
\begin{equation}
    \mathcal{L} = \mathcal{L}_r + \lambda_1\mathcal{L}_{D_G} + \lambda_2\mathcal{L}_{D_L} + \lambda_3\mathcal{L}_p
    \label{eq:face-completion-loss}
\end{equation}
where $\mathcal{L}_r$ is the $L_2$ distance between the autoencoder output and the original image, $\mathcal{L}_{D_G}$ and $\mathcal{L}_{D_L}$ are adversarial losses of $D_G$ and $D_L$,and $\mathcal{L}_p$ is the pixel-wise softmax for the parsing network~\cite{long2015fully,yang2016object}. $\lambda_1$, $\lambda_2$ and $\lambda_3$ are the hyperparameters to control the effects caused by different losses. Figure~\ref{fig:face-completion} presents the completion results by using the CelebA dataset from the original paper. 
\begin{figure}[ht!]
    \centering
    \includegraphics[width=1.\textwidth]{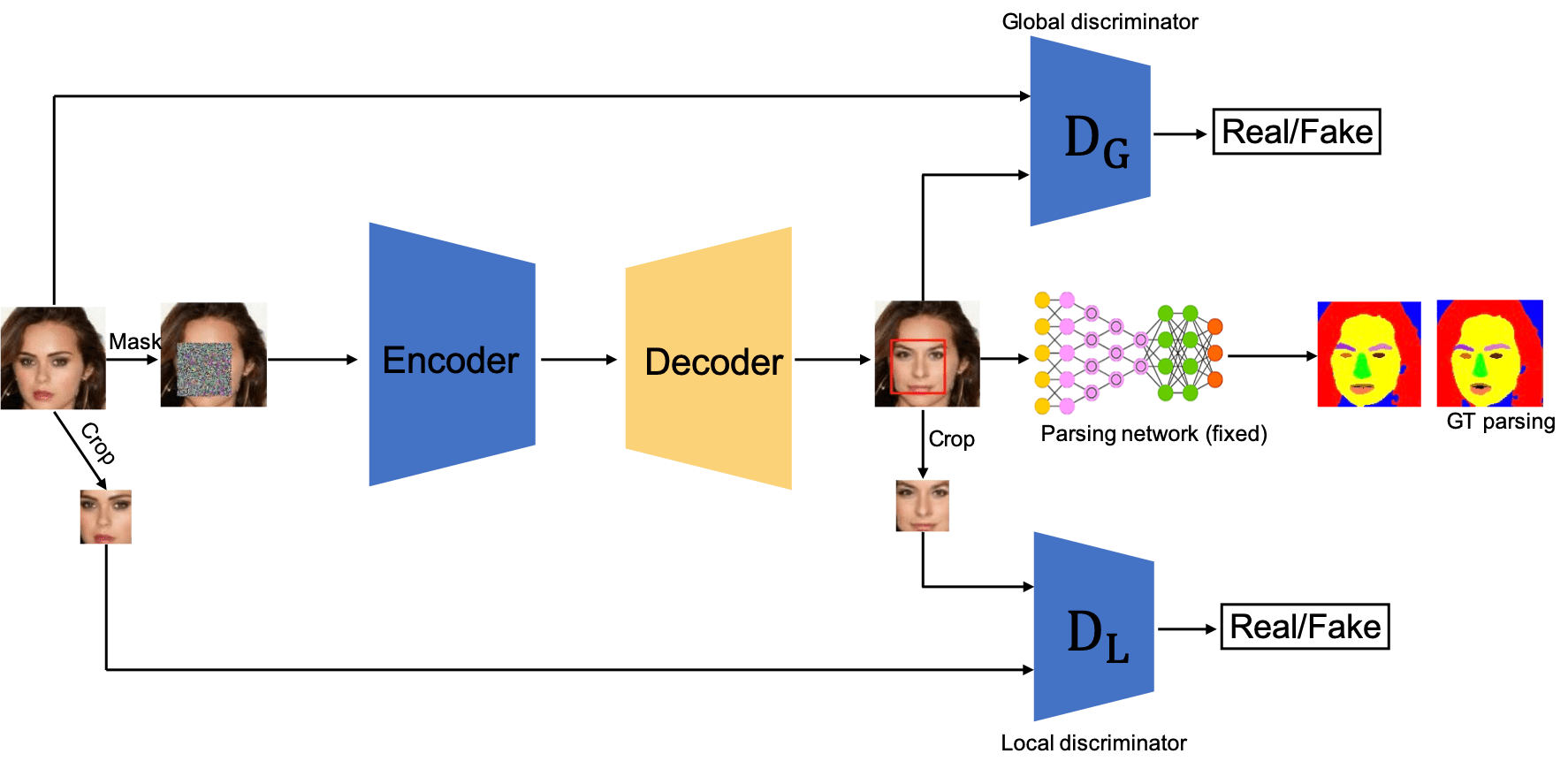}
    \caption{Network architecture for face completion in~\cite{li2017generative}.}
    \label{fig:face-completion-arch}
\end{figure}

\vspace{10pt}
\noindent
\textbf{Image Matting}\hspace{10pt} Natural image matting is defined to accurately estimate the opacity of foreground object in an image or video stream~\cite{lutz2018alphagan}. This field attracts a growing interest as it has wide applications such as image editing and film post-production. Image matting approaches formally require a input image with a foreground object and the image background, which can be mathematically expressed as
\begin{equation}\label{eq:alpha-GAN}
    I_i = \alpha_iF_i + (1-\alpha_i)B_i, \hspace{10pt} \alpha_i \in [0,1]
\end{equation}
where $\alpha_i$ is a scalar value that defines the foreground opacity at pixel $i$, $F_i$ is a scalar value of foreground object at pixel $i$ and $B_i$ is a scalar value of background at pixel $i$. Lutz \textit{et al.} introduces GANs to this field and proposes AlphaGAN that can produce visually appealing compositions. In their work, they train the discriminator on images composited with the ground-truth alpha and the predicted alpha and the generator is used to generate the compositions. Figure~\ref{fig:image-matting} illustrates the performance of AlphaGAN compared with other approaches in the literature.
\begin{figure}[!ht]
    \subfigure{
    \centering
    \includegraphics[width=1.\textwidth]{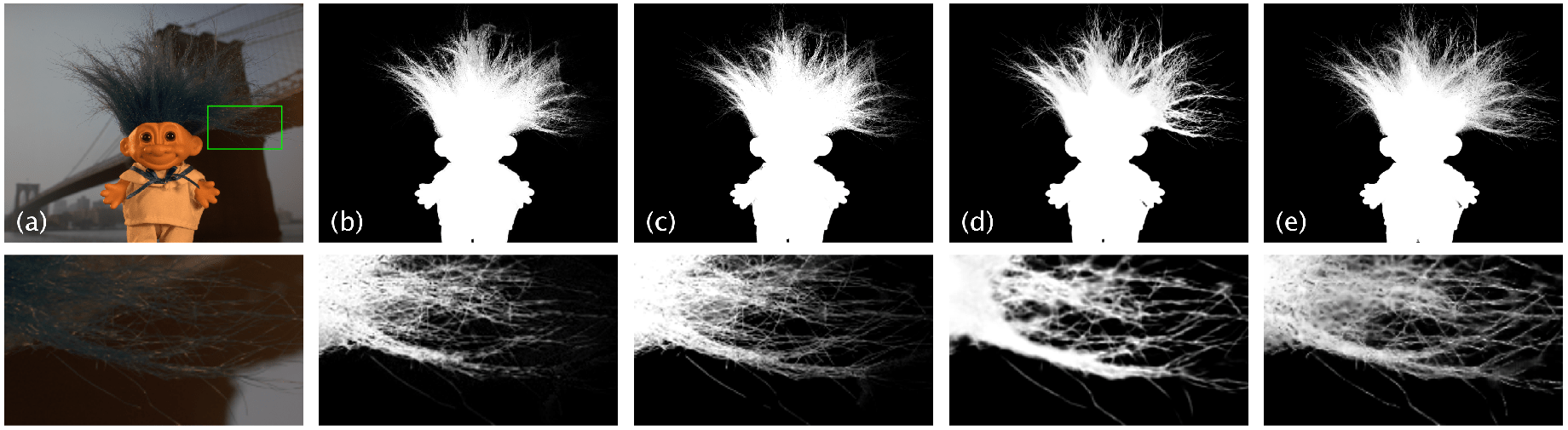}
    }
    \subfigure{
    \centering
    \includegraphics[width=1.\textwidth]{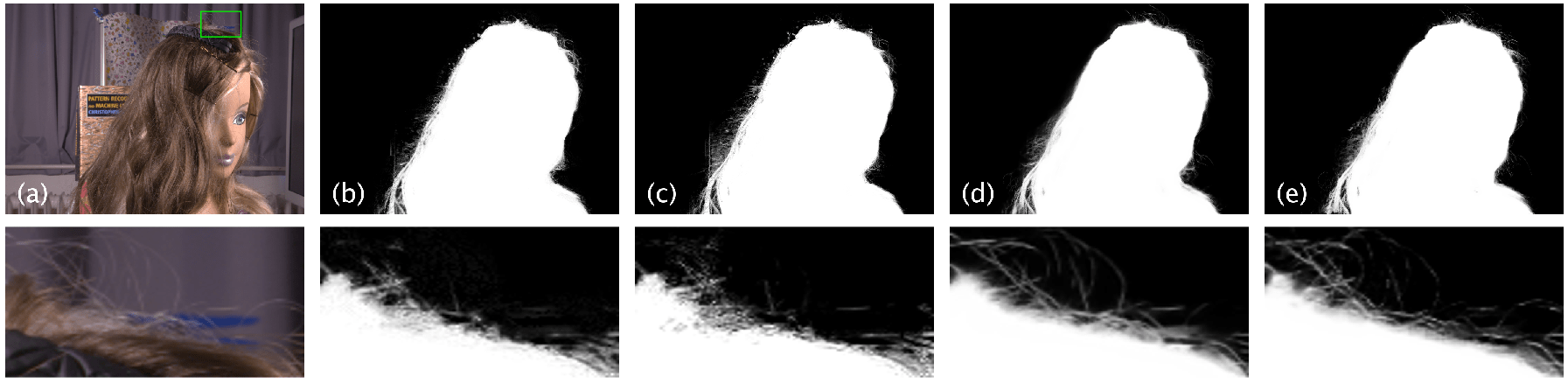}
    }
    \subfigure{
    \centering
    \includegraphics[width=1.\textwidth]{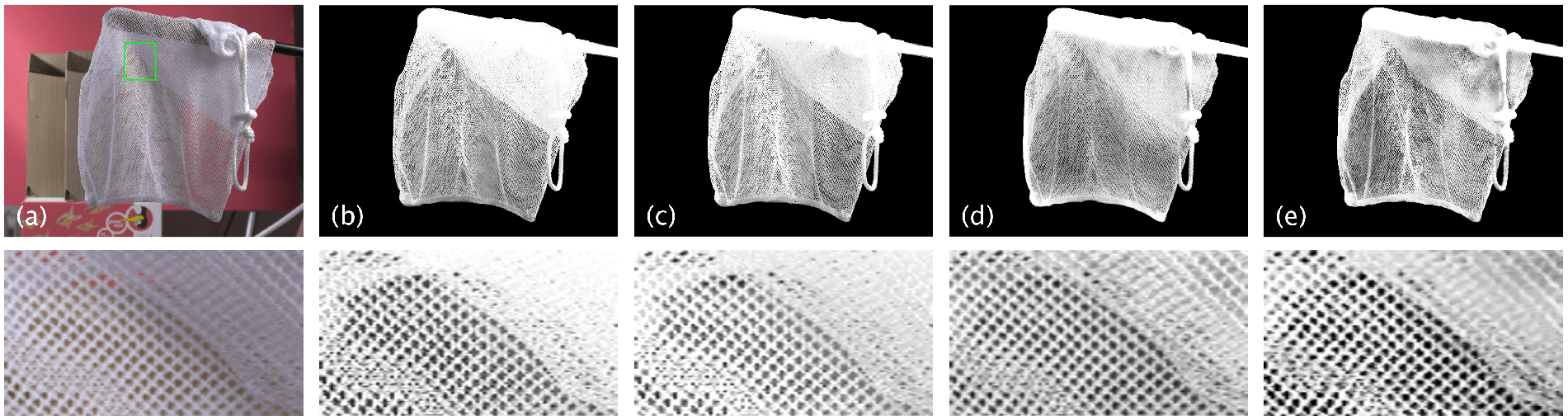}
    }
    \label{fig:image-matting}
    \caption{Examples of Alpha matting predictions using the \url{alphamatting.com} dataset. From left to right: raw images, DCNN~\cite{cho2016natural}, IF~\cite{aksoy2017designing}, DI~\cite{xu2017deep} and AlphaGAN~\cite{lutz2018alphagan}. Figures from \cite{lutz2018alphagan}.}
    \label{fig:image-matting}
\end{figure}

\begin{figure}[!ht]
    \centering
    \includegraphics[width=1.\textwidth]{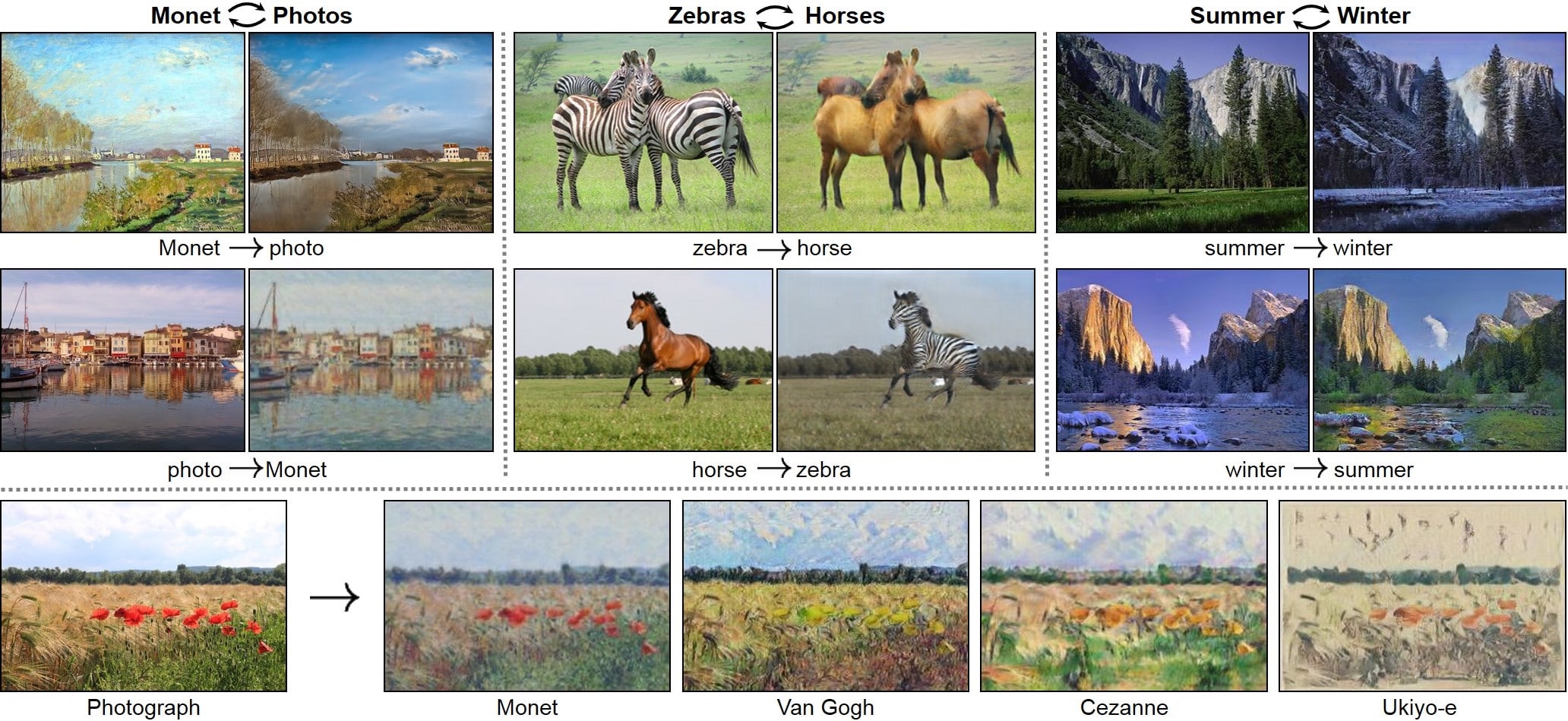}
    \caption{Top and middle: CycleGAN learns to translate an image from one into the other and vice versa given any two unordered image collections. Bottom: An example of application, in which CycleGAN learns to render natural photographs into the different styles by using a collection of paintings. Figure from \cite{zhu2017unpaired}.}
    \label{fig:cycleGAN}
\end{figure}

\vspace{10pt}
\noindent
\textbf{Image-to-image Translation}\hspace{10pt} Image-to-image translation is a class of graphic problems, in which the objective is to learn the mapping between an output image and an input image using a training set of aligned image pairs. Isola \textit{et al.}~\cite{isola2017image} propose to use CGAN for image-to-image translation when paired training data is available. CycleGAN~\cite{zhu2017unpaired} is proposed to solve when unpaired training data is not available for image-to-image translation. CycleGAN achieves this by introducing cycle consistency loss to enforce the mapping from one domain $X$ and the other domain $Y$ is roughly same. Figure~\ref{fig:cycleGAN} shows some results generated by CycleGAN, which proves to be able to be applied for lots of graphical problems. Other work such as DiscoGAN~\cite{kim2017learning} and DualGAN~\cite{yi2017dualgan} are also proposed to solve the missing paired training data issue in the area of image-to-image translation. More details can be refered to original papers.

\subsection{Video Generation}
GANs have already achieved striking results for natural images generation. Some recent work attempt to carry this success to the area of video generation~\cite{tulyakov2018mocogan,clark2019adversarial,kahembwe2019lower}. The efficient video generation using GANs remains a significant challenge as it exacerbates all the issues associated with image generation using GANs. Also it has the problems of increased memory and computation costs due to the 3D nature of video and the requirement for temporal modelling. Besides, due to the limitation of memory and training stability, the generation becomes increasingly challenging with the increase of the resolution/duration of videos. The current mainstreams of the video generation research via GANs we think are 1) producing high-resolution videos, e.g., up to/more than 256$\times$256; 2) increasing the lengths of the generated video up to/more than 48 frames, and 3) generating more realistic video so that much of the generated video content is not blurred, indistinct, or even surreal. Videos based GANs does not only have to consider the spatial modeling but also require temporal modeling i.e., the motion between each adjacent frames in a video sequence. MoCoGAN~\cite{tulyakov2018mocogan} is proposed to learn motion and content in unsupervised fashion, which divides the image latent space into content and motion spaces. DVD-GAN~\cite{clark2019adversarial} is able to generate longer and higher resolution videos based on the BigGAN architecture while introducing scalable, video-specific generator and discriminator architectures. DVD-GAN comprises two discriminators $D_S$ (spatial discriminators) and $D_T$ (temporal discriminators), in which $D_S$ scores input frames from the spatial perspective and $D_T$ scores input frames from the temporal perspective (motion).

\subsection{Feature Generation}
``Can machines think?'' This is the question raised in Alan Turing's seminal paper entitled ``Computing Machinery and Intelligence''~\cite{turing2009computing} in 1950. The ultimate goal of machines is to be as intelligent as human beings. Current AI depends on large datasets rather than being generalized from small datasets and has catastrophic forgetting problem~\cite{mccloskey1989catastrophic} when learning tasks sequentially. Few-shot learning and continual learning attract significant attention from scientific community in recent years, which aim to fill gaps between AI and humans. Both of these two fields suffer from the problem of lack of data i.e., few-shot learning uses few samples (e.g., normally 1 sample or 5 sample) for each category and continual learning comes across unseen data in sequential tasks. Mandal \textit{et al.}~\cite{mandal2019out} proposes use of conditional Wasserstein GAN with two additional terms i.e., cosine embedding and cycle-consistency losses to synthesize unseen action features for zero-shot action recognition. Shin \textit{et al.}~\cite{shin2017continual} proposes a deep generative replay framework for continual learning, in which training data for previous tasks are sampled by the generator and interleaved with those for a new task. These show that GANs have potential ability to provide solutions for other machine leanring problems.

\section{Notes on Evaluation of GANs}\label{sec:Notes_on_evaluation}
It has been known that comparison between generative models is challenging~\cite{theis2015note} and a number of evaluation criterion has been proposed in the literature~\cite{borji2019pros}. In general, evaluation for GANs is mainly divided into two main types i.e., qualitative and quantitative measurements respectively. The qualitative measurement is supposed to determine the visual quality of generated images from a human perceptual perspective. The most representative qualitative measurement is using human annotation. As human annotation approaches are time-consuming i.e., these require asking evaluators to generate behavioral responses on an image-by-image basis. Several approaches have been proposed and proven to be correlated with human annotation e.g., Inception Score (IS)~\cite{salimans2016improved}, Fre\'chet Inception Distance (FID)~\cite{heusel2017gans}, Neuroscore~\cite{wang2018use}. Quantitative metrics in contrast, are less subjective but the robustness of their performance is compromised. Quantitative measurements such as kernel maximum mean discrepancy (MMD)~\cite{gretton2012kernel}, sliced Wasserstein distance (SWD)~\cite{karras2017progressive}, classifier two-sample tests (C2ST)~\cite{lopez2016revisiting} can be applied to detect issues e.g., overfitting or memorizing, low diversity, mode dropping for GANs~\cite{borji2019pros}. Despite a number of existing metrics for evaluating GANs, Inception Score and FID are the most widely used measurement because both of these two metrics can indicate quality and the diversity of generated images~\cite{salimans2016improved,heusel2017gans}. These two metrics become main evaluation metrics for evaluating the performance for different GANs. It is obvious that human would be the most reliable to evaluate the quality of generated images. Nevertheless, direct comparison between the human annotator and evaluation metrics is absent in the literature. We have designed a comprehensive comparison among humans, humans' neural signals (Neuroscore) and evaluation metrics, which has been published~\cite{wang2018use}. Specifically, we trained three GANs i.e., DCGAN, BEGAN and PROGAN on CelebA, Three evaluation metrics Inception Score, MMD and FID are selected for comparison and 12 participants are gathered for the experiment. We have a behavioral task to record participants' responses to each type of GAN-generated images, in which participants answer "real" or "fake" when seeing a image presented on the screen. The accuracy for each type of GAN is shown in Table~\ref{tab:BE-ACC}. It is worth nothing that lower accuracy for GAN-generated images indicates better image quality i.e., participants were often convinced that synthesized faces were in fact real. It can be noticed PROGAN performs the best, followed by BEGAN and then DCGAN. Table~\ref{tab:traditional-method-score} summarizes the different results across three GANs evaluated by using different metrics. To be consistent with other metrics (smaller score indicates better GAN performance), we use 1/IS (1/Inception Score) for comparison. It can be seen that all three methods are consistent with each other and they rank the GANs in the same order of PROGAN, DCGAN and BEGAN from high to low performance. By comparing the three traditional evaluation metrics to the human, it can be seen that they are not consistent with human judgment of GAN performance. It should be noted that we are not arguing which metric is the best and we understand metrics such as Inception Score and MMD also consider other factors such as model diversity. However, those metrics might not be able to fully interpret the quality of images generated by a GAN.     
\begin{table}[!htp] 
	\centering
	\begin{tabular}{c|c|c|c}
		\toprule
		{ID}& {DCGAN} & {BEGAN} & {PROGAN}\\ \midrule
		{1} &{1.000} & {0.759} & {0.704}\\
		{2} &{0.981} & {0.741} & {0.537}\\
		{3} &{1.000} & {0.796} & {0.778}\\
		{4} &{0.981} & {0.889}& {0.704}\\
		{5} &{1.000} & {0.667} & {0.648}\\
		{6} &{1.000} & {0.926}& {0.704}\\
		{7} &{1.000} & {0.815}& {0.611}\\
		{8} &{0.981} & {0.815} & {0.870}\\
		{9} &{1.000} & {0.796}  & {0.685}\\
		{10} &{1.000} & {0.815}  & {0.759}\\
		{11} &{1.000} & {0.907}  & {0.759}\\
		{12} &{1.000} & {0.963}  & {0.704}\\\midrule
		{Mean} &{\textbf{0.995}} & {0.824} & {\textbf{0.705}}\\
		\bottomrule
	\end{tabular}
	\caption{Accuracy for face images generated from three GANs in the behavioral task. Lower accuracy for GAN-generated images indicates better image quality  i.e.,  participants were often convinced that synthesized faces were in fact real~\cite{wang2018use}.}
	\label{tab:BE-ACC}
\end{table}

\begin{table}[!htbp] 
	\centering
	\begin{tabular}{c|c|c|c}
		\toprule
		{Metrics}& {DCGAN} & {BEGAN} & {PROGAN}\\
		\midrule
		{1/IS} &{\textcolor{red}{0.44}} & {\textcolor{red}{0.57}} & {0.42} \\
		{MMD} &{\textcolor{red}{0.22}} & {\textcolor{red}{0.29}}  & {0.12}\\
		{FID} &{\textcolor{red}{63.29}} & {\textcolor{red}{83.38}} & {34.10}\\
		{1/Neuroscore} &{1.715} & {1.479} & {1.195}\\
		{Human} &{\textbf{0.995}} & {\textbf{0.824}} & {\textbf{0.705}} \\
		\bottomrule
	\end{tabular}
	\caption{ Score comparison for each GAN category. Lower score indicates better performance of GAN~\cite{wang2018use}.}
	\label{tab:traditional-method-score}
\end{table}

Table~\ref{tab:performance} demonstrates our summary for the performance of presented GANs in this work by using Inception Score and FID in the literature. Four image datasets are considered in the summary i.e., CIFAR10, ImageNet, LSUN and CelebA, which are the most widely used benchmarking datasets.
\begin{table}[!ht]
    \centering
    \begin{tabular}{c|c|c|c|c}
        \toprule
        Model &  {CIFAR10 (IS/FID)} & {ImageNet (IS/FID)} & {LSUN (FID)}  & {CelebA (FID)} \\ \midrule
        FCGAN &  6.41/42.6 & -/-& -& -\\
        BEGAN &  5.62/- & -/-& -& 83.3\\
        PROGAN &  8.80/- & -/-& 8.3& \textbf{7.3}\\
        LSGAN &  6.76/29.5 & -/-& 21.6& -\\
        DCGAN &  6.69/42.5 & -/74.2& 160.1& 63.1\\
        WGAN-GP &  8.21/21.5 & 11.6/62.1& 22.8& -\\
        SN-GAN &  8.43/18.8 & 36.8/27.6& -& -\\
        Geometric GAN &  -/27.1 & -/-& -& -\\
        RGAN &  -/15.9 & -/-& -& -\\
        AC-GAN & 8.25/- & -/-& -&-\\
        BigGAN & \textbf{9.22}/\textbf{14.7} & \textbf{166.5}/\textbf{7.4}& -&-\\
        RealnessGAN & -/34.6  & -/-& -& 23.5\\
        MSG-GAN & -/-  & -/-& \textbf{5.2}& 8.0\\
        SS-GAN & -/15.7  & -/43.9& 13.3& 24.36\\
        YLG & -/-  & 57.2/15.9& -& -\\
        Sphere GAN & -/-  & -/-& 16.9& -\\
        \bottomrule
    \end{tabular}
    \caption{Performance summary across different types of GANs discussed in this paper on different datasets in the literature i.e., CIFAR10, ImageNet, LSUN, CelebA. "-" refers that experiments have not been done in the literature. Studies from~\cite{karras2017progressive,mao2017least,radford2015unsupervised,jolicoeur2018relativistic,shmelkov2018good,miyato2018spectral,im2018quantitatively,gulrajani2017improved,odena2017conditional,song2019bridging,karnewar2020msg,xiangli2020real,karnewar2020msg,wang2018use,shmelkov2018good,mao2018effectiveness,zhang2018self}.}    
    \label{tab:performance} 
\end{table}

\section{Discussion}\label{sec:discussion}
We have introduced the most significant problems present in the original GAN design, which are mode collapse and vanishing gradient for updating $G$. We have surveyed significant GAN-variants that remedy these problems through two design considerations: (1) Architecture-variants. This aspect focuses on architectural options for GANs. This approach enables GANs to be successfully applied to different applications, however, it is not able to fully solve the problems mentioned above; (2) Loss-variant. We have provided a detail explanation why these problems arise in the original GAN. These problems are essentially caused by the loss function in the original GAN. Thus, modifying this loss function can solve this problem. It should be noted that the loss function may change for some architecture-variants. However, this loss function is changed according to the architecture thus it is architecture-specific loss. It is not able to generalize to other architectures.

Through a comparison of the different architectural approaches surveyed in this work, it is clear that the modification of the GAN architecture has significant impact on the generated images quality and their diversity. Recent research shows that the capacity and performance of GANs are related to the network size and batch size~\cite{brock2018large}, which indicates that a well designed architecture is critical for good GANs performance. However, modifications to the architecture only is not able to eliminate all the inherent training problems for GANs. Redesign of the loss function including regularization and normalization can help yield more stable training for GANs. This work introduced various approaches to the design of the loss function for GANs. Based on the comparison for each loss-variant, we find that spectral normalization as first demonstrated in the SN-GAN brings lots of benefits including ease of implementation, relatively light computational requirements and the ability to work well for almost all GANs. We suggest that researchers, who seek to apply GANs to real-world problems, include spectral normalization to the discriminator. 

There is no answer to the question of which GAN is the best. The selection of a specific GAN type depends on the application. For instance, if an application requires the production of natural scenes images (this requires generation of images which are very diverse). DCGAN with spectrum normalization applied, SAGAN and BigGAN can be good choices here. BigGAN is able to produce the most realistic images compared to the other two. However, BigGAN is much more computationally intensive. Thus it depends on the actual computational requirements set by the real-world application.

\vspace{10pt}
\noindent
\textbf{Interconnections Between Architecture and Loss}\hspace{10pt}
In this paper, we highlight the problems inherent in the original GAN design. In highlighting how subsequent researchers have remedied those problems, we explored architecture-variants and loss-variants in GAN designs separately. However, it should be noted that there are interconnections between these two types of GAN-variants. As mentioned before, loss functions are easily integrated to different architectures. Benefit from improved convergence and stabilization through a redesigned loss function, architecture-variants are able to achieve better performance and accomplish solutions to more difficult problems. For examples, BEGAN and PROGAN use Wasserstein distance instead of JS divergence. SAGAN and BigGAN deploy spectral normalization, where they achieve good performance based on multi-class image generation. These two types of variants equally contribute to the progress of GANs.  

\vspace{10pt}
\noindent
\textbf{Future Directions}\\
GANs were originally proposed to produce plausible synthetic images and have achieved exciting performance in the computer vision area. GANs have been applied to some other fields, (e.g., time series generation~\cite{esteban2017real,hartmann2018eeg,luo2018multivariate,brophy2019quick} and natural language processing~\cite{yu2017seqgan,bahdanau2016actor,li2016deep,fedus2018maskgan}) with some success. Compared to computer vision, GANs research in other areas is still somewhat limited. The limitation is caused by the different properties inherent in image versus non-image data. For instance, GANs work to produce continuous value data but natural language are based on discrete values like words, characters and bytes, so it is hard to apply GANs for natural language applications. As this is also a very promising area, success in this area will lead to lots of applications such as generating subtitle and generating comments to live streaming. As mentioned in section~\ref{sec:data-augmentation}, research areas such as neuroscience may have some privacy issue on the data and successful data augmentation will have significant impact in these areas. However, generation for other modal data such as time-series data using GANs is limitedly explored even there is lack of efficient evaluation metrics for evaluating the performance of GANs in those areas. More research is encouraged to be carried out in those areas. 

Since the first GAN has been proposed in 2014, the development of GANs has brought lots of benefits to us either in research or in our real life. However, improper use of GANs can also bring hidden concerns to society e.g., GANs can be used to generate a tampered video for specific people and inappropriate events, creating images that are detrimental to a particular person, and may even affect that personal safety~\cite{hsu2018learning,afchar2018mesonet}. We should also focus on developing forgery detector to efficiently and effectively detect the AI-genereated images (including using GANs).


\section{Conclusion}
In this paper, we review GAN-variants based on performance improvement offered in terms of higher image quality, more diverse images and more stable training. We review the current state of GAN-related research from an architecture and a loss basis. More complicated architecture with larger batch size can increase both image quality and image diversity i.e., BigGAN. However, limited GPUs memory could be an issue for processing large batched images, where the progressive training strategy used in PROGAN can increase the image quality without requiring very large batched images. Regarding the image diversity, adding the self-attention mechanism to the generator and the discriminator has achieved exciting results and the SAGAN has been successfully applied to the ImageNet. In terms of the stable training, the loss function plays an important role here and different types of loss functions have been proposed to deal with this perspective. Having reviewed different types of loss functions, we find the spectral normalization has good generalization i.e., it is able to be applied to every GAN, is easy to be implemented and has very low computational cost. Current state-of-art GANs models such as BigGAN and PROGAN are able to produce high quality images and diverse images in the computer vision field. However, research that applies GANs to video is limited. Moreover, GAN-related research in other areas such as time series generation and natural language processing lags that for computer vision in terms of performance and capability. We conclude that there are clearly opportunities for future research and application in these fields in particular.



\section*{Acknowledgment}
This work is funded as part of the Insight Centre for Data Analytics which is supported by Science Foundation Ireland under Grant Number SFI/12/RC/2289\_P2. The authors appreciate informative and insightful comments provided by anonymous reviewers, which significantly improve the quality of this survey.

\bibliographystyle{IEEEtran}
\bibliography{bare_jrnl}

\begin{thebibliography}{100}
\providecommand{\url}[1]{#1}
\csname url@samestyle\endcsname
\providecommand{\newblock}{\relax}
\providecommand{\bibinfo}[2]{#2}
\providecommand{\BIBentrySTDinterwordspacing}{\spaceskip=0pt\relax}
\providecommand{\BIBentryALTinterwordstretchfactor}{4}
\providecommand{\BIBentryALTinterwordspacing}{\spaceskip=\fontdimen2\font plus
\BIBentryALTinterwordstretchfactor\fontdimen3\font minus
  \fontdimen4\font\relax}
\providecommand{\BIBforeignlanguage}[2]{{%
\expandafter\ifx\csname l@#1\endcsname\relax
\typeout{** WARNING: IEEEtran.bst: No hyphenation pattern has been}%
\typeout{** loaded for the language `#1'. Using the pattern for}%
\typeout{** the default language instead.}%
\else
\language=\csname l@#1\endcsname
\fi
#2}}
\providecommand{\BIBdecl}{\relax}
\BIBdecl

\bibitem{goodfellow2014generative}
I.~Goodfellow, J.~Pouget-Abadie, M.~Mirza, B.~Xu, D.~Warde-Farley, S.~Ozair,
  A.~Courville, and Y.~Bengio, ``Generative adversarial nets,'' in
  \emph{Advances in Neural Information Processing Systems}, 2014, pp.
  2672--2680.

\bibitem{liu2016coupled}
M.-Y. Liu and O.~Tuzel, ``Coupled generative adversarial networks,'' in
  \emph{Advances in neural information processing systems}, 2016, pp. 469--477.

\bibitem{salimans2016improved}
T.~Salimans, I.~Goodfellow, W.~Zaremba, V.~Cheung, A.~Radford, and X.~Chen,
  ``Improved techniques for training {GAN}s,'' in \emph{Advances in Neural
  Information Processing Systems}, 2016, pp. 2234--2242.

\bibitem{turkoglu2019layer}
M.~O. Turkoglu, L.~Spreeuwers, W.~Thong, and B.~Kicanaoglu, ``A layer-based
  sequential framework for scene generation with {GAN}s,'' in
  \emph{Thirty-Third AAAI Conference on Artificial Intelligence}, Honolulu,
  Hawaii, United States, 2019.

\bibitem{wu2017gp}
H.~Wu, S.~Zheng, J.~Zhang, and K.~Huang, ``G{P}-{GAN}: {T}owards realistic
  high-resolution image blending,'' \emph{arXiv preprint arXiv:1703.07195},
  2017.

\bibitem{pan2017salgan}
J.~Pan, C.~C. Ferrer, K.~McGuinness, N.~E. O'Connor, J.~Torres, E.~Sayrol, and
  X.~Giro-i Nieto, ``Sal{GAN}: Visual saliency prediction with generative
  adversarial networks,'' \emph{arXiv preprint arXiv:1701.01081}, 2017.

\bibitem{dziugaite2015training}
G.~K. Dziugaite, D.~M. Roy, and Z.~Ghahramani, ``Training generative neural
  networks via maximum mean discrepancy optimization,'' \emph{arXiv preprint
  arXiv:1505.03906}, 2015.

\bibitem{ma2017pose}
L.~Ma, X.~Jia, Q.~Sun, B.~Schiele, T.~Tuytelaars, and L.~Van~Gool, ``Pose
  guided person image generation,'' in \emph{Advances in Neural Information
  Processing Systems}, 2017, pp. 406--416.

\bibitem{vondrick2016generating}
C.~Vondrick, H.~Pirsiavash, and A.~Torralba, ``Generating videos with scene
  dynamics,'' in \emph{Advances In Neural Information Processing Systems},
  2016, pp. 613--621.

\bibitem{yang2017high}
C.~Yang, X.~Lu, Z.~Lin, E.~Shechtman, O.~Wang, and H.~Li, ``High-resolution
  image inpainting using multi-scale neural patch synthesis,'' in
  \emph{Proceedings of the IEEE Conference on Computer Vision and Pattern
  Recognition}, 2017, pp. 6721--6729.

\bibitem{odena2017conditional}
A.~Odena, C.~Olah, and J.~Shlens, ``Conditional image synthesis with auxiliary
  classifier gans,'' in \emph{Proceedings of the 34th International Conference
  on Machine Learning}, vol.~70.\hskip 1em plus 0.5em minus 0.4em\relax JMLR,
  2017, pp. 2642--2651.

\bibitem{li2015generative}
Y.~Li, K.~Swersky, and R.~Zemel, ``Generative moment matching networks,'' in
  \emph{International Conference on Machine Learning}, 2015, pp. 1718--1727.

\bibitem{zhu2016generative}
J.-Y. Zhu, P.~Kr{\"a}henb{\"u}hl, E.~Shechtman, and A.~A. Efros, ``Generative
  visual manipulation on the natural image manifold,'' in \emph{European
  Conference on Computer Vision}.\hskip 1em plus 0.5em minus 0.4em\relax
  Springer, 2016, pp. 597--613.

\bibitem{lassner2017generative}
C.~Lassner, G.~Pons-Moll, and P.~V. Gehler, ``A generative model of people in
  clothing,'' in \emph{Proceedings of the IEEE International Conference on
  Computer Vision}, 2017, pp. 853--862.

\bibitem{fedus2018maskgan}
W.~Fedus, I.~Goodfellow, and A.~M. Dai, ``Mask{GAN}: Better text generation via
  filling in the \_ .'' \emph{arXiv preprint arXiv:1801.07736}, 2018.

\bibitem{yang2017semi}
Z.~Yang, J.~Hu, R.~Salakhutdinov, and W.~Cohen, ``Semi-supervised {QA} with
  generative domain-adaptive nets,'' in \emph{Proceedings of the 55th Annual
  Meeting of the Association for Computational Linguistics (Volume 1: Long
  Papers)}, Vancouver, Canada, 2017, pp. 1040--1050.

\bibitem{dai2017good}
Z.~Dai, Z.~Yang, F.~Yang, W.~W. Cohen, and R.~R. Salakhutdinov, ``Good
  semi-supervised learning that requires a bad {GAN},'' in \emph{Advances in
  neural information processing systems}, 2017, pp. 6510--6520.

\bibitem{jetchev2016texture}
N.~Jetchev, U.~Bergmann, and R.~Vollgraf, ``Texture synthesis with spatial
  generative adversarial networks,'' \emph{arXiv preprint arXiv:1611.08207},
  2016.

\bibitem{donahue2018synthesizing}
C.~Donahue, J.~McAuley, and M.~Puckette, ``Synthesizing audio with generative
  adversarial networks,'' \emph{arXiv preprint arXiv:1802.04208}, 2018.

\bibitem{hartmann2018eeg}
K.~G. Hartmann, R.~T. Schirrmeister, and T.~Ball, ``{EEG-GAN}: {G}enerative
  adversarial networks for electroencephalographic brain signals,'' \emph{arXiv
  preprint arXiv:1806.01875}, 2018.

\bibitem{esteban2017real}
C.~Esteban, S.~L. Hyland, and G.~R{\"a}tsch, ``Real-valued (medical) time
  series generation with recurrent conditional {GAN}s,'' \emph{arXiv preprint
  arXiv:1706.02633}, 2017.

\bibitem{li2019mad}
D.~Li, D.~Chen, L.~Shi, B.~Jin, J.~Goh, and S.-K. Ng, ``{MAD-GAN}:
  {M}ultivariate anomaly detection for time series data with generative
  adversarial networks,'' \emph{arXiv preprint arXiv:1901.04997}, 2019.

\bibitem{brophy2019quick}
E.~Brophy, Z.~Wang, and T.~E. Ward, ``Quick and easy time series generation
  with established image-based {GAN}s,'' \emph{arXiv preprint
  arXiv:1902.05624}, 2019.

\bibitem{zhu2016adversarial}
W.~Zhu, X.~Xiang, T.~D. Tran, and X.~Xie, ``Adversarial deep structural
  networks for mammographic mass segmentation,'' \emph{arXiv preprint
  arXiv:1612.05970}, 2016.

\bibitem{luc2016semantic}
P.~Luc, C.~Couprie, S.~Chintala, and J.~Verbeek, ``Semantic segmentation using
  adversarial networks,'' \emph{arXiv preprint arXiv:1611.08408}, 2016.

\bibitem{dong2017semantic}
H.~Dong, S.~Yu, C.~Wu, and Y.~Guo, ``Semantic image synthesis via adversarial
  learning,'' in \emph{Proceedings of the IEEE International Conference on
  Computer Vision}, 2017, pp. 5706--5714.

\bibitem{qiu2017deep}
Z.~Qiu, Y.~Pan, T.~Yao, and T.~Mei, ``Deep semantic hashing with generative
  adversarial networks,'' in \emph{Proceedings of the 40th International ACM
  SIGIR Conference on Research and Development in Information Retrieval}.\hskip
  1em plus 0.5em minus 0.4em\relax ACM, 2017, pp. 225--234.

\bibitem{souly2017semi}
N.~Souly, C.~Spampinato, and M.~Shah, ``Semi supervised semantic segmentation
  using generative adversarial network,'' in \emph{Proceedings of the IEEE
  International Conference on Computer Vision}, 2017, pp. 5688--5696.

\bibitem{goodfellow2016nips}
I.~Goodfellow, ``Nips 2016 tutorial: Generative adversarial networks,''
  \emph{arXiv preprint arXiv:1701.00160}, 2016.

\bibitem{karras2018style}
T.~Karras, S.~Laine, and T.~Aila, ``A style-based generator architecture for
  generative adversarial networks,'' \emph{arXiv preprint arXiv:1812.04948},
  2018.

\bibitem{wang2018high}
T.-C. Wang, M.-Y. Liu, J.-Y. Zhu, A.~Tao, J.~Kautz, and B.~Catanzaro,
  ``High-resolution image synthesis and semantic manipulation with conditional
  {GAN}s,'' in \emph{Proceedings of the IEEE Conference on Computer Vision and
  Pattern Recognition}, 2018, pp. 8798--8807.

\bibitem{poole2016improved}
B.~Poole, A.~A. Alemi, J.~Sohl-Dickstein, and A.~Angelova, ``Improved generator
  objectives for {GAN}s,'' \emph{arXiv preprint arXiv:1612.02780}, 2016.

\bibitem{choe2017face}
J.~Choe, S.~Park, K.~Kim, J.~Hyun~Park, D.~Kim, and H.~Shim, ``Face generation
  for low-shot learning using generative adversarial networks,'' in
  \emph{Proceedings of the IEEE International Conference on Computer Vision},
  2017, pp. 1940--1948.

\bibitem{zhang2017stackgan}
H.~Zhang, T.~Xu, H.~Li, S.~Zhang, X.~Wang, X.~Huang, and D.~N. Metaxas,
  ``Stackgan: {T}ext to photo-realistic image synthesis with stacked generative
  adversarial networks,'' in \emph{Proceedings of the IEEE International
  Conference on Computer Vision}, 2017, pp. 5907--5915.

\bibitem{zhu2017unpaired}
J.-Y. Zhu, T.~Park, P.~Isola, and A.~A. Efros, ``Unpaired image-to-image
  translation using cycle-consistent adversarial networks,'' \emph{arXiv
  preprint arXiv:1703.10593v6}, 2017.

\bibitem{zhu2017toward}
J.-Y. Zhu, R.~Zhang, D.~Pathak, T.~Darrell, A.~A. Efros, O.~Wang, and
  E.~Shechtman, ``Toward multimodal image-to-image translation,'' in
  \emph{Advances in Neural Information Processing Systems}, 2017, pp. 465--476.

\bibitem{tomei2018art2real}
M.~Tomei, M.~Cornia, L.~Baraldi, and R.~Cucchiara, ``{A}rt2{R}eal: {U}nfolding
  the reality of artworks via semantically-aware image-to-image translation,''
  \emph{arXiv preprint arXiv:1811.10666}, 2018.

\bibitem{liu2017unsupervised}
M.-Y. Liu, T.~Breuel, and J.~Kautz, ``Unsupervised image-to-image translation
  networks,'' in \emph{Advances in Neural Information Processing Systems},
  2017, pp. 700--708.

\bibitem{isola2017image}
P.~Isola, J.-Y. Zhu, T.~Zhou, and A.~A. Efros, ``Image-to-image translation
  with conditional adversarial networks,'' in \emph{Proceedings of the IEEE
  conference on computer vision and pattern recognition}, 2017, pp. 1125--1134.

\bibitem{choi2018stargan}
Y.~Choi, M.~Choi, M.~Kim, J.-W. Ha, S.~Kim, and J.~Choo, ``Star{GAN}: {U}nified
  generative adversarial networks for multi-domain image-to-image
  translation,'' in \emph{Proceedings of the IEEE Conference on Computer Vision
  and Pattern Recognition}, 2018, pp. 8789--8797.

\bibitem{ma2018gan}
S.~Ma, J.~Fu, C.~Wen~Chen, and T.~Mei, ``{DA-GAN}: {I}nstance-level image
  translation by deep attention generative adversarial networks,'' in
  \emph{Proceedings of the IEEE Conference on Computer Vision and Pattern
  Recognition}, 2018, pp. 5657--5666.

\bibitem{ledig2017photo}
C.~Ledig, L.~Theis, F.~Husz{\'a}r, J.~Caballero, A.~Cunningham, A.~Acosta,
  A.~Aitken, A.~Tejani, J.~Totz, Z.~Wang \emph{et~al.}, ``Photo-realistic
  single image super-resolution using a generative adversarial network,'' in
  \emph{2017 IEEE Conference on Computer Vision and Pattern Recognition}.\hskip
  1em plus 0.5em minus 0.4em\relax IEEE, 2017, pp. 105--114.

\bibitem{wang2018esrgan}
X.~Wang, K.~Yu, S.~Wu, J.~Gu, Y.~Liu, C.~Dong, Y.~Qiao, and C.~Change~Loy,
  ``{ESRGAN}: {E}nhanced super-resolution generative adversarial networks,'' in
  \emph{European Conference on Computer Vision Workshop}, 2018.

\bibitem{mahapatra2017image}
D.~Mahapatra, B.~Bozorgtabar, S.~Hewavitharanage, and R.~Garnavi, ``Image super
  resolution using generative adversarial networks and local saliency maps for
  retinal image analysis,'' in \emph{International Conference on Medical Image
  Computing and Computer-Assisted Intervention}.\hskip 1em plus 0.5em minus
  0.4em\relax Springer, 2017, pp. 382--390.

\bibitem{mahapatra2019image}
D.~Mahapatra, B.~Bozorgtabar, and R.~Garnavi, ``Image super-resolution using
  progressive generative adversarial networks for medical image analysis,''
  \emph{Computerized Medical Imaging and Graphics}, vol.~71, pp. 30--39, 2019.

\bibitem{yu2018generative}
J.~Yu, Z.~Lin, J.~Yang, X.~Shen, X.~Lu, and T.~S. Huang, ``Generative image
  inpainting with contextual attention,'' \emph{arXiv preprint
  arXiv:1801.07892}, 2018.

\bibitem{yeh2017semantic}
R.~A. Yeh, C.~Chen, T.~Yian~Lim, A.~G. Schwing, M.~Hasegawa-Johnson, and M.~N.
  Do, ``Semantic image inpainting with deep generative models,'' in
  \emph{Proceedings of the IEEE Conference on Computer Vision and Pattern
  Recognition}, 2017, pp. 5485--5493.

\bibitem{dolhansky2018eye}
B.~Dolhansky and C.~Canton~Ferrer, ``Eye in-painting with exemplar generative
  adversarial networks,'' in \emph{Proceedings of the IEEE Conference on
  Computer Vision and Pattern Recognition}, 2018, pp. 7902--7911.

\bibitem{chen2018high}
Z.~Chen, S.~Nie, T.~Wu, and C.~G. Healey, ``High resolution face completion
  with multiple controllable attributes via fully end-to-end progressive
  generative adversarial networks,'' \emph{arXiv preprint arXiv:1801.07632},
  2018.

\bibitem{li2017generative}
Y.~Li, S.~Liu, J.~Yang, and M.-H. Yang, ``Generative face completion,'' in
  \emph{Proceedings of the IEEE Conference on Computer Vision and Pattern
  Recognition}, 2017, pp. 3911--3919.

\bibitem{kossaifi2018gagan}
J.~Kossaifi, L.~Tran, Y.~Panagakis, and M.~Pantic, ``{GANGAN}: {G}eometry-aware
  generative adversarial networks,'' in \emph{Proceedings of the IEEE
  Conference on Computer Vision and Pattern Recognition}, 2018, pp. 878--887.

\bibitem{dai2018adversarial}
Q.~Dai, Q.~Li, J.~Tang, and D.~Wang, ``Adversarial network embedding,'' in
  \emph{Thirty-Second AAAI Conference on Artificial Intelligence}, 2018.

\bibitem{kodali2017convergence}
N.~Kodali, J.~Abernethy, J.~Hays, and Z.~Kira, ``On convergence and stability
  of {GANs},'' \emph{arXiv preprint arXiv:1705.07215}, 2017.

\bibitem{li2017dualing}
Y.~Li, A.~Schwing, K.-C. Wang, and R.~Zemel, ``Dualing {GAN}s,'' in
  \emph{Advances in Neural Information Processing Systems}, 2017, pp.
  5606--5616.

\bibitem{borji2019pros}
A.~Borji, ``Pros and cons of {GAN} evaluation measures,'' \emph{Computer Vision
  and Image Understanding}, vol. 179, pp. 41--65, 2019.

\bibitem{xu2018empirical}
Q.~Xu, G.~Huang, Y.~Yuan, C.~Guo, Y.~Sun, F.~Wu, and K.~Weinberger, ``An
  empirical study on evaluation metrics of generative adversarial networks,''
  \emph{arXiv preprint arXiv:1806.07755}, 2018.

\bibitem{gulrajani2017improved}
I.~Gulrajani, F.~Ahmed, M.~Arjovsky, V.~Dumoulin, and A.~C. Courville,
  ``Improved training of wasserstein {GANs},'' in \emph{Advances in Neural
  Information Processing Systems}, 2017, pp. 5767--5777.

\bibitem{heusel2017gans}
M.~Heusel, H.~Ramsauer, T.~Unterthiner, B.~Nessler, and S.~Hochreiter, ``{GANs}
  trained by a two time-scale update rule converge to a local nash
  equilibrium,'' in \emph{Advances in Neural Information Processing Systems},
  2017, pp. 6626--6637.

\bibitem{gretton2012kernel}
A.~Gretton, K.~M. Borgwardt, M.~J. Rasch, B.~Sch{\"o}lkopf, and A.~Smola, ``A
  kernel two-sample test,'' \emph{Journal of Machine Learning Research},
  vol.~13, no. Mar, pp. 723--773, 2012.

\bibitem{wang2018use}
Z.~Wang, G.~Healy, A.~F. Smeaton, and T.~E. Ward, ``Use of neural signals to
  evaluate the quality of generative adversarial network performance in facial
  image generation,'' \emph{Cognitive Computation}, vol.~12, no.~1, pp. 13--24,
  2020.

\bibitem{wang2019neuroscore}
Z.~Wang, Q.~She, A.~F. Smeaton, T.~E. Ward, and G.~Healy,
  ``{Synthetic-Neuroscore}: {U}sing a neuro-{AI} interface for evaluating
  generative adversarial networks,'' \emph{Neurocomputing}, 2020.

\bibitem{barratt2018note}
S.~Barratt and R.~Sharma, ``A note on the inception score,'' \emph{arXiv
  preprint arXiv:1801.01973}, 2018.

\bibitem{theis2015note}
L.~Theis, A.~v.~d. Oord, and M.~Bethge, ``A note on the evaluation of
  generative models,'' \emph{arXiv preprint arXiv:1511.01844}, 2015.

\bibitem{jolicoeur2018relativistic}
A.~Jolicoeur-Martineau, ``The relativistic discriminator: a key element missing
  from standard {GAN},'' \emph{arXiv preprint arXiv:1807.00734}, 2018.

\bibitem{kurach2018gan}
K.~Kurach, M.~Lucic, X.~Zhai, M.~Michalski, and S.~Gelly, ``The {GAN}
  landscape: {L}osses, architectures, regularization, and normalization,''
  \emph{arXiv preprint arXiv:1807.04720}, 2018.

\bibitem{yu2015lsun}
F.~Yu, A.~Seff, Y.~Zhang, S.~Song, T.~Funkhouser, and J.~Xiao, ``Lsun:
  Construction of a large-scale image dataset using deep learning with humans
  in the loop,'' \emph{arXiv preprint arXiv:1506.03365}, 2015.

\bibitem{liu2018large}
Z.~Liu, P.~Luo, X.~Wang, and X.~Tang, ``Large-scale celebfaces attributes
  (celeba) dataset,'' \emph{Retrieved August}, vol.~15, p. 2018, 2018.

\bibitem{krizhevsky2009learning}
A.~Krizhevsky and G.~Hinton, ``Learning multiple layers of features from tiny
  images,'' Citeseer, Tech. Rep., 2009.

\bibitem{yoshida2017spectral}
Y.~Yoshida and T.~Miyato, ``Spectral norm regularization for improving the
  generalizability of deep learning,'' \emph{arXiv preprint arXiv:1705.10941},
  2017.

\bibitem{hitawala2018comparative}
S.~Hitawala, ``Comparative study on generative adversarial networks,''
  \emph{arXiv preprint arXiv:1801.04271}, 2018.

\bibitem{wang2017generative}
K.~Wang, C.~Gou, Y.~Duan, Y.~Lin, X.~Zheng, and F.-Y. Wang, ``Generative
  adversarial networks: introduction and outlook,'' \emph{IEEE/CAA Journal of
  Automatica Sinica}, vol.~4, no.~4, pp. 588--598, 2017.

\bibitem{creswell2018generative}
A.~Creswell, T.~White, V.~Dumoulin, K.~Arulkumaran, B.~Sengupta, and A.~A.
  Bharath, ``Generative adversarial networks: {A}n overview,'' \emph{IEEE
  Signal Processing Magazine}, vol.~35, no.~1, pp. 53--65, 2018.

\bibitem{hong2019generative}
Y.~Hong, U.~Hwang, J.~Yoo, and S.~Yoon, ``How generative adversarial networks
  and their variants work: {A}n overview,'' \emph{ACM Computing Surveys
  (CSUR)}, vol.~52, no.~1, p.~10, 2019.

\bibitem{doersch2016tutorial}
C.~Doersch, ``Tutorial on variational autoencoders,'' \emph{arXiv preprint
  arXiv:1606.05908}, 2016.

\bibitem{kingma2013auto}
D.~P. Kingma and M.~Welling, ``Auto-encoding variational {B}ayes,'' \emph{arXiv
  preprint arXiv:1312.6114}, 2013.

\bibitem{radford2015unsupervised}
A.~Radford, L.~Metz, and S.~Chintala, ``Unsupervised representation learning
  with deep convolutional generative adversarial networks,'' \emph{arXiv
  preprint arXiv:1511.06434}, 2015.

\bibitem{berthelot2017began}
D.~Berthelot, T.~Schumm, and L.~Metz, ``{BEGAN}: {B}oundary equilibrium
  generative adversarial networks,'' \emph{arXiv preprint arXiv:1703.10717},
  2017.

\bibitem{karras2017progressive}
T.~Karras, T.~Aila, S.~Laine, and J.~Lehtinen, ``Progressive growing of {GAN}s
  for improved quality, stability, and variation,'' \emph{arXiv preprint
  arXiv:1710.10196}, 2017.

\bibitem{iizuka2017globally}
S.~Iizuka, E.~Simo-Serra, and H.~Ishikawa, ``Globally and locally consistent
  image completion,'' \emph{ACM Transactions on Graphics (ToG)}, vol.~36,
  no.~4, p. 107, 2017.

\bibitem{reed2016generative}
S.~Reed, Z.~Akata, X.~Yan, L.~Logeswaran, B.~Schiele, and H.~Lee, ``Generative
  adversarial text to image synthesis,'' \emph{arXiv preprint
  arXiv:1605.05396}, 2016.

\bibitem{lecun1998gradient}
Y.~LeCun, L.~Bottou, Y.~Bengio, P.~Haffner \emph{et~al.}, ``Gradient-based
  learning applied to document recognition,'' \emph{Proceedings of the IEEE},
  vol.~86, no.~11, pp. 2278--2324, 1998.

\bibitem{glorot2011deep}
X.~Glorot, A.~Bordes, and Y.~Bengio, ``Deep sparse rectifier neural networks,''
  in \emph{Proceedings of the fourteenth international conference on artificial
  intelligence and statistics}, 2011, pp. 315--323.

\bibitem{odena2016semi}
A.~Odena, ``Semi-supervised learning with generative adversarial networks,''
  \emph{arXiv preprint arXiv:1606.01583}, 2016.

\bibitem{donahue2016adversarial}
J.~Donahue, P.~Kr{\"a}henb{\"u}hl, and T.~Darrell, ``Adversarial feature
  learning,'' \emph{arXiv preprint arXiv:1605.09782}, 2016.

\bibitem{mirza2014conditional}
M.~Mirza and S.~Osindero, ``Conditional generative adversarial nets,''
  \emph{arXiv preprint arXiv:1411.1784}, 2014.

\bibitem{chen2016infogan}
X.~Chen, Y.~Duan, R.~Houthooft, J.~Schulman, I.~Sutskever, and P.~Abbeel,
  ``{InfoGAN}: {I}nterpretable representation learning by information
  maximizing generative adversarial nets,'' in \emph{Advances in neural
  information processing systems}, 2016, pp. 2172--2180.

\bibitem{paysan20093d}
P.~Paysan, R.~Knothe, B.~Amberg, S.~Romdhani, and T.~Vetter, ``A 3{D} face
  model for pose and illumination invariant face recognition,'' in \emph{2009
  Sixth IEEE International Conference on Advanced Video and Signal Based
  Surveillance}.\hskip 1em plus 0.5em minus 0.4em\relax IEEE, 2009, pp.
  296--301.

\bibitem{aubry2014seeing}
M.~Aubry, D.~Maturana, A.~A. Efros, B.~C. Russell, and J.~Sivic, ``Seeing 3{D}
  chairs: {E}xemplar part-based 2{D}-3{D} alignment using a large dataset of
  {CAD} models,'' in \emph{Proceedings of the IEEE conference on computer
  vision and pattern recognition}.\hskip 1em plus 0.5em minus 0.4em\relax IEEE,
  2014, pp. 3762--3769.

\bibitem{denton2015deep}
E.~L. Denton, S.~Chintala, R.~Fergus \emph{et~al.}, ``Deep generative image
  models using a laplacian pyramid of adversarial networks,'' in \emph{Advances
  in neural information processing systems}, 2015, pp. 1486--1494.

\bibitem{burt1983laplacian}
P.~Burt and E.~Adelson, ``The {L}aplacian pyramid as a compact image code,''
  \emph{IEEE Transactions on communications}, vol.~31, no.~4, pp. 532--540,
  1983.

\bibitem{zeiler2014visualizing}
M.~D. Zeiler and R.~Fergus, ``Visualizing and understanding convolutional
  networks,'' in \emph{European conference on computer vision}.\hskip 1em plus
  0.5em minus 0.4em\relax Springer, 2014, pp. 818--833.

\bibitem{deng2009imagenet}
J.~Deng, W.~Dong, R.~Socher, L.-J. Li, K.~Li, and L.~Fei-Fei, ``{ImageNet}: {A}
  large-scale hierarchical image database,'' in \emph{2009 IEEE conference on
  computer vision and pattern recognition}.\hskip 1em plus 0.5em minus
  0.4em\relax IEEE, 2009, pp. 248--255.

\bibitem{zhao2016energy}
J.~Zhao, M.~Mathieu, and Y.~LeCun, ``Energy-based generative adversarial
  network,'' \emph{arXiv preprint arXiv:1609.03126}, 2016.

\bibitem{rusu2016progressive}
A.~A. Rusu, N.~C. Rabinowitz, G.~Desjardins, H.~Soyer, J.~Kirkpatrick,
  K.~Kavukcuoglu, R.~Pascanu, and R.~Hadsell, ``Progressive neural networks,''
  \emph{arXiv preprint arXiv:1606.04671}, 2016.

\bibitem{brock2018large}
A.~Brock, J.~Donahue, and K.~Simonyan, ``Large scale {GAN} training for high
  fidelity natural image synthesis,'' \emph{arXiv preprint arXiv:1809.11096},
  2018.

\bibitem{vaswani2017attention}
A.~Vaswani, N.~Shazeer, N.~Parmar, J.~Uszkoreit, L.~Jones, A.~N. Gomez,
  {\L}.~Kaiser, and I.~Polosukhin, ``Attention is all you need,'' in
  \emph{Advances in Neural Information Processing Systems}, 2017, pp.
  5998--6008.

\bibitem{zhang2018self}
H.~Zhang, I.~Goodfellow, D.~Metaxas, and A.~Odena, ``Self-attention generative
  adversarial networks,'' \emph{arXiv preprint arXiv:1805.08318}, 2018.

\bibitem{miyato2018spectral}
T.~Miyato, T.~Kataoka, M.~Koyama, and Y.~Yoshida, ``Spectral normalization for
  generative adversarial networks,'' \emph{arXiv preprint arXiv:1802.05957},
  2018.

\bibitem{kaneko2019label}
T.~Kaneko, Y.~Ushiku, and T.~Harada, ``Label-noise robust generative
  adversarial networks,'' in \emph{Proceedings of the IEEE Conference on
  Computer Vision and Pattern Recognition}, 2019, pp. 2467--2476.

\bibitem{Daras_2020_CVPR}
G.~Daras, A.~Odena, H.~Zhang, and A.~G. Dimakis, ``Your local {GAN}: Designing
  two dimensional local attention mechanisms for generative models,'' in
  \emph{IEEE/CVF Conference on Computer Vision and Pattern Recognition (CVPR)},
  June 2020.

\bibitem{Gong_2019_ICCV}
X.~Gong, S.~Chang, Y.~Jiang, and Z.~Wang, ``Autogan: Neural architecture search
  for generative adversarial networks,'' in \emph{The IEEE International
  Conference on Computer Vision (ICCV)}, October 2019.

\bibitem{karnewar2020msg}
A.~Karnewar and O.~Wang, ``{MSG}-{GAN}: {M}ulti-scale gradients for generative
  adversarial networks,'' in \emph{Proceedings of the IEEE Conference on
  Computer Vision and Pattern Recognition}, 2020, pp. 7799--7808.

\bibitem{MAtowards}
M.~Arjovsky and L.~Bottou, ``Towards principled methods for training generative
  adversarial networks,'' \emph{arXiv preprint arXiv:1701.04862}, 2017.

\bibitem{arjovsky2017wasserstein}
M.~Arjovsky, S.~Chintala, and L.~Bottou, ``Wasserstein {GAN},'' \emph{arXiv
  preprint arXiv:1701.07875}, 2017.

\bibitem{rubner2000earth}
Y.~Rubner, C.~Tomasi, and L.~J. Guibas, ``The earth mover's distance as a
  metric for image retrieval,'' \emph{International journal of computer
  vision}, vol.~40, no.~2, pp. 99--121, 2000.

\bibitem{mao2017least}
X.~Mao, Q.~Li, H.~Xie, R.~Y. Lau, Z.~Wang, and S.~P. Smolley, ``Least squares
  generative adversarial networks,'' in \emph{2017 IEEE International
  Conference on Computer Vision}.\hskip 1em plus 0.5em minus 0.4em\relax IEEE,
  2017, pp. 2813--2821.

\bibitem{liu2013online}
C.-L. Liu, F.~Yin, D.-H. Wang, and Q.-F. Wang, ``Online and offline handwritten
  chinese character recognition: {B}enchmarking on new databases,''
  \emph{Pattern Recognition}, vol.~46, no.~1, pp. 155--162, 2013.

\bibitem{nowozin2016f}
S.~Nowozin, B.~Cseke, and R.~Tomioka, ``f-{GAN}: Training generative neural
  samplers using variational divergence minimization,'' in \emph{Advances in
  neural information processing systems}, 2016, pp. 271--279.

\bibitem{nielsen2013chi}
F.~Nielsen and R.~Nock, ``On the {C}hi square and higher-order {C}hi distances
  for approximating f-divergences,'' \emph{IEEE Signal Processing Letters},
  vol.~21, no.~1, pp. 10--13, 2013.

\bibitem{hiriart2012fundamentals}
J.-B. Hiriart-Urruty and C.~Lemar{\'e}chal, \emph{Fundamentals of convex
  analysis}.\hskip 1em plus 0.5em minus 0.4em\relax Springer Science \&
  Business Media, 2012.

\bibitem{metz2016unrolled}
L.~Metz, B.~Poole, D.~Pfau, and J.~Sohl-Dickstein, ``Unrolled generative
  adversarial networks,'' \emph{arXiv preprint arXiv:1611.02163}, 2016.

\bibitem{qi2017loss}
G.-J. Qi, ``Loss-sensitive generative adversarial networks on lipschitz
  densities,'' \emph{arXiv preprint arXiv:1701.06264}, 2017.

\bibitem{netzer2011reading}
Y.~Netzer, T.~Wang, A.~Coates, A.~Bissacco, B.~Wu, and A.~Y. Ng, ``Reading
  digits in natural images with unsupervised feature learning,'' 2011.

\bibitem{che2016mode}
T.~Che, Y.~Li, A.~P. Jacob, Y.~Bengio, and W.~Li, ``Mode regularized generative
  adversarial networks,'' \emph{arXiv preprint arXiv:1612.02136}, 2016.

\bibitem{lim2017geometric}
J.~H. Lim and J.~C. Ye, ``Geometric {GAN},'' \emph{arXiv preprint
  arXiv:1705.02894}, 2017.

\bibitem{marron2007distance}
J.~S. Marron, M.~J. Todd, and J.~Ahn, ``Distance-weighted discrimination,''
  \emph{Journal of the American Statistical Association}, vol. 102, no. 480,
  pp. 1267--1271, 2007.

\bibitem{carmichael2017geometric}
I.~Carmichael and J.~Marron, ``Geometric insights into support vector machine
  behavior using the {KKT} conditions,'' \emph{arXiv preprint
  arXiv:1704.00767}, 2017.

\bibitem{ahn2010maximal}
J.~Ahn and J.~Marron, ``The maximal data piling direction for discrimination,''
  \emph{Biometrika}, vol.~97, no.~1, pp. 254--259, 2010.

\bibitem{sriperumbudur2009integral}
B.~K. Sriperumbudur, K.~Fukumizu, A.~Gretton, B.~Sch{\"o}lkopf, and G.~R.
  Lanckriet, ``On integral probability metrics,$\phi$-divergences and binary
  classification,'' \emph{arXiv preprint arXiv:0901.2698}, 2009.

\bibitem{muller1997integral}
A.~M{\"u}ller, ``Integral probability metrics and their generating classes of
  functions,'' \emph{Advances in Applied Probability}, vol.~29, no.~2, pp.
  429--443, 1997.

\bibitem{donchev1998stability}
T.~Donchev and E.~Farkhi, ``Stability and euler approximation of one-sided
  lipschitz differential inclusions,'' \emph{SIAM journal on control and
  optimization}, vol.~36, no.~2, pp. 780--796, 1998.

\bibitem{armijo1966minimization}
L.~Armijo, ``Minimization of functions having lipschitz continuous first
  partial derivatives,'' \emph{Pacific Journal of mathematics}, vol.~16, no.~1,
  pp. 1--3, 1966.

\bibitem{goldstein1977optimization}
A.~Goldstein, ``Optimization of lipschitz continuous functions,''
  \emph{Mathematical Programming}, vol.~13, no.~1, pp. 14--22, 1977.

\bibitem{coates2011analysis}
A.~Coates, A.~Ng, and H.~Lee, ``An analysis of single-layer networks in
  unsupervised feature learning,'' in \emph{Proceedings of the fourteenth
  international conference on artificial intelligence and statistics}, 2011,
  pp. 215--223.

\bibitem{ioffe2015batch}
S.~Ioffe and C.~Szegedy, ``Batch normalization: {A}ccelerating deep network
  training by reducing internal covariate shift,'' \emph{arXiv preprint
  arXiv:1502.03167}, 2015.

\bibitem{salimans2016weight}
T.~Salimans and D.~P. Kingma, ``Weight normalization: {A} simple
  reparameterization to accelerate training of deep neural networks,'' in
  \emph{Advances in neural information processing systems}, 2016, pp. 901--909.

\bibitem{ba2016layer}
J.~L. Ba, J.~R. Kiros, and G.~E. Hinton, ``Layer normalization,'' \emph{arXiv
  preprint arXiv:1607.06450}, 2016.

\bibitem{brock2016neural}
A.~Brock, T.~Lim, J.~M. Ritchie, and N.~Weston, ``Neural photo editing with
  introspective adversarial networks,'' \emph{arXiv preprint arXiv:1609.07093},
  2016.

\bibitem{xiangli2020real}
Y.~Xiangli, Y.~Deng, B.~Dai, C.~C. Loy, and D.~Lin, ``Real or not real, that is
  the question,'' \emph{arXiv preprint arXiv:2002.05512}, 2020.

\bibitem{Park_2019_CVPR}
S.~W. Park and J.~Kwon, ``Sphere generative adversarial network based on
  geometric moment matching,'' in \emph{The IEEE Conference on Computer Vision
  and Pattern Recognition (CVPR)}, June 2019.

\bibitem{Chen_2019_CVPR}
T.~Chen, X.~Zhai, M.~Ritter, M.~Lucic, and N.~Houlsby, ``Self-supervised gans
  via auxiliary rotation loss,'' in \emph{The IEEE Conference on Computer
  Vision and Pattern Recognition (CVPR)}, June 2019.

\bibitem{mavratzakis2016emotional}
A.~Mavratzakis, C.~Herbert, and P.~Walla, ``Emotional facial expressions evoke
  faster orienting responses, but weaker emotional responses at neural and
  behavioural levels compared to scenes: {A} simultaneous {EEG} and facial
  {EMG} study,'' \emph{Neuroimage}, vol. 124, pp. 931--946, 2016.

\bibitem{polich2007updating}
J.~Polich, ``Updating {P}300: {A}n integrative theory of {P3a} and {P3b},''
  \emph{Clinical neurophysiology}, vol. 118, no.~10, pp. 2128--2148, 2007.

\bibitem{delaney2019synthesis}
A.~M. Delaney, E.~Brophy, and T.~E. Ward, ``Synthesis of realistic {ECG} using
  generative adversarial networks,'' \emph{arXiv preprint arXiv:1909.09150},
  2019.

\bibitem{barnes2009patchmatch}
C.~Barnes, E.~Shechtman, A.~Finkelstein, and D.~B. Goldman, ``{PatchMatch}: {A}
  randomized correspondence algorithm for structural image editing,'' in
  \emph{ACM Transactions on Graphics (ToG)}, vol.~28, no.~3.\hskip 1em plus
  0.5em minus 0.4em\relax ACM, 2009, p.~24.

\bibitem{huang2014image}
J.-B. Huang, S.~B. Kang, N.~Ahuja, and J.~Kopf, ``Image completion using planar
  structure guidance,'' \emph{ACM Transactions on graphics (TOG)}, vol.~33,
  no.~4, pp. 1--10, 2014.

\bibitem{long2015fully}
J.~Long, E.~Shelhamer, and T.~Darrell, ``Fully convolutional networks for
  semantic segmentation,'' in \emph{Proceedings of the IEEE conference on
  computer vision and pattern recognition}, 2015, pp. 3431--3440.

\bibitem{yang2016object}
J.~Yang, B.~Price, S.~Cohen, H.~Lee, and M.-H. Yang, ``Object contour detection
  with a fully convolutional encoder-decoder network,'' in \emph{Proceedings of
  the IEEE conference on computer vision and pattern recognition}, 2016, pp.
  193--202.

\bibitem{lutz2018alphagan}
S.~Lutz, K.~Amplianitis, and A.~Smolic, ``Alpha{GAN}:{G}enerative adversarial
  networks for natural image matting,'' \emph{arXiv preprint arXiv:1807.10088},
  2018.

\bibitem{cho2016natural}
D.~Cho, Y.-W. Tai, and I.~Kweon, ``Natural image matting using deep
  convolutional neural networks,'' in \emph{European Conference on Computer
  Vision}.\hskip 1em plus 0.5em minus 0.4em\relax Springer, 2016, pp. 626--643.

\bibitem{aksoy2017designing}
Y.~Aksoy, T.~Ozan~Aydin, and M.~Pollefeys, ``Designing effective inter-pixel
  information flow for natural image matting,'' in \emph{Proceedings of the
  IEEE Conference on Computer Vision and Pattern Recognition}, 2017, pp.
  29--37.

\bibitem{xu2017deep}
N.~Xu, B.~Price, S.~Cohen, and T.~Huang, ``Deep image matting,'' in
  \emph{Proceedings of the IEEE Conference on Computer Vision and Pattern
  Recognition}, 2017, pp. 2970--2979.

\bibitem{kim2017learning}
T.~Kim, M.~Cha, H.~Kim, J.~K. Lee, and J.~Kim, ``Learning to discover
  cross-domain relations with generative adversarial networks,'' in
  \emph{Proceedings of the 34th International Conference on Machine
  Learning-Volume 70}.\hskip 1em plus 0.5em minus 0.4em\relax JMLR. org, 2017,
  pp. 1857--1865.

\bibitem{yi2017dualgan}
Z.~Yi, H.~Zhang, P.~Tan, and M.~Gong, ``{DualGAN}: {U}nsupervised dual learning
  for image-to-image translation,'' in \emph{Proceedings of the IEEE
  international conference on computer vision}, 2017, pp. 2849--2857.

\bibitem{tulyakov2018mocogan}
S.~Tulyakov, M.-Y. Liu, X.~Yang, and J.~Kautz, ``{MoCoGAN}: {D}ecomposing
  motion and content for video generation,'' in \emph{Proceedings of the IEEE
  conference on computer vision and pattern recognition}, 2018, pp. 1526--1535.

\bibitem{clark2019adversarial}
A.~Clark, J.~Donahue, and K.~Simonyan, ``Adversarial video generation on
  complex datasets,'' \emph{arXiv preprint arXiv:1907.06571}, 2019.

\bibitem{kahembwe2019lower}
E.~Kahembwe and S.~Ramamoorthy, ``Lower dimensional kernels for video
  discriminators,'' \emph{arXiv preprint arXiv:1912.08860}, 2019.

\bibitem{turing2009computing}
A.~M. Turing, ``Computing machinery and intelligence,'' in \emph{Parsing the
  Turing Test}.\hskip 1em plus 0.5em minus 0.4em\relax Springer, 2009, pp.
  23--65.

\bibitem{mccloskey1989catastrophic}
M.~McCloskey and N.~J. Cohen, ``Catastrophic interference in connectionist
  networks: {T}he sequential learning problem,'' in \emph{Psychology of
  learning and motivation}.\hskip 1em plus 0.5em minus 0.4em\relax Elsevier,
  1989, vol.~24, pp. 109--165.

\bibitem{mandal2019out}
D.~Mandal, S.~Narayan, S.~K. Dwivedi, V.~Gupta, S.~Ahmed, F.~S. Khan, and
  L.~Shao, ``Out-of-distribution detection for generalized zero-shot action
  recognition,'' in \emph{Proceedings of the IEEE Conference on Computer Vision
  and Pattern Recognition}, 2019, pp. 9985--9993.

\bibitem{shin2017continual}
H.~Shin, J.~K. Lee, J.~Kim, and J.~Kim, ``Continual learning with deep
  generative replay,'' in \emph{Advances in Neural Information Processing
  Systems}, 2017, pp. 2990--2999.

\bibitem{lopez2016revisiting}
D.~Lopez-Paz and M.~Oquab, ``Revisiting classifier two-sample tests,''
  \emph{arXiv preprint arXiv:1610.06545}, 2016.

\bibitem{shmelkov2018good}
K.~Shmelkov, C.~Schmid, and K.~Alahari, ``How good is my gan?'' in
  \emph{Proceedings of the European Conference on Computer Vision (ECCV)},
  2018, pp. 213--229.

\bibitem{im2018quantitatively}
D.~J. Im, H.~Ma, G.~Taylor, and K.~Branson, ``Quantitatively evaluating {GAN}s
  with divergences proposed for training,'' \emph{arXiv preprint
  arXiv:1803.01045}, 2018.

\bibitem{song2019bridging}
J.~Song and S.~Ermon, ``Bridging the gap between $ f $-{GAN}s and {Wasserstein}
  {GAN}s,'' \emph{arXiv preprint arXiv:1910.09779}, 2019.

\bibitem{mao2018effectiveness}
X.~Mao, Q.~Li, H.~Xie, R.~Y. Lau, Z.~Wang, and S.~P. Smolley, ``On the
  effectiveness of least squares generative adversarial networks,'' \emph{IEEE
  transactions on pattern analysis and machine intelligence}, vol.~41, no.~12,
  pp. 2947--2960, 2018.

\bibitem{luo2018multivariate}
Y.~Luo, X.~Cai, Y.~Zhang, J.~Xu \emph{et~al.}, ``Multivariate time series
  imputation with generative adversarial networks,'' in \emph{Advances in
  Neural Information Processing Systems}, 2018, pp. 1596--1607.

\bibitem{yu2017seqgan}
L.~Yu, W.~Zhang, J.~Wang, and Y.~Yu, ``Seq{GAN}: Sequence generative
  adversarial nets with policy gradient,'' in \emph{Thirty-First AAAI
  Conference on Artificial Intelligence}, 2017.

\bibitem{bahdanau2016actor}
D.~Bahdanau, P.~Brakel, K.~Xu, A.~Goyal, R.~Lowe, J.~Pineau, A.~Courville, and
  Y.~Bengio, ``An actor-critic algorithm for sequence prediction,'' \emph{arXiv
  preprint arXiv:1607.07086}, 2016.

\bibitem{li2016deep}
J.~Li, W.~Monroe, A.~Ritter, M.~Galley, J.~Gao, and D.~Jurafsky, ``Deep
  reinforcement learning for dialogue generation,'' \emph{arXiv preprint
  arXiv:1606.01541}, 2016.

\bibitem{hsu2018learning}
C.-C. Hsu, C.-Y. Lee, and Y.-X. Zhuang, ``Learning to detect fake face images
  in the wild,'' in \emph{2018 International Symposium on Computer, Consumer
  and Control (IS3C)}.\hskip 1em plus 0.5em minus 0.4em\relax IEEE, 2018, pp.
  388--391.

\bibitem{afchar2018mesonet}
D.~Afchar, V.~Nozick, J.~Yamagishi, and I.~Echizen, ``Meso{N}et: {A} compact
  facial video forgery detection network,'' in \emph{2018 IEEE International
  Workshop on Information Forensics and Security (WIFS)}.\hskip 1em plus 0.5em
  minus 0.4em\relax IEEE, 2018, pp. 1--7.

\end{thebibliography}

\clearpage
\appendix
\section{Search Strategy and Results} \label{app:search}
A review of the literature was performed to identify research work describing GANs. Papers were first identified through manual search of the online datasets (Google Scholar and IEEE Xplore) through use of the keyword ``generative adversarial networks''. Secondly papers related to computer vision were manually selected. This search concluded 10th June 2020.

A total of 463 papers describing GANs related to computer vision were identified. The earliest paper was in 2014~\cite{goodfellow2014generative}.  Papers were classified into three categories regarding their repository, namely conference, arXiv and journal. More than half the papers were presented at conferences (267, 57.7\%). The rest were journal articles (116, 25.1\%) and arXiv pre-prints (80, 17.3\%). 

\begin{figure}[ht!]
	\centering
	\includegraphics[width=.6\textwidth]{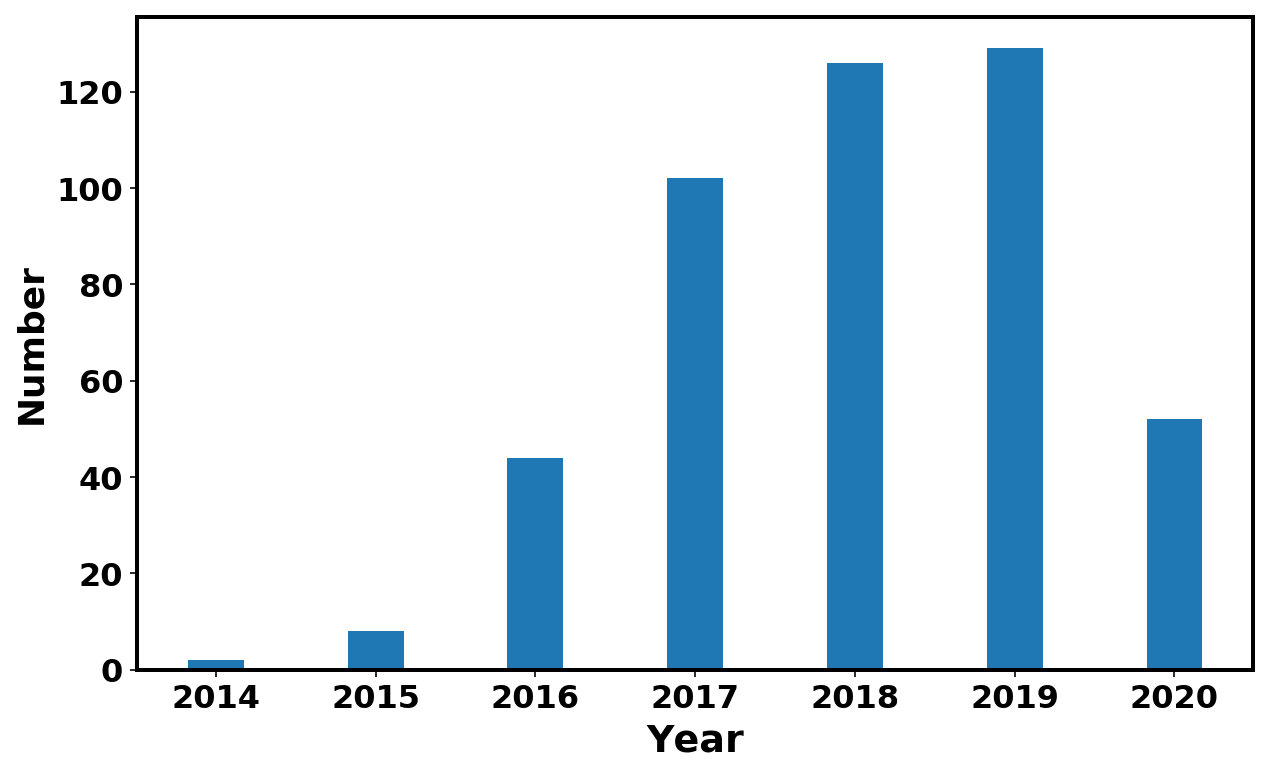}
	\caption{Number of papers in each year from 2014 to 10th June 2020.}
	\label{chap05-fig:paper_stats_mix}
\end{figure}

\begin{figure}[ht!]
	\centering
	\includegraphics[width=.6\textwidth]{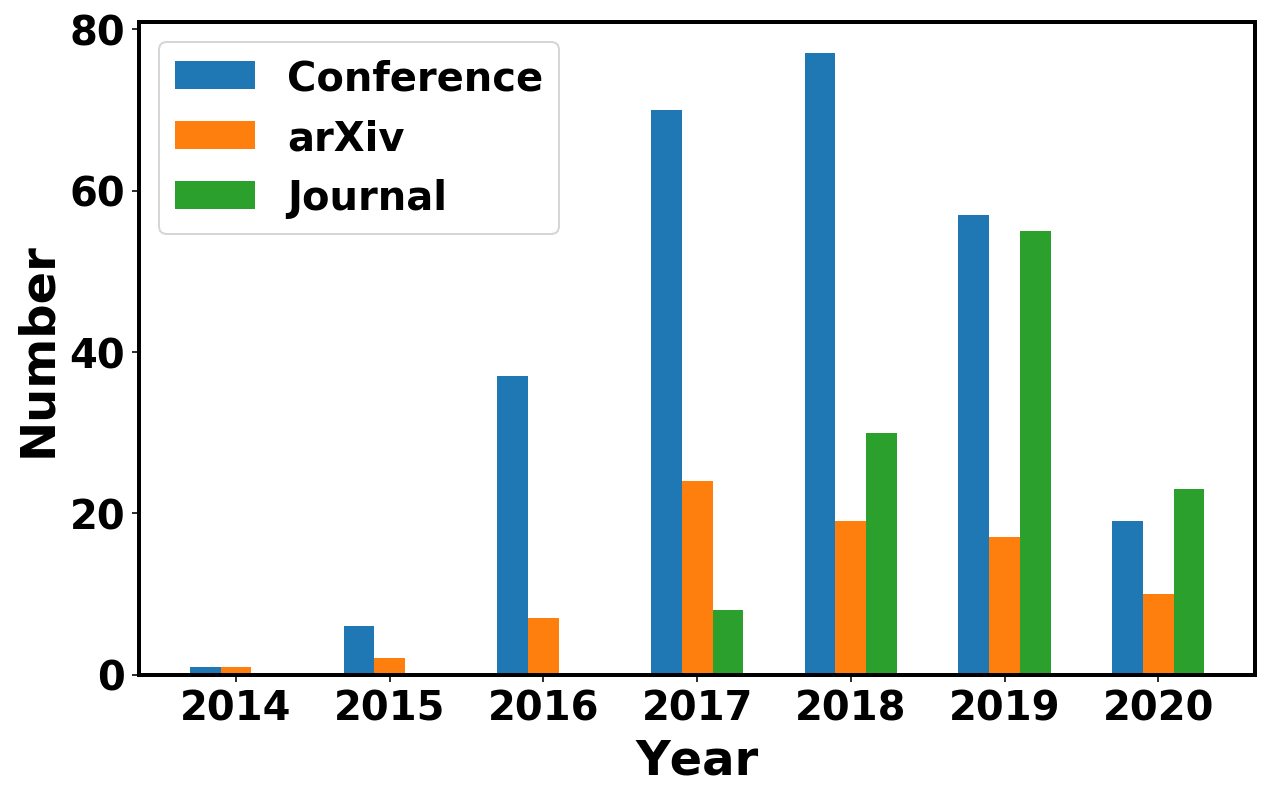}
	\caption{Categories of papers from 2014 to 10th June 2020. Papers are categorized as conference, arXiv and journal.}
	\label{chap05-fig:paper_stats_three}
\end{figure}

Details of searched papers are included in Fig.~\ref{chap05-fig:paper_stats_mix} and Fig.~\ref{chap05-fig:paper_stats_three}. Figure~\ref{chap05-fig:paper_stats_mix} 
illustrates the number of papers in each year from 2014 to 2020. It can be seen that the number of papers increases each year from 2014 to 2019. As our search ends up on 10th June 2020, this number can not represent the overall number of papers in 2020. Especially there are several upcoming top-tier conferences e.g., CVPR, NeurIPS, and ICML, where much more papers may come out later this year.  Even given this situation, the number of papers in 2020 has already exceeded that in 2016. It can be noticed that there is significant rise of papers in 2016 and 2017. Indeed we see lots of exciting research in these two years e.g., CoGAN, f-GAN in 2016 and WGAN, PROGAN in 2017, which pushes the GANs research and exposes GANs to the public. In 2018, GANs still attracts lots of attention and the number of papers is more than that in previous years.

\begin{figure}[ht!]
    \centering
    \includegraphics[width=.6\textwidth]{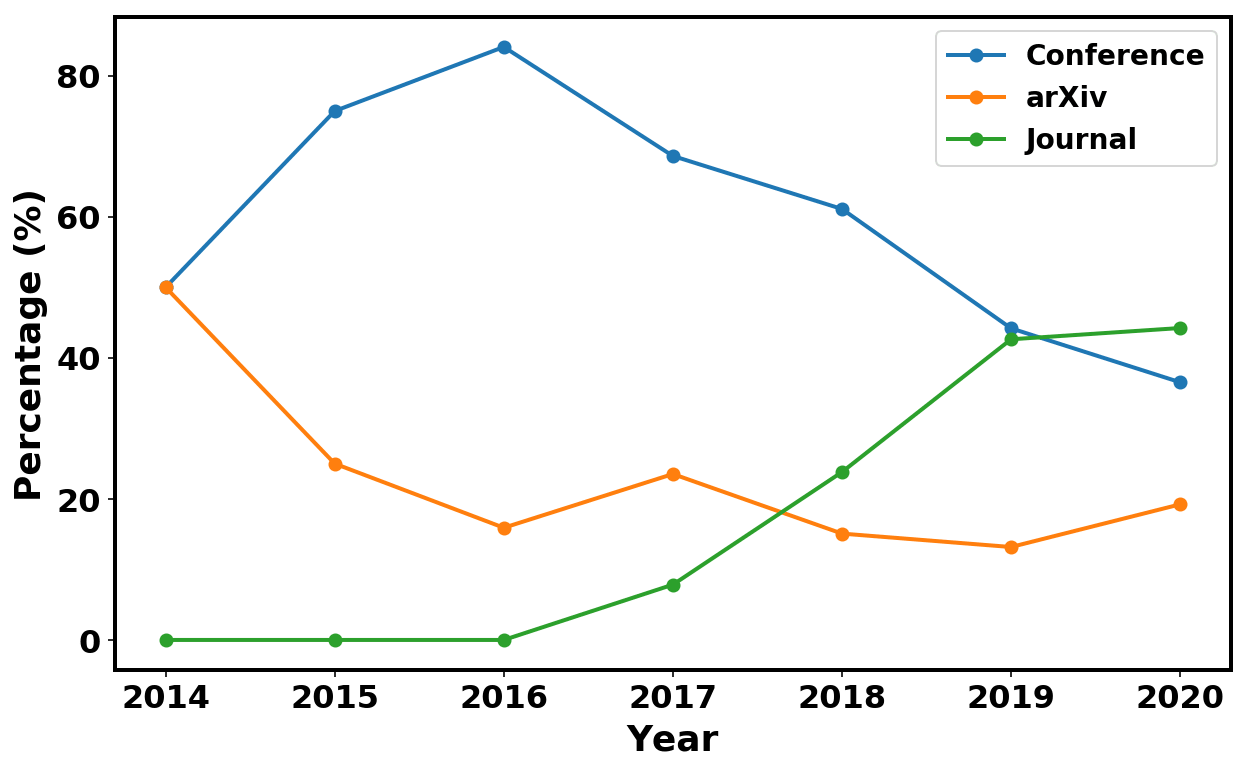}
    \caption{Percentages of each category take account the total number of papers in each year.}
    \label{chap05-fig:paper_stats_percent}
\end{figure}

Figure~\ref{chap05-fig:paper_stats_three} 
illustrates the number of papers published on three repositories, namely conference, arXiv and journal. Conference takes the largest amount from 2015 to 2019 and dramatic increase appears in 2016 and 2017. Papers published on Journal starts to increase since from 2017, which may be caused by the reviewing duration for a journal paper is longer than a conference paper and of course much longer than an arXiv paper. As GANs are well-developed and well-known to researchers from different areas today, number of journal papers related to GANs supposes to maintain the increasing tendency in 2020. It is interesting that number of arXiv pre-prints reaches the peak in 2017 and then starts to descend. We guess this is caused by more and more papers are accepted by conference and journal so arXiv pre-prints claim the publication details, which leads to the decreasing number of pre-prints on arXiv. This indicates higher quality of GANs research in recent years from the other side. Figure~\ref{chap05-fig:paper_stats_percent}
gives an illustration on the percentage of each category taking account the total number of papers in each year. Supporting results in Fig.~\ref{chap05-fig:paper_stats_three}, tendency of number of journal papers keeps going up. Percentage of number of conference papers reaches peak at 2016 then begins to descend. It should be noted that this does not mean the decrease of number of conference papers. This is due to other categories (i.e., arXiv and journal papers) start to increase. 

A detail of searched papers are listed on this link: \url{https://github.com/sheqi/GAN_Review/blob/master/GAN_CV.csv}.
\end{document}